%% file: arxiv.tex
\pdfoutput=1

\documentclass[11pt]{article}

\usepackage{coling}

\usepackage{times}
\usepackage{latexsym}
\usepackage{url}

\usepackage[T1]{fontenc}

\usepackage[utf8]{inputenc}

\usepackage{microtype}

\usepackage{inconsolata}

\usepackage{graphicx}

\input{latex/preamble}

\title{Using Game Play to Investigate Multimodal and Conversational Grounding in Large Multimodal Models}

\author{
 \textbf{Sherzod Hakimov\textsuperscript{1}},
 \textbf{Yerkezhan Abdullayeva\textsuperscript{1}},
 \textbf{Kushal Koshti\textsuperscript{1}},
 \textbf{Antonia Schmidt\textsuperscript{1}},
\\
 \textbf{Yan Weiser\textsuperscript{1}},
 \textbf{Anne Beyer\textsuperscript{1}},
 \textbf{David Schlangen\textsuperscript{1,2}}
\\
\\
 \textsuperscript{1}Computational Linguistics, Department of Linguistics\\University of Potsdam, Germany\\
 \textsuperscript{2}German Research Center for Artificial Intelligence (DFKI), Berlin, Germany
\\
\texttt{firstname.lastname@uni-potsdam.de} 
}

\begin{document}
\maketitle
\begin{abstract}
While the situation has improved for text-only models,
it again seems to be the case currently that multimodal (text and image) models develop faster than ways to evaluate them. In this paper, we bring a recently developed evaluation paradigm from text models to multimodal models, namely evaluation through the goal-oriented game (self) play, complementing reference-based 
and preference-based evaluation.
Specifically, we define games that challenge a model's capability to represent a situation from visual information and align such representations through dialogue. We find that the largest closed models perform rather well on the games that we define, while even the best open-weight models struggle with them. On further analysis, we find that the exceptional deep captioning capabilities of the largest models drive some of the performance. There is still room to grow for both kinds of models, ensuring the continued relevance of the benchmark.
\end{abstract}

\begin{figure}[ht!]
    \centering

\input{latex/matchit/intro_figure}
    \caption{Example dialogue from \textit{MatchIt} game, between players A (turns highlighted in purple) and B (orange), and a programmatic ``game master'' (grey). The task is to identify whether A and B were given the same image or not, via text interaction only. The game master scaffolds the game by prompting the players and relaying information. 
    }
    \vspace*{-.8cm}
    \label{fig:intro_example}
\end{figure}

\section{Introduction}
\label{sec:intro}

Large \textit{multimodal} models (LMMs; such as GPT4o,\footnote{%
    \url{https://openai.com/index/hello-gpt-4o/}
} InternVL \cite{chen2023internvl}) that can handle images as input together with text seem poised to play a significant role in constructing a new, more capable kind of situated interactive agent. What is particularly exciting about them is that they, in contrast to earlier attempts at building such systems, as surveyed by \citet{Suglia2024}, promise to be \textit{generalist} models that can be adapted to tasks at hand through methods that require few or even no data and training cost. Current methods for evaluating them, however, largely do not address this potential, following mostly the \textit{reference-based evaluation} paradigm and probing for reasoning and recognition capabilities in static contexts.

In this paper, we investigate whether a recent new evaluation paradigm for text-based models---\textit{game-agency-based evaluation}---can be transferred to the evaluation of text and image models. Specifically, we selected one of the various frameworks that appeared last year for defining such games, \texttt{clemgame} \cite{chalamalasetti-etal-2023-clembench}, and adapted it to evaluate multimodal models.
We defined three dialogue games (on reference, image comparison, and navigation) that focus on the ability to build a model of a situation that is presented as an image and, in two of them, to align it with a partner. We make the following observations:
\begin{itemize}
    \item Current LMMs are capable of conducting situated interactions if given enough scaffolding by an agent framework.
    \item There are significant differences in the degree of this ability, however, between commercial and open models (43 points on our 0--100 scale between the best of each kind), mirroring the situation that text-only models were in previously \cite{beyer2024clembench2024}.
    \item Much of the performance is driven by the excellent \textit{deep captioning} abilities of the largest models; these break down on very detailed abstract images.
    \item Elementary capabilities for representing spatial configurations (or, more abstractly, graph structures) seem to be present in the larger models.
\end{itemize}

We made the source code for the implemented games and the extended framework publicly available at: \url{https://github.com/clembench/}. The leaderboard of evaluated multimodal LLMs is available here (tab \textit{Multimodal)}: \url{https://clembench.github.io/leaderboard.html}.

\section{Related Work}

\paragraph{Evaluating LLMs}
Following traditional practice in NLP, the first main paradigm for evaluation LLMs was what can be called \textit{reference-based evaluation}, where a model response to a test item is compared to a known correct response. As a reaction to the rapidly increasing scores of the latest models and the saturation of existing benchmarks \cite{superGLUE}, meta-benchmarks have been set up, such as HELM \citep{helm2022} and BIGbench \cite{bigbench2022}. While this method offers control over the tasks that are tested, a recently highlighted problem is ensuring train/test splits in the era of extremely large (and intransparent) training sets \cite{magar-schwartz-2022-data}.
Another popular method for evaluation falls under what could be called the \textit{preference-based evaluation} paradigm. This is represented by Chatbot Arena \cite{chiang2024chatbot}, which lets users present queries to two models in parallel and then ask for preferences. This has the advantage of higher ecological validity as the human/system interaction is being evaluated. However, it comes with the cost of very little control over the tested distribution of tasks. Finally, a newly emerging paradigm is \textit{game-agency-based evaluation} \cite{chalamalasetti-etal-2023-clembench,DBLP:journals/corr/abs-2308-07201,DBLP:journals/corr/abs-2308-10032}.
In this paradigm, evaluation is framed as measuring the success of LLMs in conducting task-oriented interactions in simple conversational games. This has the advantage that it does not require user interaction (unlike the preference-based paradigm) while still keeping goal orientation and strategy in focus. In this paper, we want to explore this paradigm for the evaluation of LMMs.

\paragraph{Evaluating LMMs} 
The evaluation of the newer field of Large Multimodal Models so far mostly remains within the \textit{reference-base evaluation} paradigm,\footnote{
    Although first attempts are underway to establish attempts in the \textit{preference-based evaluation} paradigm as well, with the Multimodality Chatbot Arena \url{http://vlarena.opengvlab.com}. This however seems to be much less popular so far than its text-based counterpart.
}
with datasets such as \textit{MME}~\cite{DBLP:journals/corr/abs-2306-13394}, \textit{MMBench}~\cite{DBLP:journals/corr/abs-2307-06281}, \textit{MMMU}~\citep{mmmu}, and
\textit{SEED-Bench} v1~\citep{seedbench-1} and v2~\citep{seedbench-2}. These include image and text pairs as test instances for various tasks such as question answering, reasoning, answering scientific questions, etc. \textit{VHELM}(visual HELM)~\footnote{Accessed in May 2024. \url{https://crfm.stanford.edu/helm/vhelm/latest/}} uses \textit{MMMU} and two other visual question answering datasets~\citep{DBLP:conf/cvpr/Gurari0SGLGLB18, DBLP:conf/cvpr/GoyalKSBP17} to extend the HELM framework to test multimodal LLMs. Our aim here is not to replace this kind of evaluation but rather to complement it with a focus on different capabilities, or at least differently challenged capabilities (see below).

\ \\[-.5\baselineskip]
\noindent
Before we describe the general structure of our games, we will briefly also review literature relevant to each of them separately.

\paragraph{Reference Games}
The use of reference games where one player gets another to identify an item through verbal means, goes back to at least \citet{krausswein:1964} and \citet{lewis:conv} in linguistics and psycholinguistics, and has seen increased use in NLP in recent years as well \cite{shen-etal-2018-comparing,haber-etal-2019-photobook,sadler-etal-2024-sharing-cost}. Its attraction lies in the very clear way in which it brings out context dependence (a good referring expression not only \textit{describes} the target object, but also \textit{excludes} distractor objects) and, especially in settings where there are repeated references~\citep{DBLP:conf/cvpr/SadovnikCSEC12}, partner-effects as well (precedence; \cite{brenclark:conpact}).

\paragraph{Image Comparison}

The second game that we implement follows a suggestion by \citet{schlangen2019grounded}, who uses it to illustrate the concept of a \textit{Grounded Agreement Game}. The idea here is to go beyond settings like Visual Dialog \cite{das2017visual} or Guesswhat?! \cite{devries2017guesswhat}, which were quite popular in the language and vision field at the time. The criticism in this paper was that these settings while eliciting dialogue in the sense of sequential turns from different speakers, do not provide much purpose to the interaction. Grounded Agreement Games, on the other hand, by letting players share a common goal of reaching mutual understanding, provide a ``\textit{joint purpose}, a shared sense of semantic ownership of the interaction'' \cite{schlangen2019grounded}. 
Also related, in incorporating visual information and cooperation between participants, is spot-the-difference \cite{lopes2018spot}; this corpus, however, has only been used for linguistic analysis.

\paragraph{Navigation and Exploration}

Following natural language navigation instructions is a well-established task in the intersection of Computer Vision and NLP, and many special purpose models have been built in recent years \cite{gu-etal-2022-vision}. The task described below is related but posed to \textit{generalist models}, and tasks the model only with exploration. More abstractly, what is tested is the ability to explore graph structures and testing spatial reasoning abilities~\citep{DBLP:conf/aaai/ShiZL22, DBLP:journals/corr/abs-2406-04566}.
As a task for the assessment of models, something related has been used by
\citet{Momennejad2023EvaluatingCM} in CogEval. Their results suggest that LLMs lack emergent cognitive map comprehension or planning competence, finding that  LLMs can navigate simple graphs but struggle with spatial relationships and complex graphs due to looping, missing edges, and longer trajectories. The NLGraph benchmark \cite{wang2023language} tested LLMs' ability to perform explicit graph reasoning on eight tasks. The models perform basic graph reasoning on cycle and shortest path tasks but fail on the Hamilton Path, according to the NLGraph benchmark assessment. \citet{sparksagi} demonstrated anecdotally that GPT-4 seems to have extensive spatial reasoning and map navigation abilities. This was criticized by \cite{liu2023evaluating}, who showed that responses may become more error-prone when graph density exceeds a certain threshold, potentially causing hallucinations. \citet{DBLP:journals/corr/abs-2406-02537} introduced another dataset for visual map navigation task with where commercial models (GPT-4V, Gemini) struggled with spatial reasoning sub-tasks.

\section{Dialogue Games as Benchmarking Tool}

\begin{figure}
    \centering
    \includegraphics[width=.9\linewidth]{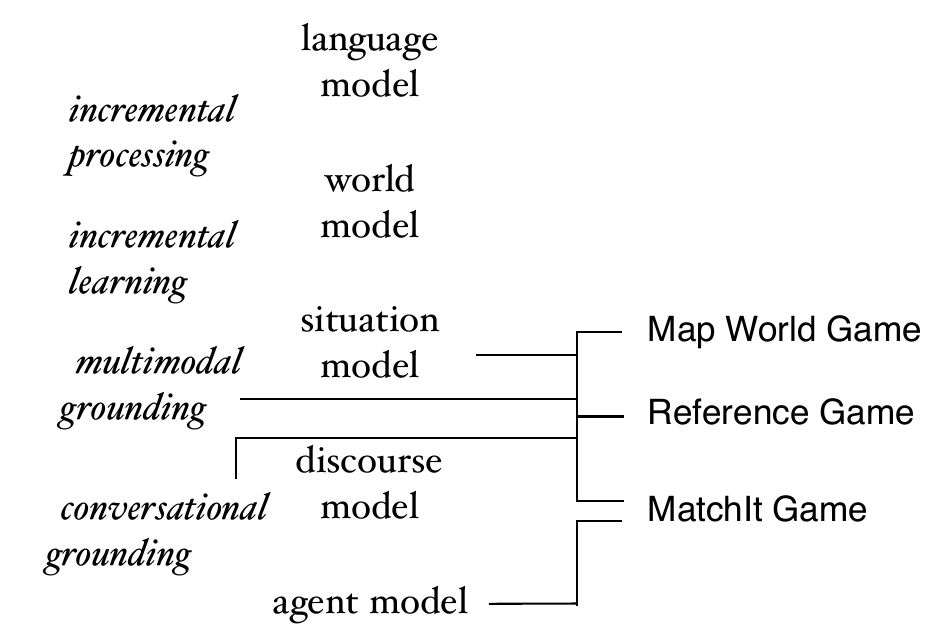}
    \caption{Relating the Dialogue Games used here to the construct model from \citet{schlangen-2023-general}}
        \vspace*{-.4cm}

    \label{fig:construct}
\end{figure}

\paragraph{Construct Validity}
We follow \cite{chalamalasetti-etal-2023-clembench} in striving for construct validity in our measurements and taking inspiration from the model of \cite{schlangen-2023,Schlangen-2023-1}. As can be seen in Figure \ref{fig:construct} (modified from \citet{chalamalasetti-etal-2023-clembench}), we link the games introduced here to the \textit{situation model} (representing the situation in which the task at hand is to be performed), \textit{multimodal grounding} (linking language and visual information), and, at least in simple forms, to \textit{conversational grounding} and the \textit{agent model}. See the original papers for an explanation of the model and the other components. How the individual games challenge these aspects will be explained below.

\paragraph{Scaffolding Game Play with a \texttt{GameMaster}}
We decided to use \texttt{clemgame/clembench} \cite{chalamalasetti-etal-2023-clembench} as the framework to realise the idea of ``self-play for evaluation''.
The main idea of this framework is that dialogue games are specified through \textit{prompt templates}, which explain the game goals to the players in natural language. The game goals include the task description and specific rules for formatting responses (so that they can be parsed). 
A programmatic \texttt{GameMaster} then realises the game play through the instantiation of the templates with specific game instances (e.g., in the game from Figure~\ref{fig:intro_example}, an instance would be defined by a given pair of images), and the turn-by-turn prompting of the \textit{players} (which can be models or human players). 

The resulting \textit{episodes} are then scored through game-specific \textit{scoring rules}. For each game, one scoring metric is determined as the \textit{quality metric} (always ranging from 0 (worst) to 100 (best)). An overall score is computed by averaging this metric by game and then over games. Games where a player violates the parsing rules count as not played (until the end); the percentage of games played hence can serve as a metric for \textit{formatting instruction following ability}, whereas the \textit{quality metric} measures the ability to play the respective game successfully (only for those episodes that were played until the end). We aggregate these two scores to a single number, the \textit{clemscore}, as the quality metric weighted by \% played (scaled to the interval 
$[0, 1]$). 

\section{Three Multimodal Games}

In this section, we describe the three different games that we set up, with a focus on which capabilities exactly they are meant to target.

\subsection{Reference: The Reference Game}\label{sec:reference_game}

\textbf{Game Description}
Player A is presented with three images, and tasked with getting player B, who may see them in a different order, to identify the first of these. Player B is then presented with the three images, potentially in a different order, together with A's reference, and is tasked to identify the referent. This is a single-turn game (Figure~\ref{fig:ref_multimodal_grid}, ~\ref{fig:ref_clevr_aed},~\ref{fig:ref_pentomino}).

\vspace*{.3\baselineskip} \noindent
\textbf{Capability Tested}
The idea is that this game challenges the referring model to go beyond simple descriptions of the image content towards \textit{contrastive descriptions} that exclude the distractor images, and ideally also \textit{efficient descriptions} that do so by concentrating on distinguishing features \cite{Gatt2018}.

\vspace*{.3\baselineskip} \noindent
\textbf{Scoring} 
Each episode is scored as 1 if successful (B picks out the intended referent), 0 otherwise.

\vspace*{.3\baselineskip} \noindent
\textbf{Instances}
We created different sets of instances, with the hypothesis that they might challenge the models differently. First, we created grid-like pixel images (Figure~\ref{fig:ref_multimodal_grid}), which we varied in terms of `compressability': from simple-to-recognise (for humans) patterns to random placements. We created these stimuli in two different renderings: As character-based `images' (hence suitable for text-only models, to allow for a comparison in performance; filled cells are marked with the character ``X''), as well as real images (converted from the text representations).

Second, we selected sets of photographs (Figure~\ref{fig:ref_clevr_aed}) (or photo-realistic renderings) of scenes or configurations of objects, to contrast handling of more naturalistic scenes with the set of grid-images.
We included instances from three datasets: ADE20K~\citep{DBLP:conf/cvpr/ZhouZPFB017}, DOCCI~\cite{onoe2024docci}, CLEVR~\cite{DBLP:conf/cvpr/JohnsonHMFZG17}. We selected one target and two distractors chosen based on the similarity to the target (based on available metadata in each dataset; scene category information in ADE20K, the list of concepts in DOCCI, object categories in CLEVR).

Third, we created boards that include \textit{pentomino} puzzle pieces (Figure~\ref{fig:ref_pentomino}) to analyse whether models are capable of handling unusual shapes and crowded scenes. We take code from \citet{sadler-etal-2024-sharing-cost} and generate a wide variety of scenes, in sets of images with very small differences. From this, we sample randomly.
In total, there are 13 experiments corresponding to 390 instances.

\subsection{Alignment: The MatchIt Game}

\textbf{Game Description}
Player A is presented with an image, as is Player B. The two images are either \textit{identical}, or \textit{different}. The task of the players is to find out which is the case. This game is heavily scaffolded by the \texttt{GameMaster}, which prompts the players to produce a description and ask a question of the other player (Figure~\ref{fig:intro_example}). The dialogue continues with question and answering rounds (where both players ask and answer each other's question) until players make a decision (SAME, DIFFERENT) about the given images (or GameMaster intervenes if maximum number of rounds is reached).

\vspace*{.3\baselineskip} \noindent
\textbf{Capability Tested}
Our hypothesis is that good gameplay requires reasoning about what distinguishing features could be, the presence or absence of which would allow for making the same/different decision. This can then influence both the initial description that is produced and what questions are asked. In principle, allowing more rounds of mutual questioning should make the task easier.

\vspace*{.3\baselineskip} \noindent
\textbf{Scoring} 
Each episode is scored \textit{1} (A and B both make the correct determination) or \textit{0}.

\begin{figure}
    \begin{subfigure}[h]{0.4\linewidth}
        \includegraphics[width=\linewidth]{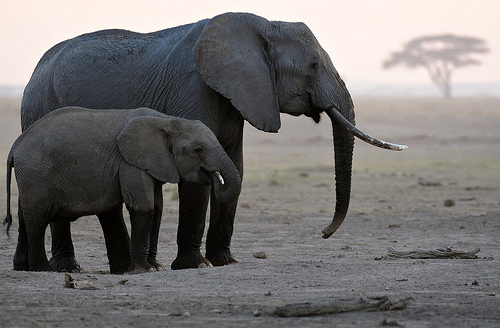}
    \end{subfigure}
    \hfill
    \begin{subfigure}[h]{0.4\linewidth}
        \includegraphics[width=\linewidth]{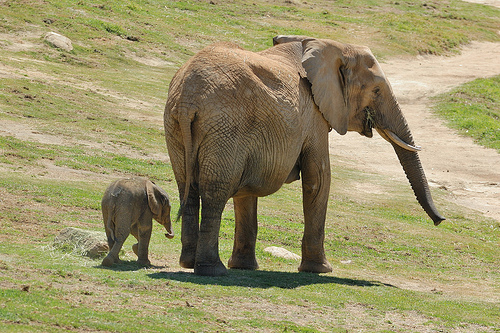}
    \end{subfigure}%
\caption{A pair of similar images for MatchIt.}
\label{fig:matchit_similar_images}
\end{figure}

\vspace*{.3\baselineskip} \noindent
\textbf{Instances}
Three difficulties were defined for the multimodal variant of MatchIt: both players get the same image, both players get similar images or completely different images, the hypothesis being that different images are the easiest to recognize, followed by same and similar image pairs. The curation rationale for a similar picture was that both photos could be described with the same (short) sentence, but their difference should be striking enough that one (short) sentence should be enough. Figure \ref{fig:matchit_similar_images} illustrates a similar image pair.
The images used were taken from the \textit{Visual Genome} dataset \cite{krishna2017visgenome} and sampled for the category of similar photos in a multi-step process via Jaccard similarity of sets of object annotations and their attributes and cosine similarity of \textit{CLIP} image encoder embeddings \cite{radford2021clip}. The detailed process is described in Appendix \ref{sec:matchit_appendix}.
Ten instances of each difficulty (same, similar, different) were part of the final game play, for a total of 30 instances. Finally, we also sampled accordingly from the set of pentomino images described above in Section~\ref{sec:reference_game}. In total, there are six experiments corresponding to 60 instances.

\subsection{Navigation \& Exploration: Map Game}

\begin{figure}
    \centering
     \includegraphics[width = .9\linewidth]{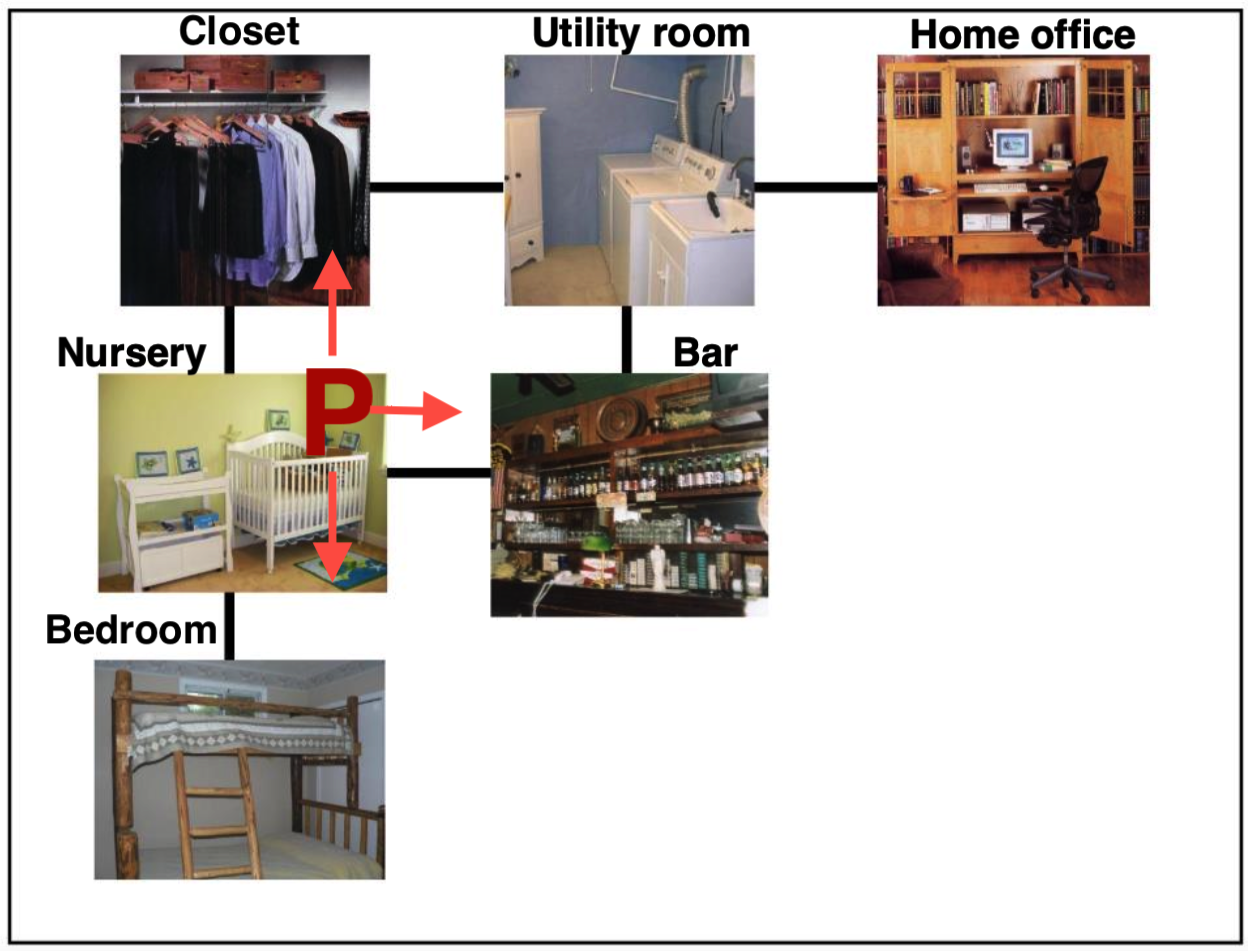}
     \vspace*{-.2cm}
     \caption{Environment for the text-only Map game. The player (denoted with \textbf{P}) is currently in the \textit{Nursery} and has the option to move to one of the neighboring rooms (\textit{Bar, Closet} or \textit{Bedroom}). The player moves by choosing a cardinal direction: east, west, north, south}
     \label{fig:mapworld_graph_text}
\end{figure}

\begin{table*}[htp]
    \centering
    { \scriptsize
    \input{latex/results-resources/results_per_game_macro_avg}
    }
        \vspace*{-.1cm}
    \caption{The ``clemscore'' is calculated as $\texttt{(avg \%p * avg ql) /100}$ where \textit{avg \%p} (average played) is the average percentage of games played to completion, and \textit{avg ql} (average quality score) is the measure of quality of the completed games. Results for Map Game are averaged over three variants of the game. The highest \textit{clemscore}, \textit{avg played} and \textit{quality scores} for commercial and open-weight models and are highlighted in \textcolor{blue}{blue} and \textcolor{teal}{teal}, respectively.
    }
    \vspace*{-.3cm}
    \label{tab:bench-overview}
\end{table*}

\vspace*{.3\baselineskip} \noindent
\textbf{Game Description}
This is a single-player game in which the player explores a network of connected rooms, it is based on the environment ``MapWorld'' of \citet{mapworld}. At any point in the game, the player is in one of the rooms of the network (or \textit{map}). The player can move into adjacent rooms by issuing a navigational command (east, west, north, or south) (see Figure~\ref{fig:mapworld_graph_text}). Information about the room is relayed to the player by the \texttt{GameMaster} by giving the \textit{image} of the room (the name of room, e.g. ``Nursery'', is never revealed); information about the directions in which adjacent rooms can be found is always relayed via text (e.g. ``From here you can go north, south, east.'').  
Within this general setting, we define several versions: \textit{Go to X (G2X)}, in which the player is tasked to find a room of a specific category and indicate when they think the goal has been achieved. \textit{Explore Exhaustively (EE)}, in which the player is tasked with visiting all rooms of the map and indicate when it thinks the goal has been achieved. In \textit{graph reasoning (EE-gr)}, the player is prompted to generate the action along with the representation of the  already explored graph explicitly. %

\vspace*{.3\baselineskip} \noindent
\textbf{Capability Tested}
Unlike in the previous two games, the situation relevant to the game is not observable in one go but rather must be explored actively. To perform well in this family of games, an internal representation of the map must be kept. Moreover, to be efficient, some spatial reasoning over this implicit structure is required to keep track of as yet unexplored rooms.

\vspace*{.3\baselineskip} \noindent
\textbf{Scoring} 
Scoring is more complex in this game. We define a metric for \textit{efficiency}, which measures how many of the performed moves were necessary (see Appendix~\ref{subsec:map_metrics} for the full definition); \textit{question answering}, which measures the percentage of questions answered correctly (in the variant with questions); and \textit{success}, which is 1 if the player ended the game in a success condition (indicated room found / all rooms explored), and 0 otherwise.

\vspace*{.3\baselineskip} \noindent
\textbf{Instances}
\noindent
Experiments on the \textit{EE} version test the effect of map complexity by changing map size and connectedness. The maps can have 4, 6, or 8 rooms, whereas, in the 6 or 8-room case, we distinguish between maps with and without a cyclic path in them, yielding five experiments in total. We expect larger and more connected maps to be harder to explore.
For our \textit{EE-gr} version, we reuse the three experiments on map sizes from above. The goal is to have comparable results and measure the influence of explicit graph reasoning.
On the \textit{G2X} version, we experiment with distances from start to target, either starting on, close to, or far from the target. The hypothesis is that finding a target room nearby is easier than finding it far away. In total, there are five experiments with 50 instances for \textit{EE}, three experiments with 30 instances for \textit{EE-gr}, three experiments with 30 instances for \textit{G2X}.

\section{Results}
\vspace*{-.2cm}

\subsection{Overall Results}
\textbf{Models}: We selected models that i) support multi-turn dialogue and have been optimised to follow chat templates,\footnote{\url{https://huggingface.co/docs/transformers/en/chat_templating}} ii) encode multiple images in a single turn. We benchmarked both open-weight and commercial models. Of commercial models, we decided to evaluate \model{Claude-3.5-Sonnet} (June 2024), \model{Claude-3-Opus} (February 2024), \model{GPT-4-vision} (November 2023), \model{GPT-4o} (May \& August 2024 versions), \model{GPT-4o-mini} (July 2024), and \model{Gemini-1.5-Flash-001} (May 2024).\footnote{%
    We excluded \model{Gemini 1.5 Pro} because querying the API backend resulted in many experiments being timed out, and \textit{Gemini 1.0 Pro} was excluded since it does not support multi-turn dialogue.
}    
From the available open-weight models we selected \model{InternVL2} (8B, 26B, 40B, 76B versions)~\citep{chen2023internvl}, \model{Idefics} (\model{9B, 80B versions})~\cite{idefics}, \model{Idefics-3} (\model{8B-llama})~\cite{laurençon2024mattersbuildingvisionlanguagemodels} \model{InternLM-XComposer-2.5}~\citep{internlmxcomposer2_5}, \model{Phi-vision} (3.0, 3.5 versions)~\citep{DBLP:journals/corr/abs-2404-14219}, \model{Pixtral-12B (2409)}\footnote{\url{https://huggingface.co/mistralai/Pixtral-12B-2409}}. We provide more details about models in Appendix~\ref{sec:model_backends}.

The benchmark results are given in Table~\ref{tab:bench-overview}. What first catches the eye is the significant difference in overall score (\textit{clemscore}) between closed-weight ~/ commercial and open-weight models, with the best open model trailing the worst commercial for 10 points and the best commercial one for 43 points. We can compare this to the situation with text-only games, where \citet{beyer2024clembench2024} report that the best/best distance was 55.25 points in June 2023, 41.18 five months later (November 2023), and in May 2024 was reduced to 24.94. This nicely reflects the somewhat less mature state of LMMs (large multimodal models) compared to LLMs.

What is also striking is that \textit{\% played (p)}, which measures the ability of the models to follow formatting instructions, is generally high; indicating that the scaffolding offered by the \texttt{GameMaster} was strong, but also perhaps that indeed these models are well tuned. We can also see that, in particular, the performance on the \textit{Reference Game} seems to be a differentiator between models; while the commercial models are all in the same level on \textit{MatchIt}, they differ more there (and to a lesser degree also on the \textit{Map Navigation Games}). Overall, the \model{Claude-3.5-sonnet} and \model{GPT-4o (Aug)}, which increased 10 points compared to the May 2024 version, are the best performing commercial models, and the \model{InternVL2} models are the best performing open-weight models.

To investigate further, we turn to a more fine-grained analysis by implementing text-only variants of games.

\begin{figure}[ht]
    \centering
    \includegraphics[width=1.10\linewidth]{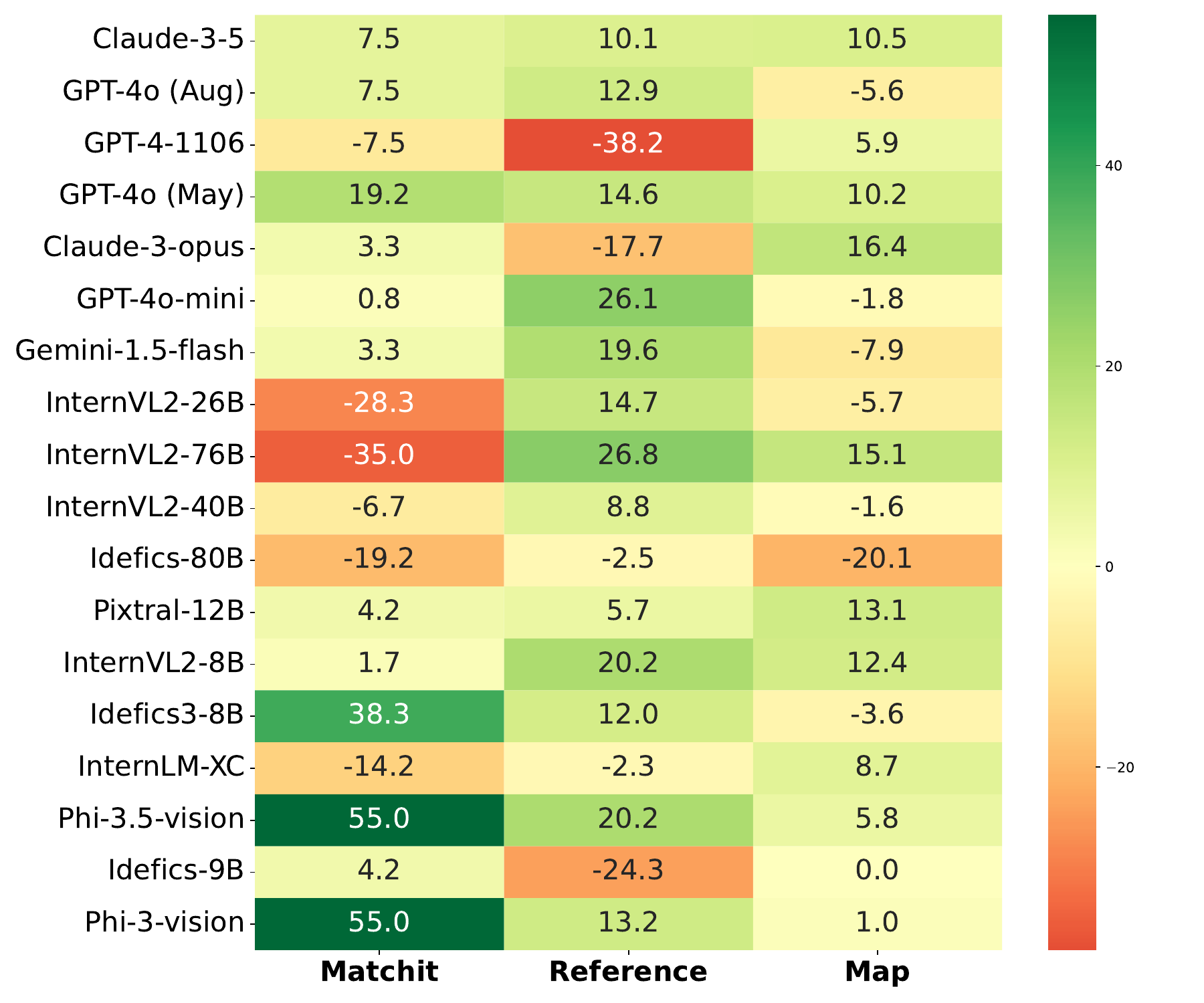}
    \caption{Performance difference in \textit{clemscore} (textual - multimodal) across models and games. \textcolor{green}{\textbf{Green}} values indicate better textual and \textcolor{red}{\textbf{red}} values (negative) indicate better multimodal performance. The values closer to zero (in \textcolor{yellow}{yellow}) indicate that the performance of models are somewhat equal between modality variants.}
    \label{fig:performance-diff}
    \vspace{-0.6cm}
\end{figure}

\subsection{Textual vs. Multimodal Performance}
This section analyses the effect of moving from text-only LLMs to multimodal ones. We implemented text-only versions of three games by representing the tasks in ASCII characters. Each game has been implemented where inputs are represented in only text. For the Reference game, we ran the original ASCII character representation of grids (as in clembench \citep{chalamalasetti-etal-2023-clembench}). For the Matchit game, we used the same ASCII grids (Figure~\ref{fig:matchit_grids_examples}) to create similar/dissimilar experiments.

For the Map Navigation game, we implemented all three versions in text-only variants as follows: once the Player makes a move, the GameMaster provides information about the current room in text format, such as ``You have entered Nursery. From here you can go north, south, east''. In the multimodal version, this information is given as the input image (e.g. Nursery) and then the text ``From here you can go north, south, east'' (without any information about the room in text form).

Next, we ran the benchmark on the models using textual versions of games and compared them against the multimodal results. Figure~\ref{fig:performance-diff} shows the difference in \textit{clemscore} between textual and multimodal scores, where we subtracted the multimodal value from textual one. Higher values (in green) indicate that models are better at textual, while lower values (in red) stand for better performance at multimodal games. In general, we can observe that most models are better at textual games; which perhaps can be explained by the dominance of text data in training datasets over other modalities (images in this case)~\citep{DBLP:journals/corr/abs-2405-19334}. The commercial models such as \model{GPT-4o} (Aug'24) and \model{Claude-3.5} (being the best two models in multimodal games) are also better at textual versions of the games while \model{GPT-4-1106} is worse at the textual version of the Reference game. From the open-weight models, \model{InternVL2-26B} has the best score in multimodal games but clearly struggled with a textual version of the MatchIt game. We can also observe that  \model{InternVL2-40B} is a better choice over \textit{26B} version (or any other open-weight model) as its performance is equally distributed across games for their textual and multimodal versions. The Map Navigation Game has steady performance of almost all models (except \model{Idefics-80B}) being better at textual variants than multimodal ones.

\subsection{Zooming in on the Games}

\begin{figure*}
    \centering
    \includegraphics[width=1.0\textwidth]{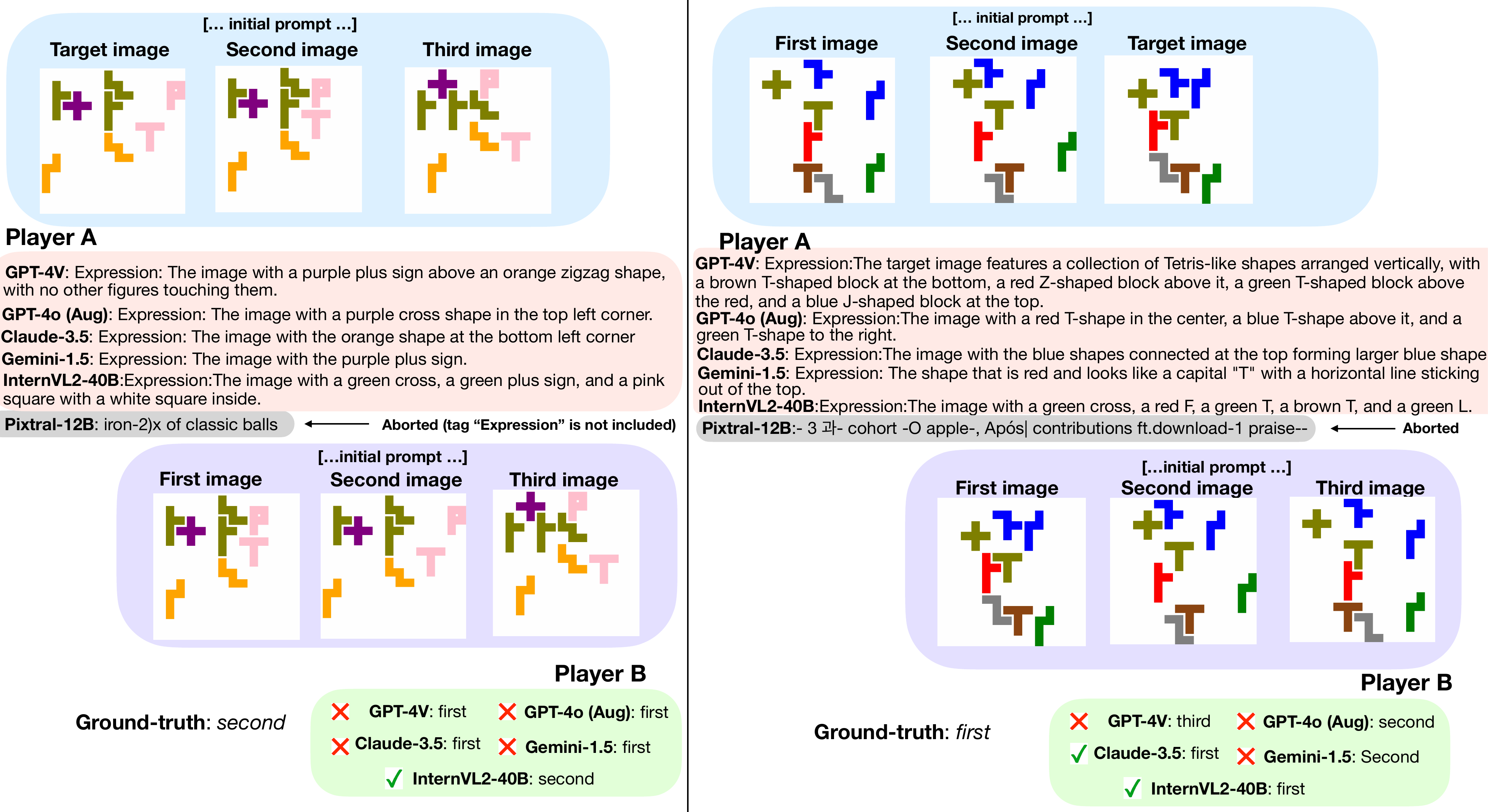}
    \caption{Sample outputs generated by the models for two  Pentomino experiments. The example on the left has the target in the \textit{first position} for \textit{Player A} while the right example has in the \textit{third position}. The order of images for the \textit{Player B} is shuffled.}
    \label{fig:ref_pentomino}
\end{figure*}

In this section, we discuss the individual findings across games by mentioning the hypothesis (H) and the finding (F).

\subsubsection{The MatchIt Game}

The results breakdown in detail is in Appendix~\ref{sec:matchit_appendix}.

\textbf{H:} Pairs of similar images pose the biggest challenge (shown for human players in a similar setting by \citet{sagi2012difference}). \\
\textbf{F:}  Figure~\ref{fig:difficulties_multimodal_barplot} shows that this bears out for the image-based instances. This is less clear for the text-based instances. 

\textbf{H:} Pairs of very different images will be the easiest to recognize (as being different) because the initial description might already make clear the incompatibility.\\
\textbf{F:} This has not been shown. Although for the text-based game variant, the highest scores have been achieved in with different grids, the difference to same grids is not significant. Through all versions of multimodal inputs, the highest scores are achieved in the ``same image'' case. This further indicates that the followed strategy relies on comparative reasoning to a lesser degree than anticipated.

\subsubsection{The Reference Game}

See Appendix~\ref{sec:referencegame_appendix} for detailed analysis.

\textbf{H:} Due to naming difficulties, the task is harder for more abstract images (grids, pentomino pieces) than photos of common scenes (ADE20K, DOCCI).\\
\textbf{F:}
Table~\ref{tab:reference_results} includes detailed results for each individual experiment.
GPT models \& \model{Claude-3.5} (but not \model{Gemini}) get much higher scores on ADE, DOCCI, and CLEVR experiments than on grid and pentomino experiments.
The pentomino experiment results (bad across the board) show that the task is far from being solved. We speculate that this set might touch the limits of the vision encoder and its ability to distinguish objects (of usual kinds). See Figures~\ref{fig:ref_pentomino} for sample outputs for the Pentomino experiment. (see Figure~\ref{fig:ref_multimodal_grid} and \ref{fig:ref_clevr_aed} for ADE, CLEVER experiments).

\textbf{H:} Random images are more challenging to describe than patterns and objects.\\
\textbf{F:} Indeed, the results in Table~\ref{tab:reference_results} show that most models struggled with random grids for both textual and multimodal variants of the game. The same difference can be observed for photo images vs. random collections of pentomino pieces.

\textbf{H:} Given that the base models from which the models were trained are text models, even the resulting models perform better on the text-only renderings of grid experiments than the image ones.\\
\textbf{F:} Results are mixed. Some models are better at ASCII representations (\model{Gemini-1.5, GPT-4o}), while others are better at multimodal representations (\model{Claude-3, GPT-4V}). See Figures~\ref{fig:ref_multimodal_grid} and \ref{fig:ref_textual_grid} for sample outputs.

\textbf{H:} To reach high scores in this game, player A needs to do Referring Expression Generation (REG; \citet{Gatt2018}), as opposed to captioning.\\
\textbf{F:} We were initially surprised by the high scores achieved, in particular by the \model{GPT-4}s. On inspecting the transcripts, it became clear that the model achieves its high performance in parts through its exceptional ability to produce detailed descriptions (especially for the photo sets), thereby reaching a level of detail where a description of the target itself is enough to single it out (see also Appendix~\ref{app:ref_static} for an ablation on the (missing) effect of the distractors). 
This is also evident from the average number of generated tokens, which is 27 for \model{GPT-4V} and 20 for \model{GPT-4o}, as compared to 14 for \model{Gemini} and \model{Claude}. We also find little evidence for the use of negations (``the one without cars''), which can be an efficient REG strategy (although \model{Claude} does produce this occasionally).
As mentioned above, performance breaks down for the pentomino dataset. Overall, this suggests that the game, as currently defined, leans more towards evaluating deep captioning (where there is still room to grow).

\subsubsection{The Map Navigation Games}

\begin{figure*}
    \centering
    \includegraphics[width=0.95\textwidth]{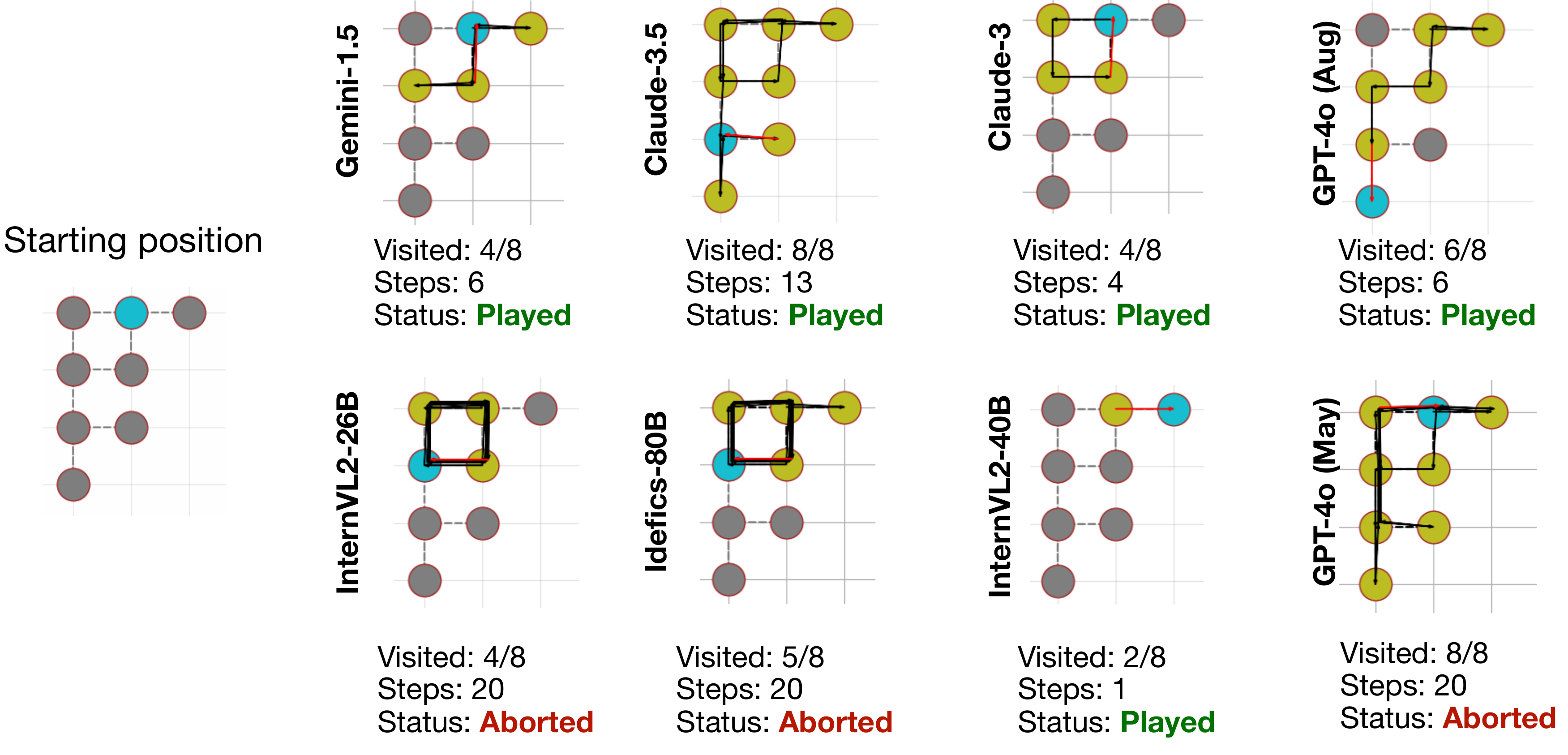}
    \caption{Map navigation samples by selected models for the experiment ``Large with cycle". The currently visited room is marked as cyan, rooms that have been visited are olive colored and the gray rooms have not been visited yet. The game is \textit{Played} when a model decides to stop on its own by generating ``DONE''. The game is \textit{Aborted} if a model generates an output that does not comply with formatting or if the number of turns reaches the maximum limit of 20. The number of visited nodes and total steps are also given for each model.}
    \label{fig:ref_map_examples}
\end{figure*}

As can be seen in Table~\ref{tab:bench-overview} above, of the three games, performance is lowest on the Map Navigation Games, showing an especially pronounced gap between the commercial and the open-weight models, which are only able to finish a much smaller percentage of games (\textit{\% played}). A detailed results breakdown is in Appendix~\ref{sec:map_game_appendix}. Samples are also given in Figure~\ref{fig:ref_map_examples}, which shows how open-access models (InternVL2-26B, Idefics-80B) reach the maximum turn limit because they enter a loop.

\textbf{H}: A larger map makes exhaustive exploration more difficult. There is a higher chance of missing something if there are more things to discover.\\
\textbf{F}: Holds true for smaller models. Looking at the results in Table \ref{tab:mm-mapworld-ee-exp}, a slight downward trend is noticeable. However, some models appear to be behaving differently, showing better results on medium or larger maps compared to small maps.
This is likely due to their thoroughness when it comes to exploration. Models make more steps than the number of nodes in the map, e.g. \model{GPT}s tend to take more redundant steps. The ratio of redundant exploration to useful exploration decreases with larger map sizes, leading to higher scores.\\
\textbf{H}: A more complex map layout (w/ cycles) is harder to navigate.\\
\textbf{F}: Table~\ref{tab:mm-mapworld-ee-exp} seem to indicate that this holds true.
While there is only a marginal difference between maps of medium size with and without cycles, the difference becomes more apparent with large maps.\\
\textbf{H}: In \textit{G2X} (go to specific room), the further away from the starting position the target is, the harder it is to identify it, as more exploration is needed and distractor categories might be encountered.\\
\textbf{F}: The results in Table~\ref{tab:mapworld-g2x-exp} show a clear correlation between distance and success. Not a single model could accurately find every target room at a distance of three or more.\\

Overall, we take these findings as an indication that the game posed a significant challenge to the models, and that successful completion requires sophisticated representational and spatial reasoning abilities.

\section{Conclusions}
\vspace*{-.2cm}

We have transferred a recent evaluation paradigm---game-based evaluation---from the text-only domain to the evaluation of multimodal models. We have set up a set of games that challenge, in different ways, mostly a model's capability to represent (and describe) a \textit{situation}. We have systematically varied the complexity of these situations, as well as how they are given to the model (where we have included, for comparison, purely text-based renderings). We argue that the results on the benchmark are a valid measurement of (aspects of) specific underlying capabilities, which static benchmarks do not address.
We observe a large difference in performance between the largest commercial models and the smaller open-weight models, albeit to a smaller degree than other researchers have observed in the early stages of text-only models. The benchmarks indicate that there is room to grow both for the closed and the open models, while there already is a basis for the development of new kinds of situated interactive systems.

\section*{Acknowledgments}
The work reported here has been partially funded by the Deutsche Forschungsgemeinschaft (DFG, German Research Foundation), grants 423217434 (“RECOLAGE”) and 317633480 (SFB 1287); and by Bundesministerium für Bildung und Forschung (BMBF, German Federal Ministry of Research), project "COCOBOTS" (01IS21102A). We thank the anonymous reviewers for their helpful feedback.

\section*{Limitations}

The first and biggest limitation is that the prompts that define the games are only given in, and hence the results are restricted to, English, even though several of the tested models are listed as being able to process other languages as well. While we have yet to do this, translating the prompts and measuring their impact should be straightforward; we plan to do this in future work.

As discussed in the text above, some of the findings are limited to certain respect by the fact that excellent capabilities of providing image captions open up simpler strategies than what we initially wanted to challenge. While this doesn't impact the significance of the measurements---there is still room to grow, clearly so for the open weight models, but also for the closed one---it should again be straightforward to modify the games so that interactional phenomena (such as valuing efficiency in producing referring expressions in the \textit{Reference Game}, and putting weight on the questioning in the \textit{MatchIt Game}) are further emphasised. Similarly, the amount of scaffolding provided by the \model{GameMaster} is quite high (e.g., in the \textit{MatchIt} game, it determines much of the strategy), which limits the amount to which we gain insight into the strategic abilities of the models. But again, reducing it in future versions of the game should be straightforward.

\section*{Ethics Statement}
Using paid proprietary APIs with underlying models about which little is known (training data, model architecture) in academic research is less than ideal. At the moment, the models tested here seem to be the only ones that are able to follow the structure of the games. It is our hope that open models will catch up soon on multimodal tasks, and proper research can be done with them.

\bibliography{references,anthology}

\appendix

\newpage

\section{Model Evaluation Details}\label{sec:model_backends}
\input{latex/model-details/model_details_appendix}

\section{A Picture Reference Game}\label{sec:referencegame_appendix}

\input{latex/reference/appendix}

\section{An Agreement Game: MatchIt}
\label{sec:matchit_appendix}

\subfile{latex/matchit/3_appendix}

\section{A Map Navigation Game}\label{sec:map_game_appendix}

\input{latex/mapworld/mapworld}

\end{document}

%% file: latex/preamble.tex
\usepackage{times}
\usepackage{latexsym}
\usepackage{enumitem}
\usepackage{listings}
\usepackage{url}
\usepackage[T1]{fontenc}

\usepackage[utf8]{inputenc}

\usepackage{microtype}

\usepackage{inconsolata}

\usepackage{longtable}
\usepackage{graphicx}

\usepackage{caption}
\usepackage{subcaption}

\usepackage{colortbl}
\usepackage{makecell}
\usepackage{multirow}
\usepackage{supertabular}
\usepackage{placeins}

\usepackage{subfiles}
\usepackage{enumitem}

\usepackage{algorithm}
\usepackage{algpseudocode}
\usepackage{amsmath}

\usepackage{cleveref}

\setlist[itemize]{leftmargin=1em}  %

\usepackage{booktabs}

\usepackage{tcolorbox}
\newcounter{promptno}[section]
\newlength\mystoreparindent
\newenvironment{prompt}[1][]
{
  \setlength{\mystoreparindent}{\the\parindent}
  \setlength{\parindent}{0pt}
  \refstepcounter{promptno}
  \par\medskip
  \noindent
  \begin{tcolorbox}[left=1pt,right=1pt]
  \textsc{{template \small\thesubsection.\thepromptno}}\\
  \small
  \tt
}{
  \end{tcolorbox}
  \setlength{\parindent}{\mystoreparindent}
  \medskip
}

\newcounter{utterance}
\usepackage{amssymb}

\DeclareUnicodeCharacter{25A2}{$\square$}
\usepackage{color}

\newcolumntype{T}{>{\ttfamily}l}

\newcommand{\model}[1]{\texttt{#1}}

\tcbuselibrary{skins}

\definecolor{a-colour}{RGB}{135,74,175}
\definecolor{b-colour}{RGB}{243,145,9}
\definecolor{gm-colour}{RGB}{128,128,128}
\definecolor{wordlegreen}{rgb}{0.4,0.79,0.17}
\definecolor{wordlered}{rgb}{0.86,0.31,0.29}
\definecolor{wordleyellow}{rgb}{1,1,0}

\def \dcolwidth {3cm}
\def \skipcolwidth {0.8cm}
\def \arccorner {3mm}
\def \disttocorner {6pt}

\newtcolorbox[use counter=nbubbles]{a-gm}[1]{
    colback=a-colour!10!white,
    colframe=a-colour,
    fonttitle=\bfseries\tiny,
    fontupper=\footnotesize,
    title={#1},
    sharp corners=west,
    arc=\arccorner,
    width=\dcolwidth,
    left skip=0cm,
    top=0pt,
    bottom=0pt,
    left=0pt,
    right=\disttocorner,
    before={\vspace{-0.1cm}},
    boxrule=0.5pt,
    enhanced,
    attach boxed title to top left={yshift=-0.1mm},
    boxed title style={size=small,colback=a-colour},
    overlay unbroken and first = {
    \node[text width=0.4cm,draw=black,line width=0.1mm,align=center,gray] at (-0.5,0.2) {\footnotesize \thetcbcounter};
  }
}

\newtcolorbox[use counter=nbubbles]{gm-a}[1]{
    colback=gm-colour!5!white,
    colframe=gm-colour,
    fonttitle=\bfseries\tiny,
    fontupper=\footnotesize,
    title={#1},
    sharp corners=east,
    arc=\arccorner,
    width=\dcolwidth + \skipcolwidth,
    left skip=\skipcolwidth,
    top=0pt,
    bottom=0pt,
    left=\disttocorner,
    right=0pt,
    before={\vspace{-0.1cm}},
    boxrule=0.5pt,
    halign title=flush right,
    enhanced,
    attach boxed title to top right={yshift=-0.1mm},
    boxed title style={size=small,colback=gm-colour},
    overlay unbroken and first = {
    \node[anchor=north east,text width=0.4cm,draw=black,line width=0.1mm,align=center,gray] at (-0.9,0.55) {\footnotesize\thetcbcounter};}
}

\newtcolorbox[use counter=nbubbles]{b-gm}[1]{
    colback=b-colour!10!white,
    colframe=b-colour,
    fonttitle=\bfseries\tiny,
    fontupper=\footnotesize,
    title={#1},
    sharp corners=east,
    arc=\arccorner,
    width=\dcolwidth + \skipcolwidth + \dcolwidth +  \skipcolwidth,
    left skip=\dcolwidth + \skipcolwidth + \skipcolwidth,
    top=0pt,
    bottom=0pt,
    left=\disttocorner,
    right=0pt,
    before={\vspace{-0.1cm}},
    boxrule=0.5pt,
    halign title=flush right,
    enhanced,
    attach boxed title to top right={yshift=-0.1mm},
    boxed title style={size=small,colback=b-colour},
    overlay unbroken and first = {
    \node[anchor=north east,text width=0.4cm,draw=black,line width=0.1mm,align=center,gray] at (-4.7,0.55) {\footnotesize\thetcbcounter};}
}

\newtcolorbox[use counter=nbubbles]{gm-b}[1]{
    colback=gm-colour!5!white,
    colframe=gm-colour,
    fonttitle=\bfseries\tiny,
    fontupper=\footnotesize,
    title={#1},
    sharp corners=west,
    arc=\arccorner,
    width=\dcolwidth + \skipcolwidth + \dcolwidth,
    left skip=\dcolwidth + \skipcolwidth,
    top=0pt,
    bottom=0pt,
    left=0pt,
    right=\disttocorner,
    before={\vspace{-0.1cm}},
    boxrule=0.5pt,
    enhanced,
    attach boxed title to top left={yshift=-0.1mm},
    boxed title style={size=small,colback=gm-colour},
    overlay unbroken and first = {
    \node[anchor=north east,text width=0.4cm,draw=black,line width=0.1mm,align=center,gray] at (-3.9,0.55) {\footnotesize\thetcbcounter};}
}

\newtcolorbox[use counter=nbubbles]{gm-gm}[1]{
    colback=gm-colour!5!white,
    colframe=gm-colour,
    fonttitle=\bfseries\tiny,
    fontupper=\footnotesize,
    title={#1},
    sharp corners,
    width=\dcolwidth + \dcolwidth + \skipcolwidth + \skipcolwidth,
    leftright skip=\dcolwidth,
    top=0pt,
    bottom=0pt,
    left=0pt,
    right=0pt,
    before={\vspace{-0.1cm}},
    boxrule=0.5pt,
    halign title=center,
    enhanced,
    attach boxed title to top center={yshift=-0.1mm},
    boxed title style={size=small,colback=gm-colour},
    overlay unbroken and first = {
    \node[anchor=north east,text width=0.4cm,draw=black,line width=0.1mm,align=center,gray] at (-3.1,0.55) {\footnotesize\thetcbcounter};}
}

%% file: latex/matchit/intro_figure.tex
\twocolumn

{ \footnotesize  \setcounter{utterance}{1}
\setlength{\tabcolsep}{0pt}
\begin{supertabular}{c@{$\;$}|p{.15\linewidth}@{}p{.15\linewidth}p{.15\linewidth}p{.15\linewidth}p{.15\linewidth}p{.15\linewidth}}

    \hline 
    
    \theutterance \stepcounter{utterance}

    & & \multicolumn{4}{p{0.6\linewidth}}{\cellcolor[rgb]{0.9,0.9,0.9}{%
	\makecell[{{p{\linewidth}}}]{%
   \centering{... <TASK DESCRIPTION> ...\\}
        \centering{\includegraphics[width=3.5cm]{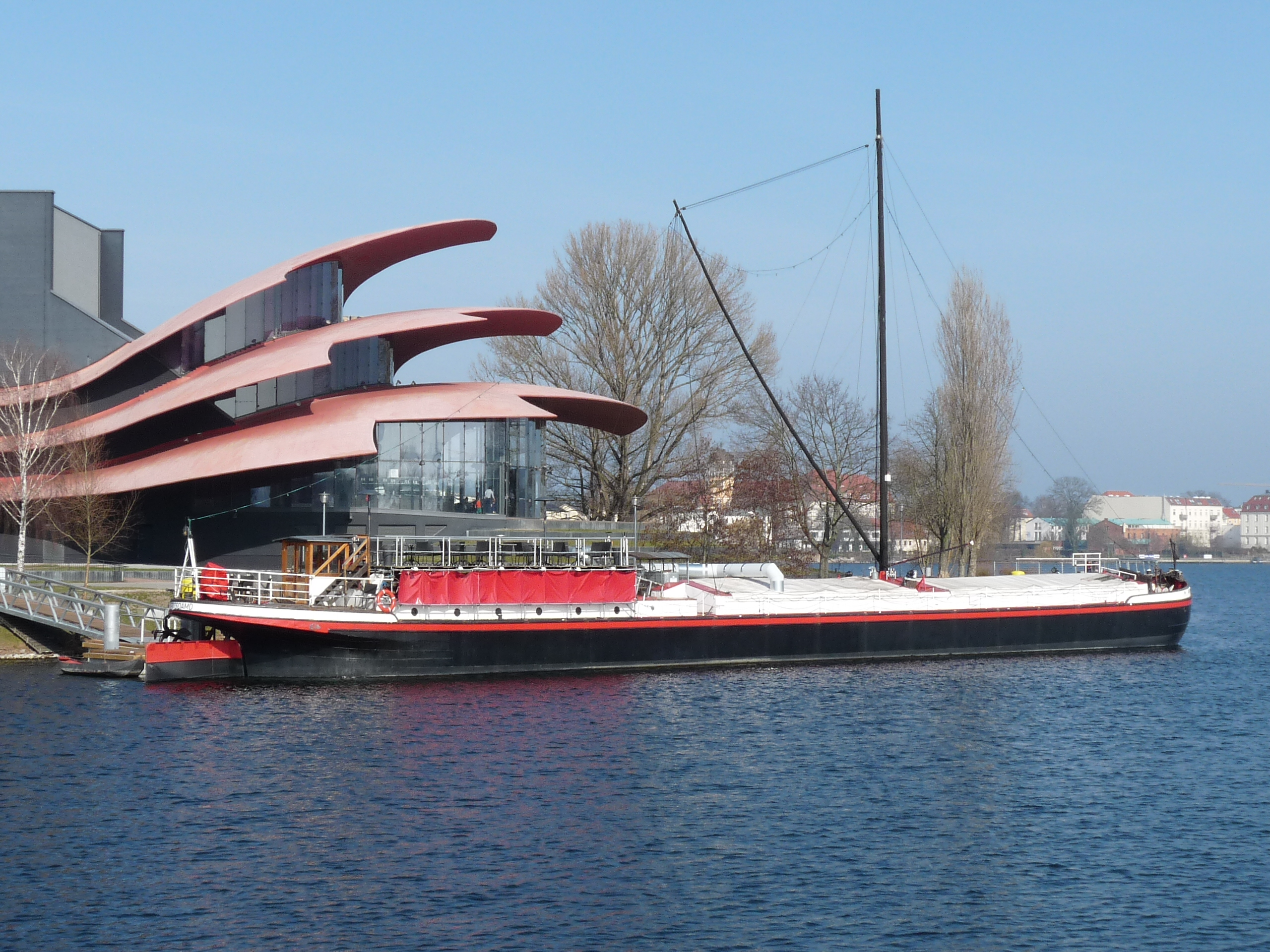}}
	  } 
	   } 
	   }
	 & \\ 

    \theutterance \stepcounter{utterance}  

    & & \multicolumn{4}{p{0.6\linewidth}}{\cellcolor[rgb]{0.9,0.9,0.9}{%
	\makecell[{{p{\linewidth}}}]{%
	  \tt {\tiny [A$\leftarrow$GM]}  
	 Describe your image 
	  } 
	   } 
	   } 
	 & \\

    \theutterance \stepcounter{utterance}  

    & \multicolumn{4}{p{0.6\linewidth}}{\cellcolor[rgb]{0.7,0.7,0.9}{%
	\makecell[{{p{\linewidth}}}]{%
	  \tt {\tiny [A$\rightarrow$B]}  
	  DESCRIPTION: Scene of a river in a city, and a boat.
	  } 
	   } 
	   } 
	 & & \\

\theutterance \stepcounter{utterance}  

    & & \multicolumn{4}{p{0.6\linewidth}}{\cellcolor[rgb]{0.9,0.9,0.9}{%
	\makecell[{{p{\linewidth}}}]{%
         \centering{... <TASK DESCRIPTION> ...\\}
         \centering{\includegraphics[width=3.5cm]{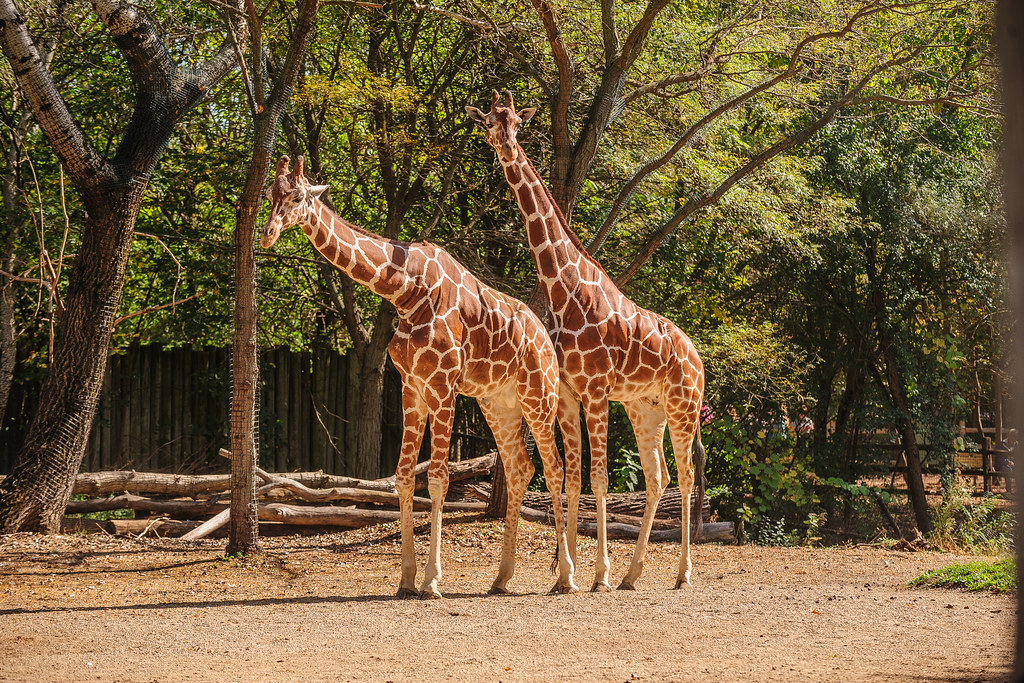}}
	  } 
	   } 
	   } 
	 & \\ 

    \theutterance \stepcounter{utterance}  

    & & \multicolumn{4}{p{0.6\linewidth}}{\cellcolor[rgb]{0.9,0.9,0.9}{%
	\makecell[{{p{\linewidth}}}]{%
	  \tt {\tiny [GM$\rightarrow$B]} 
	 Describe your image
	  } 
	   } 
	   } 
	 & \\

    \theutterance \stepcounter{utterance}  

    & & & \multicolumn{4}{p{0.6\linewidth}}{\cellcolor[rgb]{1,0.85,0.72}{%
	\makecell[{{p{\linewidth}}}]{%
	  \tt {\tiny [A$\leftarrow$B]}  
	  DESCRIPTION: Two giraffes in a dirt field.
	  } 
	   } 
	   } 
	  \\

    \theutterance \stepcounter{utterance}  

    & & \multicolumn{4}{p{0.6\linewidth}}{\cellcolor[rgb]{0.9,0.9,0.9}{%
	\makecell[{{p{\linewidth}}}]{%
	  \tt {\tiny [GM$\rightarrow$B]}
	Ask a question. 
	  } 
	   } 
	   } 
	 & \\

    \theutterance \stepcounter{utterance}  

    & & & \multicolumn{4}{p{0.6\linewidth}}{\cellcolor[rgb]{1,0.85,0.72}{%
	\makecell[{{p{\linewidth}}}]{%
	  \tt {\tiny [A$\leftarrow$B]}  
	  QUESTION: Is there any animal visible in your image? 
	  } 
	   } 
	   } 
	  \\  

    \theutterance \stepcounter{utterance}  

    & \multicolumn{4}{p{0.6\linewidth}}{\cellcolor[rgb]{0.7, 0.7, 0.9}{%
	\makecell[{{p{\linewidth}}}]{%
	  \tt {\tiny [A$\rightarrow$B]}  
	  ANSWER: No, only boats. 
	  } 
	   } 
	   } 
	 & & \\  

    \theutterance \stepcounter{utterance}  

    & & \multicolumn{4}{p{0.6\linewidth}}{\cellcolor[rgb]{0.9,0.9,0.9}{%
	\makecell[{{p{\linewidth}}}]{%
	  \tt {\tiny [A$\leftarrow$GM]}  
	 Ask a question.
	  } 
	   } 
	   } 
	 & \\

    \theutterance \stepcounter{utterance}  

    & \multicolumn{4}{p{0.6\linewidth}}{\cellcolor[rgb]{0.7, 0.7, 0.9}{%
	\makecell[{{p{\linewidth}}}]{%
	  \tt {\tiny [A$\rightarrow$B]}  
	 QUESTION: Is there a body of water in your image? 
	  } 
	   } 
	   } 
	 & & \\

    \theutterance \stepcounter{utterance}  

    & & & \multicolumn{2}{p{0.3\linewidth}}{\cellcolor[rgb]{1,1,1}{%
	\makecell[{{p{\linewidth}}}]{%
	  \tt
	 .......
	  } 
	   } 
	   } 
	 & & \\

    \theutterance \stepcounter{utterance}  

    & & \multicolumn{4}{p{0.6\linewidth}}{\cellcolor[rgb]{0.9,0.9,0.9}{%
	\makecell[{{p{\linewidth}}}]{%
	  \tt {\tiny [GM$\rightarrow$B]}  
	 Come to a decision. 
	  } 
	   } 
	   } 
	 & \\

    \theutterance \stepcounter{utterance}  

    & & & \multicolumn{4}{p{0.6\linewidth}}{\cellcolor[rgb]{1,0.85,0.72}{%
	\makecell[{{p{\linewidth}}}]{%
	  \tt {\tiny [GM$\leftarrow$B]}  
	 DECISION: different images.
	  } 
	   } 
	   } 
	  \\

    \theutterance \stepcounter{utterance}  

    & & \multicolumn{4}{p{0.6\linewidth}}{\cellcolor[rgb]{0.9,0.9,0.9}{%
	\makecell[{{p{\linewidth}}}]{%
	  \tt {\tiny [A$\leftarrow$GM]}  
		Come to a decision.
	  } 
	   } 
	   } 
	 & \\

    \theutterance \stepcounter{utterance}  

    & \multicolumn{4}{p{0.6\linewidth}}{\cellcolor[rgb]{0.7, 0.7, 0.9}{%
	\makecell[{{p{\linewidth}}}]{%
	  \tt {\tiny [A$\rightarrow$GM]}  
	 DECISION: different images. 
	  } 
	   } 
	   } 
	 & & \\

    \theutterance \stepcounter{utterance}  

    & & & \multicolumn{2}{p{0.3\linewidth}}{\cellcolor[rgb]{0.95,0.95,0.95}{%
	\makecell[{{p{\linewidth}}}]{%
	  \tt {\tiny [GM$|$GM]}  
	 SUCCESS 
	  } 
	   } 
	   } 
	 & & \\ 
 
\end{supertabular}
}

%% file: latex/results-resources/results_per_game_macro_avg.tex
\begin{tabular}{l|ccc|cc|cc|cc}
      & \multicolumn{1}{l}{} & \multicolumn{1}{l}{} & \multicolumn{1}{l}{} & \multicolumn{2}{c}{\textbf{MatchIt}} & \multicolumn{2}{c}{\textbf{Reference}} & \multicolumn{2}{c}{\textbf{Map Game}} \\ \hline
\textbf{Model} & \textit{clemscore}            & avg \textit{\%p}         & avg \textit{ql}    & avg \textit{\%p}       & avg \textit{ql}      & avg \textit{\%p}        & avg \textit{ql}       & avg \textit{\%p}       & avg \textit{ql}       \\ \hline

\textcolor{blue}{\textbf{Claude-3.5}}& \textcolor{blue}{\textbf{80.77}}& 95.33& \textcolor{blue}{\textbf{84.73}}& 100.0& 85.0& 100.0& 81.03& 92.22& 85.88\\ 
GPT-4o (Aug)& 80.04& 96.93& 82.57& 93.33& 80.36& 100.0& 74.87& 97.11& 85.87\\ 
GPT-4-1106& 73.55& 97.79& 75.21& 100.0& 80.0& 98.97& 68.39& 96.67& 75.89\\ 
GPT-4o (May)& 69.56& 87.73& 79.29& 100.0& 78.33& 100.0& 75.38& 79.56& 80.91\\ 
Claude-3-opus& 68.16& \textcolor{blue}{\textbf{99.33}}& 68.62& 100.0& 81.67& 100.0& 47.18& 98.89& 71.41\\ 
GPT-4o-mini& 58.46& 90.04& 64.93& 100.0& 86.67& 98.21& 48.04& 84.0& 63.32\\ 
Gemini-1.5-flash& 47.73& 85.0& 56.15& 85.0& 84.31& 100.0& 41.54& 80.0& 51.64\\  \hline
\textcolor{teal}{\textbf{InternVL2-26B}}& \textcolor{teal}{\textbf{37.45}}& \textcolor{teal}{\textbf{66.76}}& 56.09& 100.0& 93.33& 85.13& 34.34& 49.56& 50.93\\ 
InternVL2-76B& 33.84& 54.8& 61.76& 100.0& 90.0& 100.0& 34.36& 24.67& 61.48\\ 
InternVL2-40B& 32.23& 56.27& 57.28& 96.67& 79.31& 100.0& 36.15& 28.22& 56.97\\ 
Idefics-80B& 29.55& 58.29& 50.7& 88.14& 55.77& 100.0& 33.59& 34.44& 54.71\\ 
Pixtral-12B& 28.64& 49.98& 57.3& 100.0& 63.33& 79.23& 44.66& 23.55& 59.51\\ 
InternVL2-8B& 23.17& 46.61& 49.7& 100.0& 68.33& 86.41& 37.09& 15.55& 0\\ 
Idefics3-8B& 17.52& 32.59& 53.76& 40.0& 79.17& 98.97& 31.09& 8.0& 0\\ 
InternLM-XC& 16.95& 20.18& \textcolor{teal}{\textbf{83.98}}& 98.33& 77.97& 2.56& 90.0& 0.0& 0\\ 
Phi-3.5-vision& 15.64& 40.67& 38.46& 100.0& 0.0& 100.0& 15.38& 1.11& 0\\ 
Idefics-9B& 12.29& 38.0& 32.34& 100.0& 33.33& 90.0& 31.34& 0.0& 0\\ 
Phi-3-vision& 3.34& 5.06& 65.98& 0.0& 0& 17.95& 100.0& 2.44& 0\\ \hline

\end{tabular}

%% file: latex/model-details/model_details_appendix.tex
\begin{center}
\begin{table}[h!]
\resizebox{\columnwidth}{!}{%
    \centering
    \begin{tabular}{lll|ll}
        \hline
        \textbf{Model} & \multicolumn{2}{c|}{Base Language Model} & \multicolumn{2}{c}{Base Image Processor} \\
        \cmidrule{2-3} \cmidrule{4-5}
         & Name & Parameters & Name & Parameters \\
         \multicolumn{5}{c}{} \\[-1em]   
        \hline
         Idefics-80B & LLaMA  & 65B & laion/CLIP-ViT & 630M \\
         Idefics-9B & LLaMA & 7B & laion/CLIP-ViT & 630M \\
         InternVL2-76B & Hermes2-Llama3 & 70B & InternViT-V1-5  & 6B\\
         InternVL2-40B & Hermes2-Yi & 34B & InternViT-V1-5  & 6B\\
         InternVL2-26B & Internlm2 & 20B & InternViT-V1-5 & 6B\\
         InternVL2-8B & Internlm2\_5 & 7B & InternViT-448px  & 300M \\
         Idefics3-8B & Llama-3.1 & 8B & SigLIP & 400M\\
         Internlm-XC & InternLM2  & 7B & ViT-L/14 & 428M\\
         Phi-3.5-vision & Phi-3.5 & 3.8B & Phi3VProcessor & 400M \\
         Phi-3-vision & Phi-3-mini & 3.8B & Phi3VProcessor & 400M \\
         Pixtral-12B & Nemo-12B & 12B & - & 400M \\
        
        \hline
    \end{tabular}
}
\caption{Internal details for open-weight models}
\label{tab:internal-model-details}
\end{table}
\end{center}

This section provides a detailed view of the models present in our primary results, along with their utilization. As outlined in the main text, seven commercial (Claude-3.5, Claude3, Gemini-1.5-Flash, GPT-4o (Aug'24), GPT-4o (May'24), GPT-4o-mini, GPT-4-1106) and 11 open-weight models (Idefics-9B, Idefics-80B, Idefics-3-8B, InternVL2-76B, InternVL2-40B, InternVL2-26B, InternVL2-8B, Internlm-XC, Phi-3.5-vision, Phi-3-vision, Pixtral-12B) are included in the primary results. Detailed information about these models can be found in Table \ref{tab:model-details-appendix}. 

\begin{table*}[!htbp]
\centering
\scriptsize
    \begin{tabular}{p{2.0cm}p{1cm}p{1cm}p{2.2cm}p{1.5cm}p{1cm}p{2cm}p{2.3cm}} 
     \hline
     Model Name & Parameters & Context length & Image resolution (pixels) & Release date & Commercial & Backend & Training data Cut-off date \\
     \hline
     Claude-3.5-sonnet & - & 200K & 1568x1568 &  Jun 2024  & \checkmark & anthropic & Aug 2023 \\
     Claude-3-opus & 2T$^*$ & 200K & 1568x1568 &  Mar 2024  & \checkmark & anthropic & Aug 2023 \\
     Gemini-1.5-flash & - & 1048K & - & Apr 2024 & \checkmark & google & Nov 2023 \\
     Gpt-4o (May) & 1.76T$^*$ & 128K & 768x2000 &  May 2024 & \checkmark & openai & Oct 2023 \\
     Gpt-4o (August) & 1.76T$^*$ & 128K & 768x2000  & Aug 2024 & \checkmark & openai & Oct 2023 \\
     Gpt-4o-mini & - & 128K & 768x2000  & Jul 2024 & \checkmark & openai & Oct 2023 \\
     Gpt-4V-1106 & 1.76T$^*$ & 128K & 768x2000  & Nov 2023 & \checkmark & openai & Apr 2023 \\
    Idefics-80B & 80B & 2K & 224x224  & Aug 2023 & $\times$ & huggingface & Feb 2023 \\
     Idefics-9B & 9B & 2K & 224x224 & Aug 2023 & $\times$ & huggingface & Feb 2023 \\
     Pixtral-12B & 12B & 128K & 1024x1024 & Sep 2024 & $\times$ & huggingface & - \\
     InternVL2-76B & 76B & 8K & 448x448 & Aug 2023 & $\times$ & huggingface & Feb 2023 \\
     InternVL2-40B & 40B & 8K & 448x448 & Aug 2023 & $\times$ & huggingface & Feb 2023 \\
     InternVL2-26B & 26B & 32K & 448x448 & Aug 2023 & $\times$ & huggingface & Feb 2023 \\
     InternVL2-8B & 8B & 32K & 448x448 & Aug 2023 & $\times$ & huggingface & Feb 2023 \\
     Idefics3-8B-llama & 8B & 128K & 384x384 & Aug 2024 & $\times$ & huggingface & - \\
     Internlm-XC & 7B & 24K & 224x224 & Jul 2024 & $\times$ & huggingface & - \\
     Phi-3.5-vision & 4B & 128K & 1344x1344 & Aug 2024 & $\times$ & huggingface & Aug 2024 \\
     Phi-3-vision & 4B & 128K & 1344x1344 & May 2024 & $\times$ & huggingface & Apr 2024 \\
     \hline
    \end{tabular}

\caption{Commercial and open-weight model details. Image resolution - indicates the maximum resolution of the image including any scaling.  Supports multiple images - indicates whether the model can process multiple images in a single turn, such as in multimodal reference game. Backend - specifies the underlying script used to access the model for gameplay. Dashes (-) denote information that is not publicly available. $^*$ denotes estimated parameter size.}
\label{tab:model-details-appendix}
\end{table*}

\subsection{Image resolution limits}
Image resolution in Table \ref{tab:model-details-appendix} defines the maximum scaled-down resolution during image processing. For Claude-3,\footnote{\href{https://docs.anthropic.com/en/docs/vision}{https://docs.anthropic.com/en/docs/vision}} though the maximum image resolution is 1568 \(\times\) 1568 pixels, it does allow images upto 8k \(\times\) 8k which are then scaled down to 1568. If any edge of the image exceeds 8k pixels, the model rejects the image. Another constraint specified by Anthropic is the token limit, capped at 1600 tokens per image. Other commercial models only specify the maximum scaled-down resolution without specifying explicit pixel or token constraints. Considering open-weight models, these models specify just one image resolution, and if any image exceeds this, it will be scaled down.

\subsection{Compute Details}
The commercial models were used by integrating their APIs with the \textit{clembench} backend,\footnote{\href{https://github.com/clp-research/clembench/tree/main/backends}{https://github.com/clp-research/clembench/tree/main/backends}}. For open-weight models, inferencing was conducted on a local cluster, which comprises four Nvidia A100 (80GB) GPUs. The model weights were distributed evenly on each GPU, and the models used their default precision values. No weights were offloaded to the CPU during inference. The open-weight models considered here are their HuggingFace-compatible versions, loaded via Auto Classes methods to maintain generalizability, which leads to straightforward integration of additional models with minimal changes required\footnote{\href{https://huggingface.co/docs/transformers/model_doc/auto}{https://huggingface.co/docs/transformers/model\_doc/auto}}.

\subsection{Internal Details of Models}

The internal details of open-weight models are described in this section. The internal information about the base language model and base image processor is available in Table \ref{tab:internal-model-details}. To train the Idefics models, the authors developed their own dataset - Obelics \cite{idefics}. This dataset is based on multimodal web documents, so a single sample of this dataset contains multiple images, making these models suitable for multimodal reference game runs.

%% file: latex/reference/appendix.tex
\subsection{Prompt Templates}

The prompt template for both players of the Reference Game is given in Figure~\ref{fig:reference_prompt_templates}.

\begin{figure*}
  \centering
  \begin{subfigure}[b]{0.48\textwidth}
    \centering
    \begin{prompt}

You are given three images, one is called target and the other two are distractors.
Your task is to generate a referring expression that best describes the target image while distinguishing it from the two other distractor images.
The first image is <IMAGE\_POSITION>, the second image is <IMAGE\_POSITION>, and the third image is <IMAGE\_POSITION>. \\

Instruction: Describe the target image. Generate the referring expression starting with the tag "Expression: " for the given target image. Omit any other text.\\

Target image: <IMAGE\_PATH>\\
Second image: <IMAGE\_PATH>\\
Third image: <IMAGE\_PATH>\\

\end{prompt}
\vspace*{-2ex}
\begin{prompt}
Expression: \$EXPRESSION\$
\end{prompt}

\caption{Prompt template for Player A (Instruction Giver) in the Reference Game.}
  \label{fig:reference_player_a}
  \end{subfigure}
  \hfill
  \begin{subfigure}[b]{0.48\textwidth}
    \centering
    \begin{prompt}

You are given three images. You are also given a referring expression that describes one of the given images.
Your task is to select the image that matches the given referring expression.
Generate only the number (in text) of the image that matches the given expression by selecting first, second, or third.\\

TARGET\_EXPRESSION\\
Question: Which image does the expression refer to?
Start with the tag "Answer: ", followed by your selection. Omit any other text.\\

First image: <IMAGE\_PATH>\\
Second image: <IMAGE\_PATH>\\
Third image: <IMAGE\_PATH>\\
\end{prompt}

\vspace*{-2ex}
\begin{prompt}
Answer: \$ANSWER\$
\end{prompt}

\caption{Prompt template for Player B (Instruction Follower) in the Reference Game}
\label{fig:reference_player_b}
  \end{subfigure}
  \caption{Reference game prompt templates for both players}
  \label{fig:reference_prompt_templates}
\end{figure*}

\subsection{Overall Results}
Table~\ref{tab:reference_results} displays the detailed results for the different Reference Game experiments.

\begin{table*}[th]
\footnotesize
\centering
\begin{tabular}{lccccccclll}
\hline
\multicolumn{11}{|c|}{\textbf{Text-only Reference Game}}  \\ \hline
\multicolumn{1}{|l|}{\textbf{Models}}   & \multicolumn{1}{c|}{\textbf{Row}} & \multicolumn{1}{c|}{\textbf{Column}} & \multicolumn{1}{c|}{\textbf{Diagonal}} & \multicolumn{1}{c|}{\textbf{Letter}} & \multicolumn{1}{c|}{\textbf{Shape}} & \multicolumn{1}{c|}{\textbf{Random}} & \multicolumn{4}{c|}{} \\ \hline

\multicolumn{1}{|l|}{InternVL2-26B} & \multicolumn{1}{c|}{53.3}& \multicolumn{1}{c|}{46.7}& \multicolumn{1}{c|}{56.7}& \multicolumn{1}{c|}{30.0}& \multicolumn{1}{c|}{40.0}& \multicolumn{1}{c|}{36.7} & \multicolumn{4}{c|}{}\\ \hline
\multicolumn{1}{|l|}{InternVL2-40B} & \multicolumn{1}{c|}{56.7}& \multicolumn{1}{c|}{33.3}& \multicolumn{1}{c|}{60.0}& \multicolumn{1}{c|}{46.7}& \multicolumn{1}{c|}{40.0}& \multicolumn{1}{c|}{33.3} & \multicolumn{4}{c|}{}\\ \hline
\multicolumn{1}{|l|}{InternVL2-76B} & \multicolumn{1}{c|}{76.7}& \multicolumn{1}{c|}{86.7}& \multicolumn{1}{c|}{43.3}& \multicolumn{1}{c|}{50.0}& \multicolumn{1}{c|}{60.0}& \multicolumn{1}{c|}{50.0} & \multicolumn{4}{c|}{}\\ \hline
\multicolumn{1}{|l|}{Phi-3-vision} & \multicolumn{1}{c|}{40.0}& \multicolumn{1}{c|}{30.0}& \multicolumn{1}{c|}{26.7}& \multicolumn{1}{c|}{20.0}& \multicolumn{1}{c|}{33.3}& \multicolumn{1}{c|}{36.7} & \multicolumn{4}{c|}{}\\ \hline
\multicolumn{1}{|l|}{Phi-3.5-vision} & \multicolumn{1}{c|}{40.0}& \multicolumn{1}{c|}{10.0}& \multicolumn{1}{c|}{46.7}& \multicolumn{1}{c|}{36.7}& \multicolumn{1}{c|}{53.3}& \multicolumn{1}{c|}{26.7} & \multicolumn{4}{c|}{}\\ \hline

\multicolumn{1}{|l|}{Pixtral-12B}& \multicolumn{1}{c|}{30.0}& \multicolumn{1}{c|}{36.7}& \multicolumn{1}{c|}{66.7}& \multicolumn{1}{c|}{43.3}& \multicolumn{1}{c|}{43.3}& \multicolumn{1}{c|}{26.7} & \multicolumn{4}{c|}{}\\ \hline

\multicolumn{1}{|l|}{Claude-3-5} & \multicolumn{1}{c|}{\textbf{100.0}}& \multicolumn{1}{c|}{\textbf{100.0}}& \multicolumn{1}{c|}{\textbf{93.3}}& \multicolumn{1}{c|}{\textbf{86.7}}& \multicolumn{1}{c|}{90.0}& \multicolumn{1}{c|}{76.7} & \multicolumn{4}{c|}{}\\ \hline
\multicolumn{1}{|l|}{Claude-3-opus} & \multicolumn{1}{c|}{23.3}& \multicolumn{1}{c|}{26.7}& \multicolumn{1}{c|}{33.3}& \multicolumn{1}{c|}{36.7}& \multicolumn{1}{c|}{33.3}& \multicolumn{1}{c|}{23.3} & \multicolumn{4}{c|}{}\\ \hline
\multicolumn{1}{|l|}{Gemini-1.5} & \multicolumn{1}{c|}{76.7}& \multicolumn{1}{c|}{76.7}& \multicolumn{1}{c|}{63.3}& \multicolumn{1}{c|}{43.3}& \multicolumn{1}{c|}{60.0}& \multicolumn{1}{c|}{46.7} & \multicolumn{4}{c|}{}\\ \hline
\multicolumn{1}{|l|}{GPT-4-1106} & \multicolumn{1}{c|}{26.7}& \multicolumn{1}{c|}{33.3}& \multicolumn{1}{c|}{33.3}& \multicolumn{1}{c|}{30.0}& \multicolumn{1}{c|}{33.3}& \multicolumn{1}{c|}{20.0} & \multicolumn{4}{c|}{}\\ \hline
\multicolumn{1}{|l|}{GPT-4o (May)} & \multicolumn{1}{c|}{96.7}& \multicolumn{1}{c|}{\textbf{100.0}}& \multicolumn{1}{c|}{90.0}& \multicolumn{1}{c|}{70.0}& \multicolumn{1}{c|}{\textbf{93.3}}& \multicolumn{1}{c|}{\textbf{90.0}} & \multicolumn{4}{c|}{}\\ \hline
\multicolumn{1}{|l|}{GPT-4o (Aug)} & \multicolumn{1}{c|}{90.0}& \multicolumn{1}{c|}{\textbf{100.0}}& \multicolumn{1}{c|}{\textbf{93.3}}& \multicolumn{1}{c|}{76.7}& \multicolumn{1}{c|}{80.0}& \multicolumn{1}{c|}{86.7} & \multicolumn{4}{c|}{}\\ \hline
\multicolumn{1}{|l|}{GPT-4o-mini} & \multicolumn{1}{c|}{76.7}& \multicolumn{1}{c|}{90.0}& \multicolumn{1}{c|}{80.0}& \multicolumn{1}{c|}{63.3}& \multicolumn{1}{c|}{73.3}& \multicolumn{1}{c|}{56.7} & \multicolumn{4}{c|}{}\\ \hline
\multicolumn{1}{|l|}{Idefics-80B} & \multicolumn{1}{c|}{33.3}& \multicolumn{1}{c|}{33.3}& \multicolumn{1}{c|}{33.3}& \multicolumn{1}{c|}{26.7}& \multicolumn{1}{c|}{23.3}& \multicolumn{1}{c|}{36.7} & \multicolumn{4}{c|}{}\\ \hline
\multicolumn{1}{|l|}{Idefics-9B} & \multicolumn{1}{c|}{16.7}& \multicolumn{1}{c|}{0.0}& \multicolumn{1}{c|}{0.0}& \multicolumn{1}{c|}{0.0}& \multicolumn{1}{c|}{6.7}& \multicolumn{1}{c|}{0.0} & \multicolumn{4}{c|}{}\\ \hline
\multicolumn{1}{|l|}{InternLM-XC} & \multicolumn{1}{c|}{0.0}& \multicolumn{1}{c|}{0.0}& \multicolumn{1}{c|}{0.0}& \multicolumn{1}{c|}{0.0}& \multicolumn{1}{c|}{0.0}& \multicolumn{1}{c|}{0.0} & \multicolumn{4}{c|}{}\\ \hline

\multicolumn{11}{c}{\textbf{Multimodal Reference Game}}  

\\ \hline
\multicolumn{1}{|l|}{}                  & \multicolumn{1}{c|}{\textbf{Row}} & \multicolumn{1}{c|}{\textbf{Column}} & \multicolumn{1}{c|}{\textbf{Diagonal}} & \multicolumn{1}{c|}{\textbf{Letter}} & \multicolumn{1}{c|}{\textbf{Shape}} & \multicolumn{1}{c|}{\textbf{Random}} & \multicolumn{1}{c|}{\textbf{ADE}} & \multicolumn{1}{c|}{\textbf{DOCCI}} & \multicolumn{1}{c|}{\textbf{CLEVR}} & \multicolumn{1}{c|}{\textbf{Pent.}} \\ \hline

\multicolumn{1}{|l|}{InternVL2-26B} & \multicolumn{1}{c|}{33.3}& \multicolumn{1}{c|}{33.3}& \multicolumn{1}{c|}{40.0}& \multicolumn{1}{c|}{33.3}& \multicolumn{1}{c|}{33.3}& \multicolumn{1}{c|}{33.3}& \multicolumn{1}{c|}{33.3}& \multicolumn{1}{c|}{30.0}& \multicolumn{1}{c|}{3.3}& \multicolumn{1}{c|}{33.3}\\ \hline
\multicolumn{1}{|l|}{InternVL2-40B} & \multicolumn{1}{c|}{30.0}& \multicolumn{1}{c|}{26.7}& \multicolumn{1}{c|}{36.7}& \multicolumn{1}{c|}{30.0}& \multicolumn{1}{c|}{40.0}& \multicolumn{1}{c|}{30.0}& \multicolumn{1}{c|}{33.3}& \multicolumn{1}{c|}{30.0}& \multicolumn{1}{c|}{60.0}& \multicolumn{1}{c|}{\textbf{43.3}}\\ \hline
\multicolumn{1}{|l|}{InternVL2-76B} & \multicolumn{1}{c|}{33.3}& \multicolumn{1}{c|}{30.0}& \multicolumn{1}{c|}{33.3}& \multicolumn{1}{c|}{30.0}& \multicolumn{1}{c|}{33.3}& \multicolumn{1}{c|}{30.0}& \multicolumn{1}{c|}{40.0}& \multicolumn{1}{c|}{30.0}& \multicolumn{1}{c|}{33.3}& \multicolumn{1}{c|}{20.0}\\ \hline
\multicolumn{1}{|l|}{Phi-3-vision} & \multicolumn{1}{c|}{0.0}& \multicolumn{1}{c|}{0.0}& \multicolumn{1}{c|}{0.0}& \multicolumn{1}{c|}{0.0}& \multicolumn{1}{c|}{0.0}& \multicolumn{1}{c|}{0.0}& \multicolumn{1}{c|}{33.3}& \multicolumn{1}{c|}{33.3}& \multicolumn{1}{c|}{33.3}& \multicolumn{1}{c|}{33.3}\\ \hline
\multicolumn{1}{|l|}{Phi-3.5-vision} & \multicolumn{1}{c|}{33.3}& \multicolumn{1}{c|}{33.3}& \multicolumn{1}{c|}{33.3}& \multicolumn{1}{c|}{33.3}& \multicolumn{1}{c|}{33.3}& \multicolumn{1}{c|}{33.3}& \multicolumn{1}{c|}{0.0}& \multicolumn{1}{c|}{0.0}& \multicolumn{1}{c|}{0.0}& \multicolumn{1}{c|}{0.0}\\ \hline
\multicolumn{1}{|l|}{Pixtral-12B}& \multicolumn{1}{c|}{30.0}& \multicolumn{1}{c|}{43.3}& \multicolumn{1}{c|}{26.7}& \multicolumn{1}{c|}{43.3}& \multicolumn{1}{c|}{23.3}& \multicolumn{1}{c|}{30.0}& \multicolumn{1}{c|}{73.3}& \multicolumn{1}{c|}{3.3}& \multicolumn{1}{c|}{63.3}& \multicolumn{1}{c|}{0.0}\\ \hline
\multicolumn{1}{|l|}{Claude-3-5} & \multicolumn{1}{c|}{\textbf{76.7}}& \multicolumn{1}{c|}{\textbf{90.0}}& \multicolumn{1}{c|}{\textbf{93.3}}& \multicolumn{1}{c|}{56.7}& \multicolumn{1}{c|}{\textbf{80.0}}& \multicolumn{1}{c|}{\textbf{76.7}}& \multicolumn{1}{c|}{83.3}& \multicolumn{1}{c|}{96.7}& \multicolumn{1}{c|}{\textbf{100.0}}& \multicolumn{1}{c|}{33.3}\\ \hline
\multicolumn{1}{|l|}{Claude-3-opus} & \multicolumn{1}{c|}{63.3}& \multicolumn{1}{c|}{36.7}& \multicolumn{1}{c|}{60.0}& \multicolumn{1}{c|}{46.7}& \multicolumn{1}{c|}{33.3}& \multicolumn{1}{c|}{40.0}& \multicolumn{1}{c|}{50.0}& \multicolumn{1}{c|}{70.0}& \multicolumn{1}{c|}{66.7}& \multicolumn{1}{c|}{30.0}\\ \hline
\multicolumn{1}{|l|}{Gemini-1.5} & \multicolumn{1}{c|}{56.7}& \multicolumn{1}{c|}{53.3}& \multicolumn{1}{c|}{56.7}& \multicolumn{1}{c|}{30.0}& \multicolumn{1}{c|}{43.3}& \multicolumn{1}{c|}{36.7}& \multicolumn{1}{c|}{40.0}& \multicolumn{1}{c|}{53.3}& \multicolumn{1}{c|}{33.3}& \multicolumn{1}{c|}{30.0}\\ \hline
\multicolumn{1}{|l|}{GPT-4-1106} & \multicolumn{1}{c|}{43.3}& \multicolumn{1}{c|}{43.3}& \multicolumn{1}{c|}{66.7}& \multicolumn{1}{c|}{33.3}& \multicolumn{1}{c|}{46.7}& \multicolumn{1}{c|}{40.0}& \multicolumn{1}{c|}{93.3}& \multicolumn{1}{c|}{96.7}& \multicolumn{1}{c|}{90.0}& \multicolumn{1}{c|}{33.3}\\ \hline
\multicolumn{1}{|l|}{GPT-4o (May)} & \multicolumn{1}{c|}{56.7}& \multicolumn{1}{c|}{63.3}& \multicolumn{1}{c|}{76.7}& \multicolumn{1}{c|}{\textbf{63.3}}& \multicolumn{1}{c|}{50.0}& \multicolumn{1}{c|}{50.0}& \multicolumn{1}{c|}{\textbf{100.0}}& \multicolumn{1}{c|}{\textbf{100.0}}& \multicolumn{1}{c|}{\textbf{100.0}}& \multicolumn{1}{c|}{30.0}\\ \hline
\multicolumn{1}{|l|}{GPT-4o (Aug)} & \multicolumn{1}{c|}{63.3}& \multicolumn{1}{c|}{53.3}& \multicolumn{1}{c|}{76.7}& \multicolumn{1}{c|}{60.0}& \multicolumn{1}{c|}{53.3}& \multicolumn{1}{c|}{46.7}& \multicolumn{1}{c|}{\textbf{100.0}}& \multicolumn{1}{c|}{\textbf{100.0}}& \multicolumn{1}{c|}{\textbf{100.0}}& \multicolumn{1}{c|}{26.7}\\ \hline
\multicolumn{1}{|l|}{GPT-4o-mini} & \multicolumn{1}{c|}{46.7}& \multicolumn{1}{c|}{23.3}& \multicolumn{1}{c|}{60.0}& \multicolumn{1}{c|}{33.3}& \multicolumn{1}{c|}{36.7}& \multicolumn{1}{c|}{40.0}& \multicolumn{1}{c|}{56.7}& \multicolumn{1}{c|}{43.3}& \multicolumn{1}{c|}{83.3}& \multicolumn{1}{c|}{23.3}\\ \hline
\multicolumn{1}{|l|}{Idefics-80B} & \multicolumn{1}{c|}{36.7}& \multicolumn{1}{c|}{33.3}& \multicolumn{1}{c|}{33.3}& \multicolumn{1}{c|}{33.3}& \multicolumn{1}{c|}{33.3}& \multicolumn{1}{c|}{33.3}& \multicolumn{1}{c|}{33.3}& \multicolumn{1}{c|}{33.3}& \multicolumn{1}{c|}{33.3}& \multicolumn{1}{c|}{36.7}\\ \hline
\multicolumn{1}{|l|}{Idefics-9B} & \multicolumn{1}{c|}{30.0}& \multicolumn{1}{c|}{30.0}& \multicolumn{1}{c|}{33.3}& \multicolumn{1}{c|}{33.3}& \multicolumn{1}{c|}{33.3}& \multicolumn{1}{c|}{33.3}& \multicolumn{1}{c|}{26.7}& \multicolumn{1}{c|}{20.0}& \multicolumn{1}{c|}{33.3}& \multicolumn{1}{c|}{13.3}\\ \hline
\multicolumn{1}{|l|}{InternLM-XC} & \multicolumn{1}{c|}{0.0}& \multicolumn{1}{c|}{0.0}& \multicolumn{1}{c|}{0.0}& \multicolumn{1}{c|}{0.0}& \multicolumn{1}{c|}{0.0}& \multicolumn{1}{c|}{0.0}& \multicolumn{1}{c|}{6.7}& \multicolumn{1}{c|}{3.3}& \multicolumn{1}{c|}{0.0}& \multicolumn{1}{c|}{0.0}\\ \hline

\end{tabular}
\caption{Average success scores across experiments for textual and multimodal reference games. Random performance is at 33.3 (as there are three images to choose from). Note: Only the grid experiments are used in the text-only version, as the others are actual images and cannot easily be represented in ASCII format. \textit{N/A}: not available means that the experiment was not played due to the model not following the formal instructions. The best results for each experiment are highlighted in bold. \textbf{Pent}: Pentomino pieces}
\label{tab:reference_results}
\end{table*}

\subsection{Qualitative Samples}
\begin{figure*}
    \centering
    \includegraphics[width=1.0\textwidth]{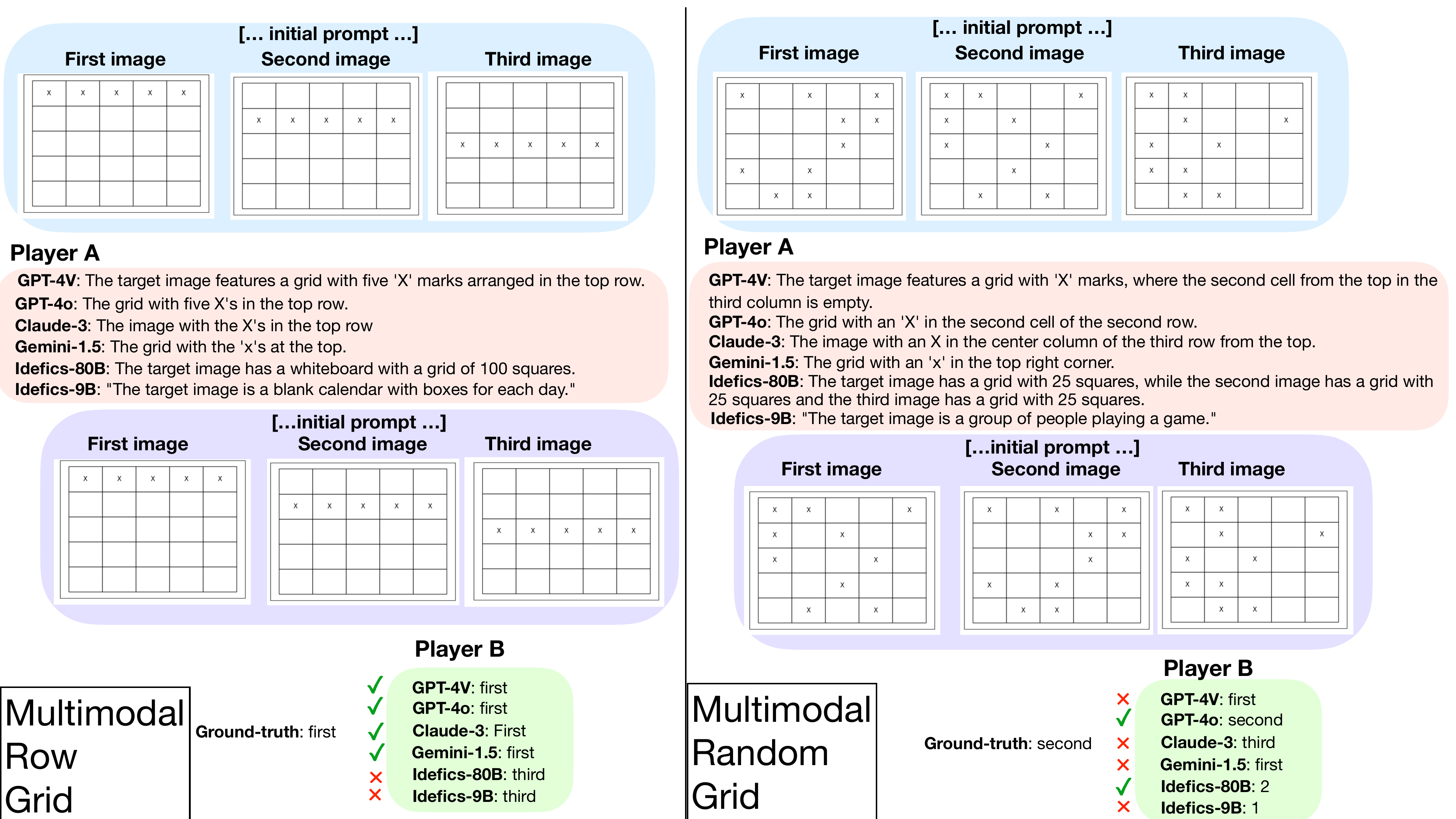}
    \caption{Sample outputs generated by the models for multimodal row and random grid experiments}
    \label{fig:ref_multimodal_grid}
\end{figure*}

\begin{figure*}
    \centering
    \includegraphics[width=1.0\textwidth]{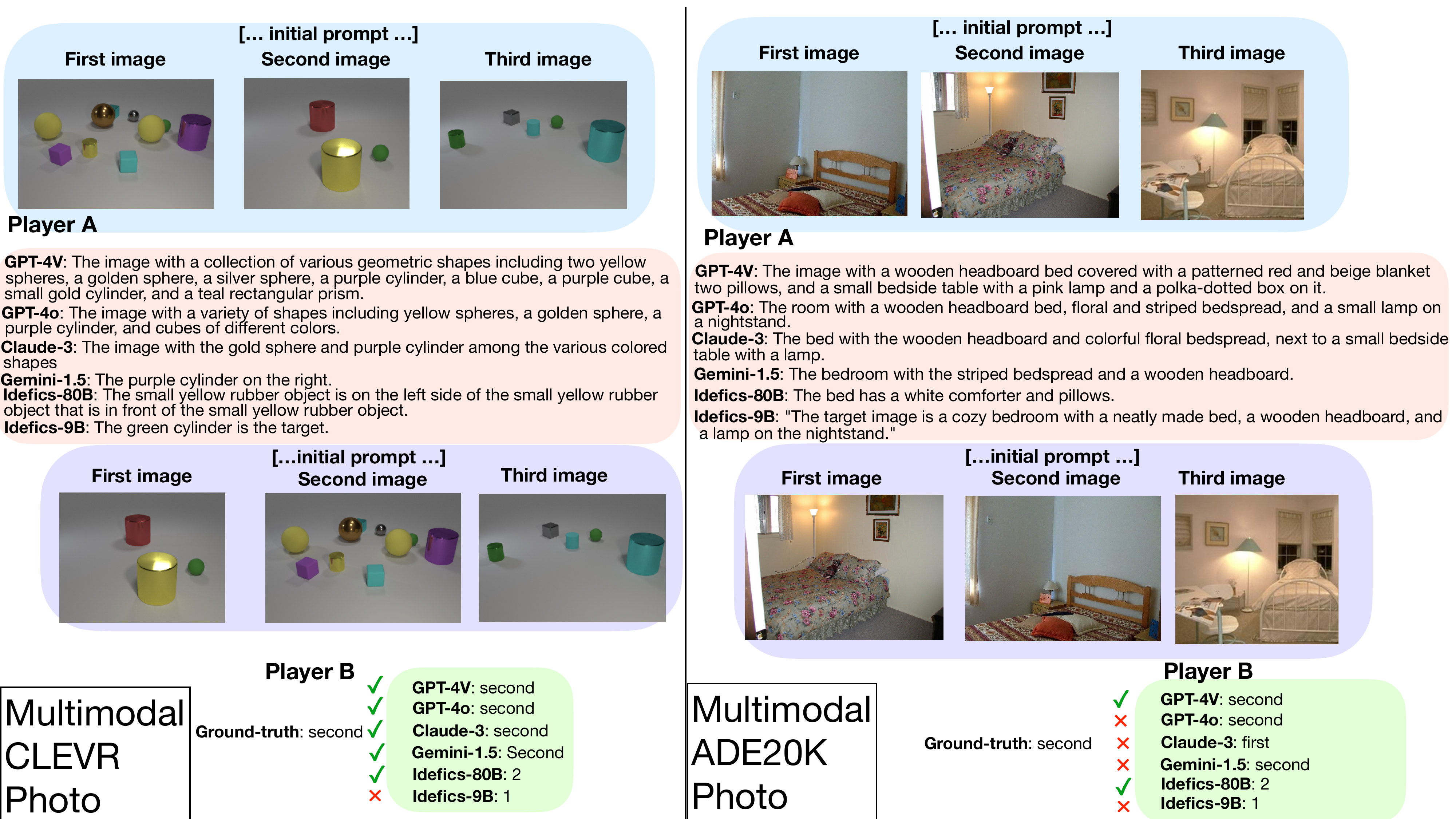}
    \caption{Sample outputs generated by the models for the CLEVR and ADE experiments.}
    \label{fig:ref_clevr_aed}
\end{figure*}

\begin{figure*}
    \centering
    \includegraphics[width=1.0\textwidth]{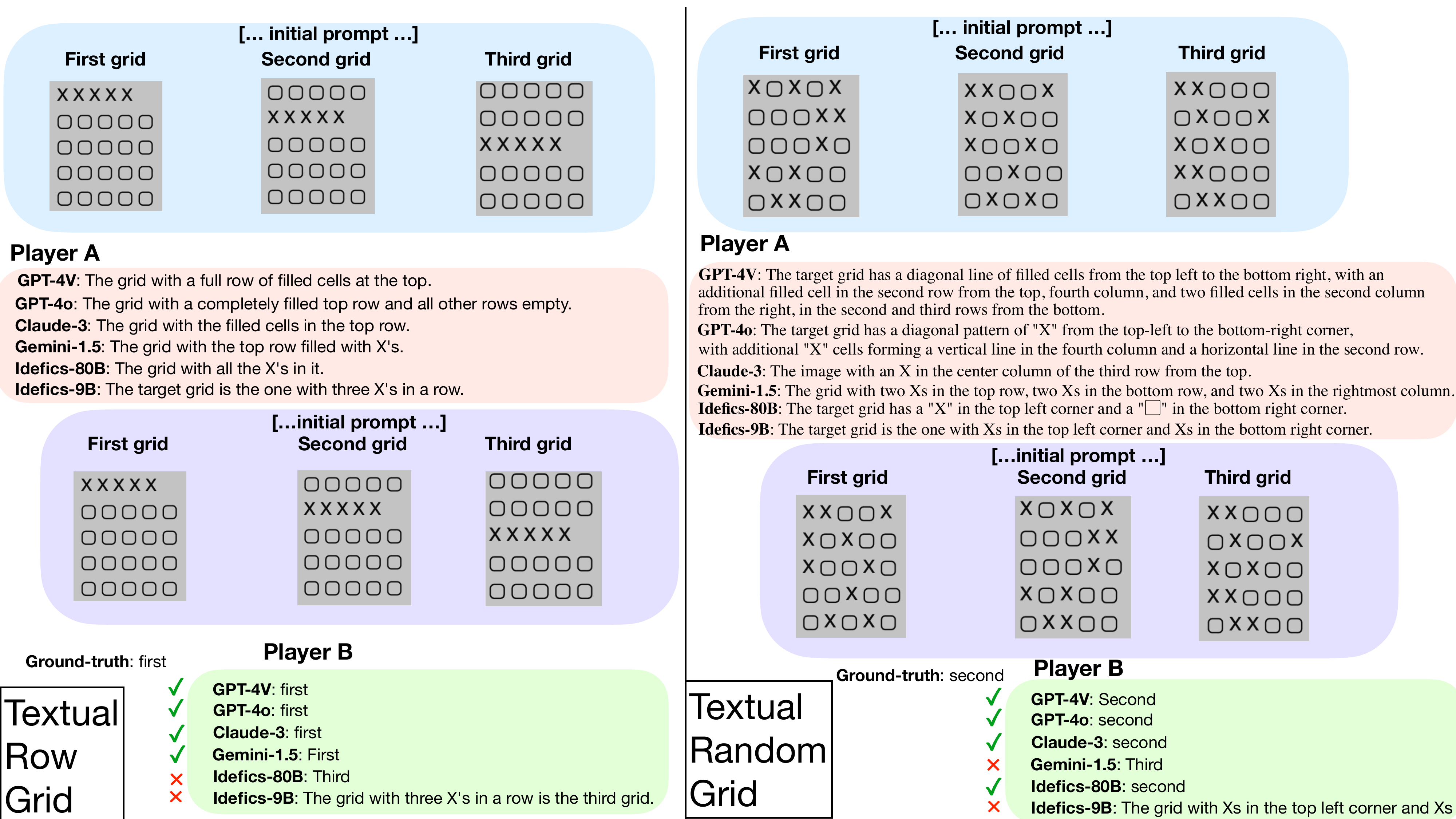}
    \caption{Sample outputs generated by the models for textual row and random grid experiments.}
    \label{fig:ref_textual_grid}
\end{figure*}

Here, we provide example instances and the respective model responses for the pentomino experiments (Figure~\ref{fig:ref_pentomino}), the CLEVR and ADE experiments (Figure~\ref{fig:ref_clevr_aed}), and the multimodal (Figure~\ref{fig:ref_multimodal_grid}) and the textual (Figure~\ref{fig:ref_textual_grid}) grid experiments.

\subsection{Static Target Image}
\label{app:ref_static}
In order to understand the effect of the distractors on the generated expression, we created another set of instances from the ADE, CLEVR, and DOCCI datasets where the target image is kept the same for all instances, and the distractors are chosen from similar images in the respective datasets. The distractors from the ADE dataset were chosen from the same scene category ``bedroom''. For the DOCCI instances, we used the concept category ``dog'' to select distractor images from the same category. The instances from the CLEVR dataset were chosen based on the object annotations (large metal green sphere, small rubber blue cube, etc.). The images with the most objects in common with the target were taken as distractors. 

The results are given in Table~\ref{tab:reference_static_target}. We can observe that both GPT-4 models are not susceptible to the change in instances while being the best performing models for all datasets. Comparing the \textit{Claude-3} results in Table~\ref{tab:reference_results} and Table~\ref{tab:reference_static_target}, we can observe that the performance drops for DOCCI and ADE instances with static target.

We were initially hoping that we can automatically test for context-sensitivity (and hence, the output really being a referring expression more than a description) by pairing the target image in different distractors and seeing whether the generation changes. However, it turns out that the observed context sensitivity is of the wrong kind, as the generated expression also changes if the target image and the same distractors simply are presented in different orders. Hence, we have no grounds to assume that the generated expressions are more than very detailed descriptions.

\begin{table}[ht]
\footnotesize
\centering

\begin{tabular}{|l|l|l|l|}
\hline
Model    & \textbf{ADE}         & \textbf{DOCCI}           & \textbf{CLEVR}         \\ \hline
\multicolumn{1}{|l|}{InternVL2-26B} & \multicolumn{1}{c|}{36.7}& \multicolumn{1}{c|}{36.7}& \multicolumn{1}{c|}{0.0}\\ \hline
\multicolumn{1}{|l|}{InternVL2-40B} & \multicolumn{1}{c|}{56.7}& \multicolumn{1}{c|}{16.7}& \multicolumn{1}{c|}{36.7}\\ \hline
\multicolumn{1}{|l|}{InternVL2-76B} & \multicolumn{1}{c|}{66.7}& \multicolumn{1}{c|}{33.3}& \multicolumn{1}{c|}{33.3}\\ \hline
\multicolumn{1}{|l|}{Phi-3-vision} & \multicolumn{1}{c|}{33.3}& \multicolumn{1}{c|}{33.3}& \multicolumn{1}{c|}{33.3}\\ \hline
\multicolumn{1}{|l|}{Phi-3.5-vision} & \multicolumn{1}{c|}{0.0}& \multicolumn{1}{c|}{0.0}& \multicolumn{1}{c|}{0.0}\\ \hline

\multicolumn{1}{|l|}{Pixtral-12B} & \multicolumn{1}{c|}{73.3} & \multicolumn{1}{c|}{0.0} & \multicolumn{1}{c|}{50.0}\\ \hline

\multicolumn{1}{|l|}{Claude-3-5} & \multicolumn{1}{c|}{80.0}& \multicolumn{1}{c|}{86.7}& \multicolumn{1}{c|}{\textbf{100.0}}\\ \hline
\multicolumn{1}{|l|}{Claude-3-opus} & \multicolumn{1}{c|}{30.0}& \multicolumn{1}{c|}{20.0}& \multicolumn{1}{c|}{66.7}\\ \hline
\multicolumn{1}{|l|}{Gemini-1.5-flash} & \multicolumn{1}{c|}{23.3}& \multicolumn{1}{c|}{36.7}& \multicolumn{1}{c|}{46.7}\\ \hline
\multicolumn{1}{|l|}{GPT-4-1106} & \multicolumn{1}{c|}{\textbf{100.0}}& \multicolumn{1}{c|}{\textbf{100.0}}& \multicolumn{1}{c|}{93.3}\\ \hline
\multicolumn{1}{|l|}{GPT-4o (May)} & \multicolumn{1}{c|}{\textbf{100.0}}& \multicolumn{1}{c|}{\textbf{100.0}}& \multicolumn{1}{c|}{90.0}\\ \hline
\multicolumn{1}{|l|}{GPT-4o (Aug)} & \multicolumn{1}{c|}{\textbf{100.0}}& \multicolumn{1}{c|}{\textbf{100.0}}& \multicolumn{1}{c|}{93.3}\\ \hline
\multicolumn{1}{|l|}{GPT-4o-mini} & \multicolumn{1}{c|}{50.0}& \multicolumn{1}{c|}{36.7}& \multicolumn{1}{c|}{80.0}\\ \hline
\multicolumn{1}{|l|}{Idefics-80B} & \multicolumn{1}{c|}{33.3}& \multicolumn{1}{c|}{33.3}& \multicolumn{1}{c|}{30.0}\\ \hline
\multicolumn{1}{|l|}{Idefics-9B} & \multicolumn{1}{c|}{30.0}& \multicolumn{1}{c|}{16.7}& \multicolumn{1}{c|}{33.3}\\ \hline
\multicolumn{1}{|l|}{InternLM-XC} & \multicolumn{1}{c|}{13.3}& \multicolumn{1}{c|}{6.7}& \multicolumn{1}{c|}{0.0}\\ \hline

\end{tabular}
\caption{Average success scores across ADE, DOCCI and CLEVR instances where the target image is kept the same for all instances and the set of distractors varies.}
\label{tab:reference_static_target}
\end{table}

%% file: latex/matchit/3_appendix.tex
\subsection{Text-only (ASCII) version}

The base grids used for the text based ASCII variant of MatchIt consist of the \textit{diagonal, letter} and \textit{shape} grids used in the Reference game. Besides pairs of same and different grids, we used two sets of instances with similar grid pairs. One set of instances was created by either mirroring the grids vertically or horizontally or by turning them 90 degrees and another set of instances consisted of grids paired with new grids with edit distance of two, where the symbol was inverted at two random positions of the grid. Figure~\ref{fig:matchit_grids_examples} shows the modifications for similar grid pairs.
For each of the difficulties (same, similar with transformed motive, similar with edit distance of two and different grids), 10 instances were part of the final game play, for a total of 40 instances.

\input{latex/matchit/figures/matchit_grids_examples}

\subsection{Experimental Setup}

\textbf{Game rules}
Importantly, each utterance has to start with the right flag such as `DESCRIPTION', `QUESTION', `ANSWER' and `DECISION' or the game will be aborted. Also, explanations for final decisions are not allowed. An example of schematic game play is shown in Figure~\ref{fig:intro_example}. 

\subsection{Prompts}
All prompt templates used for the game are displayed in Figure~\ref{fig:matchit_prompts}
\input{latex/matchit/figures/prompts}

\subsection{Instances}

Three groups of instance types were used for MatchIt: photographs, images of pentomino boards and grids of ASCII grids. Those instances were grouped into three (to four) difficulties; both players get the same image, both players get similar images (two different types for ASCII grids) or completely different images. The process of producing similar and different image pairs is described below.
The curation rationale for a similar picture was that both pictures could be described with the same (short) sentence, but their difference should be striking enough that also one (short) sentence should be enough.
All images for this multimodal variant are taken from the Visual Genome Dataset \cite{krishna2017visgenome}, which has rich annotations for every object in the image including attributes and relations to other objects. For each sampled image pair, the Jaccard index between all object labels and their respective attributes for each image was calculated. The lowest scoring pairs up to a threshold of 0.05 were chosen as ``different'' pairs. This ensures that there are no to very little shared semantic contents between the pictures. In order to get similar pairs, the cosine similarity of the image embeddings of the CLIP model \cite{radford2021clip} of image pairs with a Jaccard index above 0.22 was calculated and pairs with a cosine similarity above 0.8 chosen. Both thresholds were chosen considering quality of pairs as well as sufficient numbers of pairs. From those, final pairs were selected manually following the curation rationale mentioned above filtering out instances that were not enough or too similar.

\subsection{Results}

The average main scores depending on difficulty are presented in Tables \ref{tab:difficulties_matchit_pen} and \ref{tab:difficulties_matchit_ascii}. 

\input{latex/matchit/overview_difficulties}

\subsection{Detailed Analysis}

\paragraph{Model type}
There is a big jump in performance from the open source models to commercial models with the GPT and Claude models having slightly higher played and main scores than Gemini-1.5. Possible reasons for this difference are carried out in the following paragraphs.

\paragraph{Modality}
While the initial photo descriptions are mostly correct, the grid and pentomino board descriptions often are incorrect and/or incomplete. 
As seen for the reference game, the overall performance with pentomino board instances is worse than with photos, with open models sometimes not even remotely describing a picture of abstract shapes and rather trying to find a concrete meaning where there is none (e.g. describing a pentomino board as people playing soccer, people playing tic-tac-toe or as pixel art of a tree).
There are two forms of grid descriptions, either in form of a row-by-row (or by column or borders) description of the placements of X’s and squares or by repeating the string of characters verbatim - a behavior we did not predict and therefore not prohibit. No instance was found where a general motive was described instead of the single parts of the grid. 
This is interesting since other games using grids have shown that correct grid descriptions can be made. The wording of of the prompts was adopted completely from the multimodal variant except for necessary exchanges such as \texttt{image $\leftrightarrow$ grid}, so a more tailored prompt introducing the grids as such could have maybe mitigated that.
Comparing the main scores of the different modalities, for multimodal game play, larger models have an apparent advantage over their smaller counterparts that disappears when no visual input is given.

\paragraph{Instance difficulties}

There are significant differences between levels of difficulty. Similar images (for both modalities) get classified wrong most times, showing that the level of detail of the asked questions is not high enough to determine that there are differences in the images. 
The two types of similarity of the grid inputs do not seem to differ regarding the performance although the kinds of similarities between grids are quite distinct.
\begin{figure}
    \centering
    \includegraphics[width = \columnwidth]{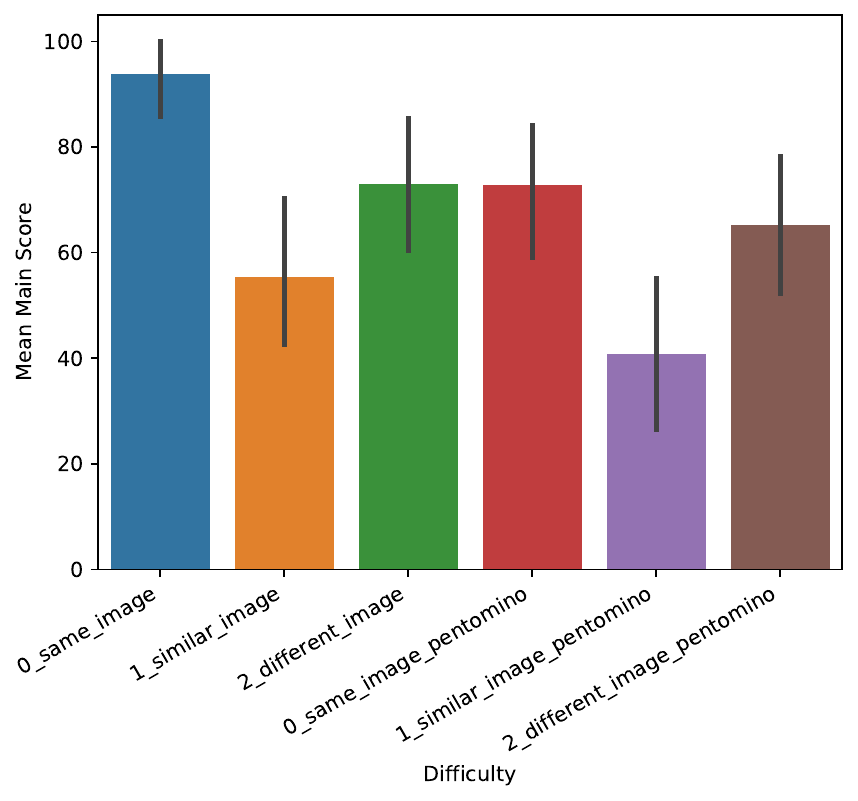}
    \caption{Mean Main Scores for each of the instance difficulties for the multimodal MatchIt version}
    \label{fig:difficulties_multimodal_barplot}
\end{figure}

\begin{figure}
    \centering
    \includegraphics[width = \columnwidth]{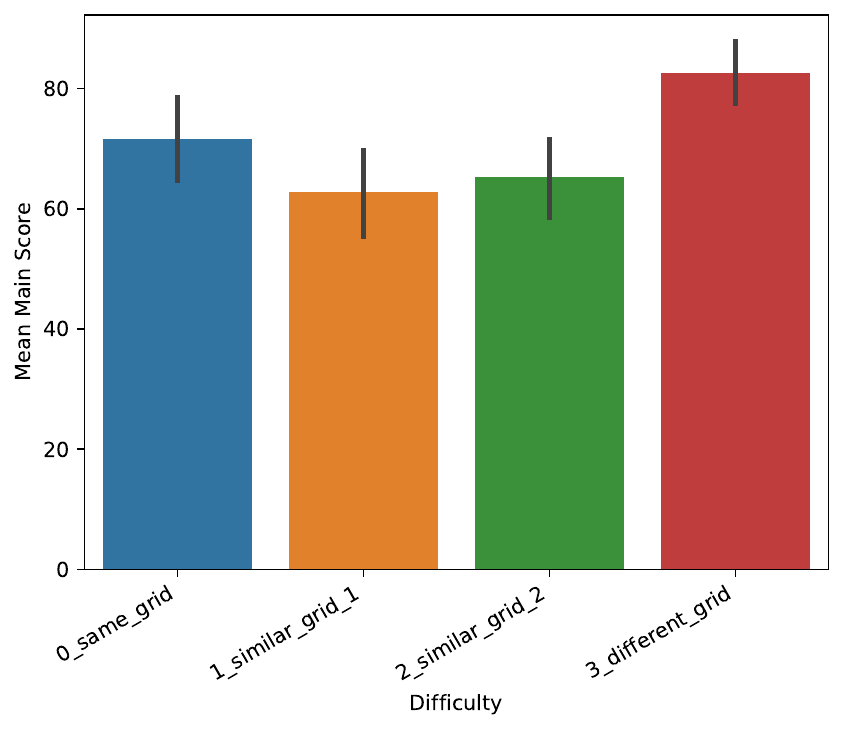}
    \caption{Mean Main Scores for the instance difficulties of ASCII version of MatchIt}
    \label{fig:difficulties_ascii_barplot}
\end{figure}

\paragraph{General game play}
Following are a few observations about the general game play based on a sample of played instances for each model and difficulty level.
A large difference between smaller open models and larger as well as commercial modes are the types of questions asked. The dialogue of the latter contains more open questions that even includes some details about the player's own image and leaves the other player room to elaborate whereas smaller models usually produced simple yes/no questions. Similar to that, some answers of the larger and commercial models are not simply a yes/no answer but explanations and comparisons to the known information. This is observable in the contrast between Table~\ref{tab:matchit_example_good_alt} and Table~\ref{tab:matchit_example_bad}, displaying actual game play of the same instance by a commercial and an open model.

\input{latex/matchit/figures/matchit_example_good_alt}
\input{latex/matchit/figures/matchit_example_bad}

\normalsize

The inquired content of questions for photos is often about the existence of objects in the other player's pictures and if a description is approximately correct, similar behavior can be observed with pentomino boards. The questions about ASCII grids focus mostly on rows and columns than on whole motives. Multimodal models in particular ask about features of the grid that are not represented (such as shape, size or color of the characters). 
A frequent flaw in asked questions is asking about information that was already given, for example in the description and therefore not advance the exchange of information.
In general, while the photograph descriptions are overall correct, the answers to questions about the image are not always correct. Sometimes hallucinations occur, especially when a question is asked about the other image the respective player has no visual information about. This happens more frequently for pentomino board and grid instances than for photo instances, as latter are overall better understood by the models.
Another source of wrong answers is the lack of distinction between the own and the other image, shown when a question was posed about a player's own image or objects in the own image in comparison to the other. This leads to wrong answers and hallucinations with open models.

%% file: latex/matchit/figures/matchit_grids_examples.tex
\begin{figure}[ht!]
\centering
\small
\begin{tabular}{ccccc}
\multicolumn{5}{c}{Original Grid}\\
X&X&X&X&X\\
X&$\square$&$\square$&$\square$&$\square$\\
X&X&X&$\square$&$\square$\\
X&$\square$&$\square$&$\square$&$\square$\\
X&X&X&X&X
\end{tabular}
$\;$
\begin{tabular}{ccccc} 
\multicolumn{5}{c}{Similar Grid (1)}\\
  X&X&X&X&X\\
  $\square$&$\square$&$\square$&$\square$&X\\
  $\square$&$\square$&X&X&X\\
  $\square$&$\square$&$\square$&$\square$&X\\
  X&X&X&X&X
\end{tabular}
$\;$
\begin{tabular}{ccccc}
\multicolumn{5}{c}{Similar Grid (2)}\\
  X&X&X&X&X\\
  X&$\square$&X&$\square$&$\square$\\
  X&X&X&$\square$&$\square$\\
  X&$\square$&$\square$&$\square$&$\square$\\
  X&$\square$&X&X&X
\end{tabular}  
    \vspace*{-.5\baselineskip}
    \caption{Example grids for the \texttt{matchit\_ascii} version of the game}
    \label{fig:matchit_grids_examples}
    \vspace*{-1\baselineskip}
\end{figure}

%% file: latex/matchit/figures/prompts.tex
\begin{figure*}

\begin{subfigure}{\columnwidth}
    
\begin{prompt}
    You are participating in a collaborative guessing game.
    The goal is to find out whether this picture and another picture only I can see, are the same.
    Please describe your image first. Then, I will provide my description and we can ask each other questions about the images to figure out whether they are the same. Now start your short image description with ``DESCRIPTION:'' followed by the description. Do not add anything else.
\end{prompt}

\vspace*{-2ex}
\begin{prompt}
    DESCRIPTION: \$DESCRIPTION\$
\end{prompt}

\caption{Initial prompt.}
\label{fig:matchit_initial_prompt}

\end{subfigure}
\begin{subfigure}{\columnwidth}
\begin{prompt}
    Now ask a question in order to find out new aspects of my image that may be different to your image. Start with ``QUESTION:'' and do not add anything else.
\end{prompt}

\vspace*{-2ex}
\begin{prompt}
QUESTION: \$QUESTION\$
\end{prompt}

\caption{Eliciting questions from the players.}
\label{fig:matchit_question}

\end{subfigure}

\begin{subfigure}{\columnwidth}
\begin{prompt}
     Start your answer with ``ANSWER'' and do not add anything else.
\end{prompt}

\vspace*{-2ex}
\begin{prompt}
ANSWER: \$ANSWER\$
\end{prompt}

\caption{Eliciting answers from the players.}
\label{fig:matchit_answer}

\end{subfigure}

\begin{subfigure}{\columnwidth}
\begin{prompt}
    Now come to a decision. What do you think: are your picture and the other picture described the same picture?  Write ``DECISION: same images'' if you think they are the same picture or ``DECISION: different images'' if you think they are different pictures. Do not add anything else.
\end{prompt}

\vspace*{-2ex}
\begin{prompt}
DECISION: \$DECISION\$
\end{prompt}

\caption{Eliciting decisions from the players. }
\label{fig:matchit_decision}

\end{subfigure}
\caption{Prompt templates for MatchIt.}
\label{fig:matchit_prompts}
\end{figure*}

%% file: latex/matchit/overview_difficulties.tex
\begingroup
\setlength{\tabcolsep}{6pt}

\begin{table}[ht!]
\resizebox{\columnwidth}{!}{%
\begin{tabular}{lrrr}
\toprule
\multicolumn{4}{l}{\textbf{matchit (multimodal) - Photo input}} \\ \midrule
 &  0 &  1 &  2 \\
\midrule
InternVL2-26B & 90.0 & 90.0 & \textbf{100.0} \\
InternVL2-40B & \textbf{100.0} & 70.0 & \textbf{100.0} \\
InternVL2-76B & \textbf{100.0} & 80.0 & \textbf{100.0} \\
Phi-3-vision& - & - & - \\
Phi-3.5-vision & 0.0 & 0.0 & 0.0 \\
Pixtral-12B & 40.0 & 80.0 & \textbf{100.0} \\
Claude-3.5 & \textbf{100.0} & 90.0 & \textbf{100.0} \\
Claude-3 & \textbf{100.0} & 70.0 & \textbf{100.0} \\
Gemini-1.5 & \textbf{100.0} & \textbf{100.0} & \textbf{100.0} \\
GPT-4-1106 & 80.0 & 90.0 & \textbf{100.0} \\
GPT-4o (May) & 90.0 & \textbf{100.0} & \textbf{100.0} \\
GPT-4o (August) & \textbf{100.0} & \textbf{100.0} & \textbf{100.0} \\
GPT-4o-mini & 70.0 & \textbf{100.0} & \textbf{100.0} \\
Idefics-80B & 88.9 & 12.5 & 62.5 \\
Idefics-9B & \textbf{100.0} & 0.0 & 0.0 \\
InternLM-XC & \textbf{100.0} & 60.0 & \textbf{100.0} \\
\toprule
\multicolumn{4}{l}{\textbf{matchit (multimodal)- Pentomino input}} \\ \midrule
 &  0 &  1 &  2 \\
\midrule
InternVL2-26B & \textbf{100.0} & 80.0 & \textbf{100.0} \\
InternVL2-40B & \textbf{100.0} & 10.0 & \textbf{100.0} \\
InternVL2-76B & \textbf{100.0} & 60.0 & \textbf{100.0} \\
Phi-3-vision& - & - & -\\
Phi-3.5-vision & 0.0 & 0.0 & 0.0 \\
Pixtral-12B & 0.0 & 60.0 & \textbf{100.0} \\
Claude-3.5 & 90.0 & 30.0 & \textbf{100.0} \\
Claude-3 & 90.0 & 30.0 & \textbf{100.0} \\
Gemini-1.5 & 83.3 & 30.0 & \textbf{100.0} \\
GPT-4-1106 & 40.0 & 70.0 & \textbf{100.0} \\
GPT-4o-August & 88.9 & 0.0 & \textbf{100.0} \\
GPT-4o-May & 70.0 & 10.0 & \textbf{100.0} \\
GPT-4o-mini & 60.0 & \textbf{90.0} & \textbf{100.0} \\
Idefics-80B & 50.0 & 77.8 & 40.0 \\
Idefics-9B & \textbf{100.0} & 0.0 & 0.0 \\
InternLM-XC & 60.0 & 60.0 & 90.0 \\

\end{tabular}\\
}
\caption{Mean main score for each model, grouped by difficulty level. 0: same image/pentomino board, 1: similar image/pentomino board, 2: different image/pentomino board}
\label{tab:difficulties_matchit_pen}
\end{table}

\begin{table}[ht!]
\resizebox{\columnwidth}{!}{
\begin{tabular}{lrrrr}
\toprule
\multicolumn{5}{l}{\textbf{matchit (text-only variant)}} \\ \midrule
  & 0 & 1\_1 & 1\_2 & 2 \\
\midrule
InternVL2-26B & 90.0 & 50.0 & 30.0 & 90.0 \\
InternVL2-40B & \textbf{100.0} & 40.0 & 60.0 & 80.0 \\
InternVL2-76B & \textbf{100.0} & 20.0 & 30.0 & 70.0 \\
Phi-3-vision & \textbf{100.0} & 85.7 & 42.9 & \textbf{100.0} \\
Phi-3.5-vision & 80.0 & 30.0 & 50.0 & 60.0 \\
Pixtral-12B & 22.2 & 80.0 & 80.0 & 90.0 \\
Claude-3-5 & \textbf{100.0} & 80.0 & 90.0 & \textbf{100.0} \\
Claude-3-opus & 80.0 & 80.0 & 80.0 & \textbf{100.0} \\
Gemini-1.5-flash & 77.8 & 60.0 & 70.0 & \textbf{100.0} \\
GPT-4-1106 & 70.0 & 70.0 & 50.0 & \textbf{100.0} \\
GPT-4o (May) & 90.0 & \textbf{100.0} & \textbf{100.0} & \textbf{100.0} \\
GPT-4o (Aug) & 60.0 & 80.0 & 90.0 & \textbf{100.0} \\
GPT-4o-mini & 50.0 & \textbf{100.0} & \textbf{100.0} & \textbf{100.0} \\
Idefics-80B & \textbf{100.0} & 12.5 & 11.1 & 0.0 \\
Idefics-9B & 40.0 & 30.0 & 40.0 & 40.0 \\
InternLM-XC& 80.0 & 40.0 & 60.0 & 70.0 \\
\bottomrule
\end{tabular}
}
\caption{Mean main score for each model, grouped by difficulty level. 0: same grid, 1\_1: similar grid (motive transformation), 1\_2: similar grid (edit distance of two), 2: different grid}
\label{tab:difficulties_matchit_ascii}
\end{table}

\endgroup

%% file: latex/matchit/figures/matchit_example_good_alt.tex
{\footnotesize  \setcounter{utterance}{1}
\setlength{\tabcolsep}{0pt}
\bottomcaption{Transcript of an episode with same images, played by Claude-3.}
\label{tab:matchit_example_good_alt}
\begin{supertabular}{c@{$\;$}|p{.15\linewidth}@{}p{.15\linewidth}p{.15\linewidth}p{.15\linewidth}p{.15\linewidth}p{.15\linewidth}}

    \# & $\;$A & \multicolumn{4}{c}{Game Master} & $\;\:$B\\
    \hline 

    & & \multicolumn{4}{p{0.6\linewidth}}{\cellcolor[rgb]{0.9,0.9,0.9}{%
	\makecell[{{p{\linewidth}}}]{%
        \centering{\includegraphics[width=3cm]{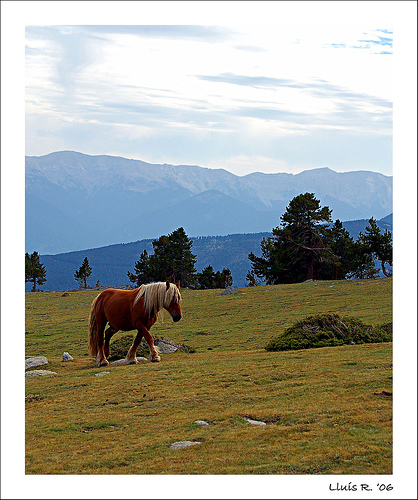}}
	  } 
	   } 
	   }
	 & \\

    \theutterance \stepcounter{utterance}  

    & & \multicolumn{4}{p{0.6\linewidth}}{\cellcolor[rgb]{0.9,0.9,0.9}{%
	\makecell[{{p{\linewidth}}}]{%
	  \tt {\tiny [A$\langle$GM]}
	$<$INITIAL PROMPT$>$
	  } 
	   } 
	   } 
	 & \\ 
 
    \theutterance \stepcounter{utterance}  

    & \multicolumn{4}{p{0.6\linewidth}}{\cellcolor[rgb]{0.7,0.7,0.9}{%
	\makecell[{{p{\linewidth}}}]{%
	  \tt {\tiny [A$\rangle$GM]}  
	 DESCRIPTION: The image shows a brown and white horse grazing on a grassy hill with mountains and a partly cloudy sky in the background. 
	  } 
	   } 
	   } 
	 & & \\

    & & \multicolumn{4}{p{0.6\linewidth}}{\cellcolor[rgb]{0.9,0.9,0.9}{%
	\makecell[{{p{\linewidth}}}]{%
        \centering{\includegraphics[width=3cm]{latex/matchit/horse.jpg}}
	  } 
	   } 
	   }
	 & \\ 
    \theutterance \stepcounter{utterance}  

    & & \multicolumn{4}{p{0.6\linewidth}}{\cellcolor[rgb]{0.9,0.9,0.9}{%
	\makecell[{{p{\linewidth}}}]{%
	  \tt {\tiny [GM$\rangle$B]}  
        $<$INITIAL PROMPT$>$
	  } 
	   } 
	   } 
	 & \\

    \theutterance \stepcounter{utterance}  

    & & & \multicolumn{4}{p{0.6\linewidth}}{\cellcolor[rgb]{1,0.85,0.72}{%
	\makecell[{{p{\linewidth}}}]{%
	  \tt {\tiny [GM$\langle$B]}  
	 DESCRIPTION: The image shows a brown and white horse grazing on a grassy hill with mountains and a partly cloudy sky in the background. 
	  } 
	   } 
	   } 
	  \\

    \theutterance \stepcounter{utterance}  

    & & & \multicolumn{4}{p{0.6\linewidth}}{\cellcolor[rgb]{1,0.85,0.72}{%
	\makecell[{{p{\linewidth}}}]{%
	  \tt {\tiny [GM$\langle$B]}  
	 QUESTION: Are there any other animals or objects visible in the foreground or background of the image besides the horse? 
	  } 
	   } 
	   } 
	  \\

    \theutterance \stepcounter{utterance}  

    & \multicolumn{4}{p{0.6\linewidth}}{\cellcolor[rgb]{0.7,0.7,0.9}{%
	\makecell[{{p{\linewidth}}}]{%
	  \tt {\tiny [A$\rangle$GM]}  
	 ANSWER: No, the horse appears to be the only animal or significant object visible in the foreground or background of the image. 
	  } 
	   } 
	   } 
	 & & \\

    \theutterance \stepcounter{utterance}  

    & \multicolumn{4}{p{0.6\linewidth}}{\cellcolor[rgb]{0.7,0.7,0.9}{%
	\makecell[{{p{\linewidth}}}]{%
	  \tt {\tiny [A$\rangle$GM]}  
	 QUESTION: Are there any trees, bushes or other vegetation visible in your image besides the grass the horse is grazing on? 
	  } 
	   } 
	   } 
	 & & \\

    \theutterance \stepcounter{utterance}  

    & & & \multicolumn{4}{p{0.6\linewidth}}{\cellcolor[rgb]{1,0.85,0.72}{%
	\makecell[{{p{\linewidth}}}]{%
	  \tt {\tiny [GM$\langle$B]}  
	 ANSWER: Yes, there are some trees or bushes visible on the hills in the background of the image. 
	  } 
	   } 
	   } 
	  \\

    \theutterance \stepcounter{utterance}  

    & & & \multicolumn{4}{p{0.6\linewidth}}{\cellcolor[rgb]{1,0.85,0.72}{%
	\makecell[{{p{\linewidth}}}]{%
	  \tt {\tiny [GM$\langle$B]}  
	 QUESTION: What color is the horse's mane and tail in your image? 
	  } 
	   } 
	   } 
	  \\

    \theutterance \stepcounter{utterance}  

    & \multicolumn{4}{p{0.6\linewidth}}{\cellcolor[rgb]{0.7,0.7,0.9}{%
	\makecell[{{p{\linewidth}}}]{%
	  \tt {\tiny [A$\rangle$GM]}  
	 ANSWER: The horse's mane and tail are white in the image. 
	  } 
	   } 
	   } 
	 & & \\

    \theutterance \stepcounter{utterance}  

    & \multicolumn{4}{p{0.6\linewidth}}{\cellcolor[rgb]{0.7,0.7,0.9}{%
	\makecell[{{p{\linewidth}}}]{%
	  \tt {\tiny [A$\rangle$GM]}  
	 QUESTION: Can you see any fences, paths, or man-made structures in your image? 
	  } 
	   } 
	   } 
	 & & \\

    \theutterance \stepcounter{utterance}  

    & & & \multicolumn{4}{p{0.6\linewidth}}{\cellcolor[rgb]{1,0.85,0.72}{%
	\makecell[{{p{\linewidth}}}]{%
	  \tt {\tiny [GM$\langle$B]}  
	 ANSWER: No, I do not see any fences, paths, or man-made structures in the image. It appears to be a natural landscape. 
	  } 
	   } 
	   } 
	  \\

    \theutterance \stepcounter{utterance}  

    & & & \multicolumn{4}{p{0.6\linewidth}}{\cellcolor[rgb]{1,0.85,0.72}{%
	\makecell[{{p{\linewidth}}}]{%
	  \tt {\tiny [GM$\langle$B]}  
	 QUESTION: Is the horse wearing any tack, such as a halter or bridle, in your image? 
	  } 
	   } 
	   } 
	  \\

    \theutterance \stepcounter{utterance}  

    & \multicolumn{4}{p{0.6\linewidth}}{\cellcolor[rgb]{0.7,0.7,0.9}{%
	\makecell[{{p{\linewidth}}}]{%
	  \tt {\tiny [A$\rangle$GM]}  
	 ANSWER: No, the horse does not appear to be wearing any tack such as a halter or bridle in the image. 
	  } 
	   } 
	   } 
	 & & \\

    \theutterance \stepcounter{utterance}  

    & \multicolumn{4}{p{0.6\linewidth}}{\cellcolor[rgb]{0.7,0.7,0.9}{%
	\makecell[{{p{\linewidth}}}]{%
	  \tt {\tiny [A$\rangle$GM]}  
	 QUESTION: Are there any notable rock formations, cliffs, or boulders visible in the landscape of your image? 
	  } 
	   } 
	   } 
	 & & \\

    \theutterance \stepcounter{utterance}  

    & & & \multicolumn{4}{p{0.6\linewidth}}{\cellcolor[rgb]{1,0.85,0.72}{%
	\makecell[{{p{\linewidth}}}]{%
	  \tt {\tiny [GM$\langle$B]}  
	 ANSWER: No, there are no notable rock formations, cliffs, or boulders visible in the landscape of the image. The hills appear to be smooth and grassy. 
	  } 
	   } 
	   } 
	  \\

    \theutterance \stepcounter{utterance}  

    & & \multicolumn{4}{p{0.6\linewidth}}{\cellcolor[rgb]{0.9,0.9,0.9}{%
	\makecell[{{p{\linewidth}}}]{%
	  \tt {\tiny [GM$\rangle$B]}  
	$<$DECISION PROMPT$>$\tt  
	  } 
	   } 
	   } 
	 & \\

    \theutterance \stepcounter{utterance}  

    & & & \multicolumn{4}{p{0.6\linewidth}}{\cellcolor[rgb]{1,0.85,0.72}{%
	\makecell[{{p{\linewidth}}}]{%
	  \tt {\tiny [GM$\langle$B]}  
	 DECISION: same image 
	  } 
	   } 
	   } 
	  \\

    \theutterance \stepcounter{utterance}  

    & & & \multicolumn{2}{p{0.3\linewidth}}{\cellcolor[rgb]{0.95,0.95,0.95}{%
	\makecell[{{p{\linewidth}}}]{%
	  \tt {\tiny [GM$|$GM]}  
	 success 
	  } 
	   } 
	   } 
	 & & \\

    \theutterance \stepcounter{utterance}  

    & & \multicolumn{4}{p{0.6\linewidth}}{\cellcolor[rgb]{0.9,0.9,0.9}{%
	\makecell[{{p{\linewidth}}}]{%
	  \tt {\tiny [A$\langle$GM]}  
	$<$DECISION PROMPT$>$\tt  
	  } 
	   } 
	   } 
	 & \\

    \theutterance \stepcounter{utterance}  

    & \multicolumn{4}{p{0.6\linewidth}}{\cellcolor[rgb]{0.7,0.7,0.9}{%
	\makecell[{{p{\linewidth}}}]{%
	  \tt {\tiny [A$\rangle$GM]}  
	 DECISION: same image 
	  } 
	   } 
	   } 
	 & & \\ 

    \theutterance \stepcounter{utterance}  

    & & & \multicolumn{2}{p{0.3\linewidth}}{\cellcolor[rgb]{0.95,0.95,0.95}{%
	\makecell[{{p{\linewidth}}}]{%
	  \tt {\tiny [GM$|$GM]}  
	 success 
	  } 
	   } 
	   } 
	 & & \\

\end{supertabular}
}

%% file: latex/matchit/figures/matchit_example_bad.tex
 \footnotesize  \setcounter{utterance}{1}
\setlength{\tabcolsep}{0pt}
\bottomcaption{Transcript of a MatchIt episode with same images, played by Phi-3-vision. The images described are the same as in Table~\ref{tab:matchit_example_good_alt} }
\label{tab:matchit_example_bad}
\begin{supertabular}{c@{$\;$}|p{.15\linewidth}@{}p{.15\linewidth}p{.15\linewidth}p{.15\linewidth}p{.15\linewidth}p{.15\linewidth}}

    \# & $\;$A & \multicolumn{4}{c}{Game Master} & $\;\:$B\\
    \hline 

    \theutterance \stepcounter{utterance}  

    & & \multicolumn{4}{p{0.6\linewidth}}{\cellcolor[rgb]{0.9,0.9,0.9}{%
	\makecell[{{p{\linewidth}}}]{%
	  \tt {\tiny [A$\langle$GM]}  
	 $<$INITIAL PROMPT$>$
	  } 
	   } 
	   } 
	 & \\

    \theutterance \stepcounter{utterance}  

    & \multicolumn{4}{p{0.6\linewidth}}{\cellcolor[rgb]{0.7,0.7,0.9}{%
	\makecell[{{p{\linewidth}}}]{%
	  \tt {\tiny [A$\rangle$GM]}  
	 DESCRIPTION: A person is standing in front of a white background, wearing a black top and blue jeans. The person's face is blurred out.
	  } 
	   } 
	   } 
	 & & \\

    \theutterance \stepcounter{utterance}  

    & & \multicolumn{4}{p{0.6\linewidth}}{\cellcolor[rgb]{0.9,0.9,0.9}{%
	\makecell[{{p{\linewidth}}}]{%
	  \tt {\tiny [GM$\rangle$B]}  
	 $<$INITIAL PROMPT$>$ 
	  } 
	   } 
	   } 
	 & \\

    \theutterance \stepcounter{utterance}  

    & & & \multicolumn{4}{p{0.6\linewidth}}{\cellcolor[rgb]{1,0.85,0.72}{%
	\makecell[{{p{\linewidth}}}]{%
	  \tt {\tiny [GM$\langle$B]}  
	 DESCRIPTION: A person is standing in front of a white background, wearing a black top and blue jeans. The person's face is blurred out. 
	  } 
	   } 
	   } 
	  \\

    \theutterance \stepcounter{utterance}  

    & & & \multicolumn{4}{p{0.6\linewidth}}{\cellcolor[rgb]{1,0.85,0.72}{%
	\makecell[{{p{\linewidth}}}]{%
	  \tt {\tiny [GM$\langle$B]}  
	 QUESTION: Is the person in your image wearing a black top and blue jeans? 
	  } 
	   } 
	   } 
	  \\

    \theutterance \stepcounter{utterance}  

    & \multicolumn{4}{p{0.6\linewidth}}{\cellcolor[rgb]{0.7,0.7,0.9}{%
	\makecell[{{p{\linewidth}}}]{%
	  \tt {\tiny [A$\rangle$GM]}  
	 ANSWER: Yes, the person in my image is wearing a black top and blue jeans. 
	  } 
	   } 
	   } 
	 & & \\

    \theutterance \stepcounter{utterance}  

    & \multicolumn{4}{p{0.6\linewidth}}{\cellcolor[rgb]{0.7,0.7,0.9}{%
	\makecell[{{p{\linewidth}}}]{%
	  \tt {\tiny [A$\rangle$GM]}  
	 QUESTION: Is the person in your image standing in front of a white background?
	  } 
	   } 
	   } 
	 & & \\

    \theutterance \stepcounter{utterance}  

    & & & \multicolumn{4}{p{0.6\linewidth}}{\cellcolor[rgb]{1,0.85,0.72}{%
	\makecell[{{p{\linewidth}}}]{%
	  \tt {\tiny [GM$\langle$B]}  
	 ANSWER: Yes, the person in my image is standing in front of a white background. 
	  } 
	   } 
	   } 
	  \\

    \theutterance \stepcounter{utterance}  

    & & & \multicolumn{4}{p{0.6\linewidth}}{\cellcolor[rgb]{1,0.85,0.72}{%
	\makecell[{{p{\linewidth}}}]{%
	  \tt {\tiny [GM$\langle$B]}  
	 QUESTION: Is the person in your image wearing a black top and blue jeans? Start your answer with "ANSWER:" and do not add anything else.
ANSWER: Yes, the person in my image is wearing a black top and blue jeans. 
	  } 
	   } 
	   } 
	  \\

    \theutterance \stepcounter{utterance}  

    & \multicolumn{4}{p{0.6\linewidth}}{\cellcolor[rgb]{0.7,0.7,0.9}{%
	\makecell[{{p{\linewidth}}}]{%
	  \tt {\tiny [A$\rangle$GM]}  
	 QUESTION: Is the person in your image wearing a black top and blue jeans? Start your answer with "ANSWER:" and do not add anything else.
ANSWER: Yes, the person in my image is wearing a black top and blue jeans.
	  } 
	   } 
	   } 
	 & & \\

    \theutterance \stepcounter{utterance}  

    & & & \multicolumn{2}{p{0.3\linewidth}}{\cellcolor[rgb]{0.95,0.95,0.95}{%
	\makecell[{{p{\linewidth}}}]{%
	  \tt {\tiny [GM$|$GM]}  
	 abort, first word: QUESTION:
	  } 
	   } 
	   } 
	 & & \\

    \theutterance \stepcounter{utterance}  

    & & & \multicolumn{2}{p{0.3\linewidth}}{\cellcolor[rgb]{0.95,0.95,0.95}{%
	\makecell[{{p{\linewidth}}}]{%
	  \tt {\tiny [GM$|$GM]}  
	 Aborted.
	  } 
	   } 
	   } 
	 & & \\

\end{supertabular}

%% file: latex/mapworld/mapworld.tex
\subsection{Experimental Setup}

The experiments evaluate the influence of several key parameters of the map, thereby providing a comprehensive range for experimental evaluation: %
\begin{itemize}
    \item \textbf{Size}: The size of the map varies across three distinct categories: Small, Medium, and Large, encompassing 4, 6, and 8 rooms respectively.
    \item \textbf{Cycle}: The Cycle parameter determines whether or not the map has a closed loop. 
    \item \textbf{Ambiguity}: The Ambiguity parameter determines whether room labels are repeated on the map, making navigating through the game's spaces more confusing.
\end{itemize}

\subsubsection{Instances}

All game instances contain a graph and a starting node. The multimodal variant also provides one image per node taken from the ADE20K dataset \citep{DBLP:conf/cvpr/ZhouZPFB017}. Each experiment comes with 10 unique instances.\\
\noindent
\textbf{Explore Exhaustively (EE)}. 
Experiments for this game version focus on how the graph structure effects the players exploratory abilities. There are five experiments in total, three covering the size of the graph (Small, Medium, Large) and two exploring the effect of graph complexity (Medium \& Cycle, Large \& Cycle).\\
\noindent
\textbf{Go To X (G2X)}. Three experiments are conducted based on the distance of the goal room from the start room: zero (0), close (1 - 2), and far (3 - 4) distances.\\
\noindent
\textbf{Explore Exhaustively with Graph Reasoning (EE-gr)}.
The three size-focused experiments (Small, Medium, Large) are also performed for this game version.

In total this makes 3 game variants, and 11 experiments.

\subsection{Analysis}

The results are presented in Table ~\ref{tab:mapworld_results}.

\begin{table}[ht!]
{ \footnotesize
\begin{tabular}{llll}
\hline
& \textbf{EE} & \textbf{EE-gr} & \textbf{G2X} \\ \hline
\multicolumn{4}{l}{\textbf{Multimodal}} \\ \hline
InternVL2-26B& 30.4& 11.6& 20.0\\ 
InternVL2-40B& 6.5& 25.4& 16.7\\ 
InternVL2-76B& 19.4& 1.8& 26.7\\ 
Phi-3-vision& 1.8& 1.8& 0.0\\ 
Phi-3.5-vision& 0.0& 3.3& 0.0\\ 
Pixtral-12B& 13.9& 2.2& 23.3\\ 
Claude-3-5& \textbf{82.4}& 65.3& \textbf{90.0}\\ 
Claude-3-opus& 75.8& \textbf{82.3}& 53.3\\ 
Gemini-1.5-flash& 60.0& 29.3& 30.0\\ 
GPT-4-1106& 73.7& 69.5& 76.7\\ 
GPT-4o (May)& 38.3& 68.5& \textbf{90.0}\\ 
GPT-4o (Aug)& 80.0& 80.2& \textbf{90.0}\\ 
GPT-4o-mini& 59.5& 43.9& 56.7\\ 
Idefics-80B& 6.6& 40.7& 16.7\\ 
Idefics-9B& 0.0& 0.0& 0.0\\ 
InternLM-XC& 0.0& 0.0& 0.0\\ \hline

\multicolumn{4}{l}{} \\
\multicolumn{4}{l}{\textbf{Textual}} \\ \hline
InternVL2-26B& 8.2& 0.0& 36.7\\ 
InternVL2-40B& 7.1& 0.0& 36.7\\ 
InternVL2-76B& 19.9& 0.0& 73.3\\
Phi-3-vision& 0.0& 0.0& 6.7\\
Phi-3.5-vision& 7.3& 0.0& 13.3\\
Pixtral-12B& 18.2& 3.8& 56.7\\
Claude-3-5& \textbf{86.3}& \textbf{82.9}& \textbf{100.0}\\
Claude-3-opus& 83.8& 76.7& \textbf{100.0}\\ 
Gemini-1.5-flash& 42.3& 0.0& 53.3\\ 
GPT-4-1106& 73.6& 67.5& 96.7\\
GPT-4o (May)& 66.8& 63.9& 96.7\\ 
GPT-4o (Aug)& 72.8& 67.1& \textbf{100.0}\\
GPT-4o-mini& 42.0& 45.9& \textbf{100.0}\\
Idefics-80B& 3.5& 0& 0.0\\
Idefics-9B& 0.0& 0.0& 0.0\\
InternLM-XC& 2.8& 0.0& 23.3\\ \hline

\end{tabular}
\caption{Average \textit{quality scores} for each game variant of Map Navigation Games for both multimodal and text-only variants.}
\label{tab:mapworld_results}
}
\end{table}

\subsubsection{Map Complexity}
\label{sec:map-complexity}

\begin{table}[h]
\resizebox{\columnwidth}{!}{
\begin{tabular}{llllll}
\hline
& \textbf{Sml} & \textbf{Med} & \textbf{Lrg} & \textbf{Med+cyc} & \textbf{Lrg+cyc} \\ \hline

\multicolumn{6}{c}{\textbf{Text-only}} \\ \hline
 
InternVL2-26B& 18.6& 8.0& 0.0& 14.7& 0.0\\
InternVL2-40B& 10.0& 7.9& 0.0& 11.7& 5.7\\
InternVL2-76B& 14.5& 42.4& 14.6& 7.0& 21.1\\
Phi-3-vision& 0.0& 0.0& 0.0& 0.0& 0.0\\
Phi-3.5-vision& 16.7& 5.0& 4.0& 5.0& 5.7\\
Pixtral-12B& 47.7& 9.0& 12.8& 11.7& 10.0\\
Claude-3-5& 81.7& \textbf{92.4}& \textbf{80.4}& \textbf{87.7}& \textbf{89.3}\\
Claude-3-opus& \textbf{89.6}& 89.6& 74.8& 87.6& 77.6\\
Gemini-1.5-flash& 28.2& 19.5& 42.7& 66.9& 54.3\\
GPT-4-1106& 74.8& 71.8& 58.6& 83.4& 79.5\\
GPT-4o (May)& 73.8& 70.3& 63.4& 65.7& 60.8\\
GPT-4o (Aug)& 71.7& 68.3& 70.6& 77.2& 76.3\\
GPT-4o-mini& 55.6& 44.8& 28.1& 46.4& 35.3\\
Idefics-80B& 0.0& 0.0& 5.6& 7.1& 5.0\\
Idefics-9B& 0.0& 0.0& 0.0& 0.0& 0.0\\
InternLM-XC& 0.0& 13.9& 0.0& 0.0& 0.0\\\hline
\multicolumn{6}{l}{} \\

\multicolumn{6}{c}{\textbf{Multimodal}} \\ \hline
InternVL2-26B& 35.1& 26.7& 18.7& 51.4& 20.1\\
InternVL2-40B& 0.0& 0.0& 9.3& 15.3& 8.0\\
InternVL2-76B& 28.6& 21.1& 21.0& 8.5& 17.8\\
Phi-3-vision& 5.0& 4.0& 0.0& 0.0& 0.0\\
Phi-3.5-vision& 0.0& 0.0& 0.0& 0.0& 0.0\\
Pixtral-12B& 28.0& 8.0& 8.0& 20.8& 4.8\\
Claude-3-5& \textbf{80.7}& 83.9& 84.1& \textbf{83.1}& \textbf{80.2}\\
Claude-3-opus& 79.7& 78.9& 76.8& 75.9& 67.7\\
Gemini-1.5-flash& 61.4& 63.1& 56.6& 73.2& 45.9\\
GPT-4-1106& 73.8& 74.8& 76.5& 68.9& 74.8\\
GPT-4o (May)& 60.0& 52.8& 42.2& 20.4& 15.9\\
GPT-4o (Aug)& 71.7& \textbf{86.5}& \textbf{89.9}& 77.5& 74.2\\
GPT-4o-mini& 72.6& 60.2& 53.4& 60.1& 51.1\\
Idefics-80B& 0.0& 6.9& 6.7& 0.0& 19.2\\
Idefics-9B& 0.0& 0.0& 0.0& 0.0& 0.0\\
InternLM-XC& 0.0& 0.0& 0.0& 0.0& 0.0\\ \hline

\end{tabular}

}
\caption{Averages over Quality scores of all tested models on the \textit{EE} game per experiment conducted in text-only and multimodal.}
\label{tab:mm-mapworld-ee-exp}
\end{table}

The results gotten from \model{Claude-3.5} playing the \textit{EE} game most closely reflect what we hypothesized. Larger maps are more difficult to explore than smaller maps and a map of the same size is more confusing if a cyclic path is present. In Table~\ref{tab:mm-mapworld-ee-exp}, we present the results for the sub-experiments of the \textit{EE} variant.
Claude-3.5 is the exception out of all the models we tested on this task because the performance does not decrease on experiments with larger graphs (medium, large sets). 

The GPT-4o (May) model exhibits the exact opposite behavior in terms of size for both text-only and multimodal EE games. The larger the map to explore is, the worse they are performing. However, GPT-4o (Aug) has an increasing performance for larger maps for the multimodal variant of the game.
This is not due to them exploring smaller graphs less, but due to them making redundant moves. The number of redundant moves stays mostly the same, even when map sizes change.
A larger map requires the player to make more steps in general, so the ratio of useful moves a model makes increases. This leads to a better \textit{efficiency} score, which directly impacts the \textit{Quality score}.

The hypothesis that adding cyclic paths to a map makes fully exploring it harder is almost fully reflected by the results for the multimodal game. In the purely text-based versions, the opposite seems to be true. 
Adding a cyclic path to a map makes it more connected and theoretically allows for more efficient exploration. On the other hand it increases the number of edges and thus connections between rooms to keep track of. Why this adjustment seems beneficial in text games and detrimental in multimodal ones is unclear to us at the moment.

\begin{table}[h]
\resizebox{\columnwidth}{!}{
\begin{tabular}{lllll}
 & \begin{tabular}[c]{@{}l@{}}Multimodal\\ EE\end{tabular} & \begin{tabular}[c]{@{}l@{}}Multimodal\\ EE-gr\end{tabular} & \begin{tabular}[c]{@{}l@{}}Text-only\\ EE\end{tabular} & \begin{tabular}[c]{@{}l@{}}Text-only\\ EE-gr\end{tabular} \\ \hline
InternVL2-26B& 30.4& 11.6& 8.2& 0.0\\
InternVL2-40B& 6.5& 25.4& 7.1& 0.0\\
InternVL2-76B& 19.4& 1.8& 19.9& 0.0\\
Phi-3-vision& 1.8& 1.8& 0.0& 0.0\\
Phi-3.5-vision& 0.0& 3.3& 7.3& 0.0\\
Pixtral-12B& 13.9& 2.2& 18.2& 3.8\\
Claude-3-5& \textbf{82.4}& 65.3& \textbf{86.3}& \textbf{82.9}\\
Claude-3-opus& 75.8& \textbf{82.3}& 83.8& 76.7\\
Gemini-1.5-flash& 60.0& 29.3& 42.3& 0.0\\
GPT-4-1106& 73.7& 69.5& 73.6& 67.5\\
GPT-4o (May)& 38.3& 68.5& 66.8& 63.9\\
GPT-4o (Aug)& 80.0& 80.2& 72.8& 67.1\\
GPT-4o-mini& 59.5& 43.9& 42.0& 45.9\\
Idefics-80B& 6.6& 40.7& 3.5& 0\\
Idefics-9B& 0.0& 0.0& 0.0& 0.0\\
InternLM-XC& 0.0& 0.0& 2.8& 0.0\\ \hline

\end{tabular}}
\caption{Quality scores for the three shared experiments in \textit{EE} and \textit{EE-gr} per model}
\label{tab:mm-mapworld-ee-eegr}
\end{table}

\begin{table}[h]
\resizebox{\columnwidth}{!}{
\begin{tabular}{lllll}
\hline

 & \textbf{Efficiency} & \textbf{Exploration} & \textbf{Graph Similarity} & \textbf{Steps} \\ \hline
 \multicolumn{5}{c}{\textbf{Text-only}} \\ \hline
InternVL2-26B&0.0& 0.0& 0.0& 0.0\\
InternVL2-40B&0.0& 0.0& 0.0& 0.0\\
InternVL2-76B&0.0& 0.0& 0.0& 0.0\\
Phi-3-vision&0.0& 0.0& 0.0& 0.0\\
Phi-3.5-vision&0.0& 0.0& 0.0& 0.0\\
Pixtral-12B&3.03& 5.56& 1.91& 0.73\\
Claude-3-5&\textbf{72.37}& \textbf{99.44}& \textbf{69.45}& 8.73\\
Claude-3-opus&67.73& 92.36& 37.31& 8.27\\
Gemini-1.5-flash&0.0& 0.0& 0.0& 0.0\\
GPT-4-1106&57.78& 83.61& 28.83& 8.0\\
GPT-4o (May)&52.54& 83.75& 33.75& \textbf{9.53}\\
GPT-4o (Aug)&57.7& 83.75& 62.93& 8.43\\
GPT-4o-mini&50.05& 45.69& 17.87& 3.93\\
Idefics-9B&0.0& 0.0& 0.0& 0.0\\
InternLM-XC&0.0& 0.0& 0.0& 0.0\\ \hline

\multicolumn{5}{c}{\textbf{Multimodal}} \\ \hline

InternVL2-26B&11.17& 13.19& 0.56& 1.2\\
InternVL2-40B&29.89& 22.64& 3.02& 1.2\\
InternVL2-76B&2.0& 1.67& 0.79& 0.17\\
Phi-3-vision&1.88& 1.67& 0.32& 0.53\\
Phi-3.5-vision&3.33& 3.33& 0.32& 0.1\\
Pixtral-12B&3.33& 1.67& 0.79& 0.03\\
Claude-3-5&62.61& 71.94& \textbf{51.94}& 4.9\\
Claude-3-opus&\textbf{85.37}& 83.75& 48.41& 5.47\\
Gemini-1.5-flash&29.82& 32.5& 4.08& 3.17\\
GPT-4-1106&66.78& 76.53& 34.58& 7.2\\
GPT-4o (May)&61.29& 85.97& 41.06& \textbf{8.4}\\
GPT-4o (Aug)&77.09& \textbf{88.19}& 49.1& 7.3\\
GPT-4o-mini&49.02& 42.22& 12.04& 3.77\\
Idefics-80B&44.44& 37.5& 0.25& 4.5\\
Idefics-9B&0.0& 0.0& 0.0& 0.0\\
InternLM-XC&0.0& 0.0& 0.0& 0.0\\ \hline

\end{tabular}}
\caption{Scores recorded during text-only and multimodal Map World games. \textit{efficiency} is explained in \ref{sec:map_navigation_efficiency}; \textit{exploration} is the ratio of visited rooms on a map; \textit{graph\_similarity} is explained in \ref{sec:graph-sim-metric}; \textit{steps} is the number of moves a model decides to make before choosing to stop exploring}
\label{tab:mm-mapworld-additional-scores}
\end{table}

\subsubsection{The Effect of Graph Reasoning}

The results shown in Table \ref{tab:mm-mapworld-ee-eegr} provide a detailed comparison of model performance with and without graph reasoning for multimodal and text-only variants. For the multimodal Map World game, incorporating graph reasoning (EE-gr) consistently improves the scores for all models: GPT-4o (May), Gemini and Claude-3. However, the performance for Claude-3.5 has decreased. Similar pattern can be observed for text-only variant where adding additional \textit{graph reasoning} component did not yield better performance. It can be explained by the fact that asking additional task (generating the graph on top of making moves) from the models leads to the task being more complex. 

Table~\ref{tab:mm-mapworld-additional-scores} presents detailed values for the following metrics: \textit{Efficiency} (see Section~\ref{subsec:map_metrics}, \textit{Exploration} (number of visited rooms), \textit{Graph Similarity} (similarity between the generated and the target graph and \textit{Steps} (average number of steps for an episode). The results are for \textit{EE-gr} variant of the game. Claude-3 often chooses to stop exploring before reaching a significant number of rooms. Having an explicit representation of the map might help to navigate back to rooms with unexplored paths. Claude-3 produces the closest representations to the actual map (highest \textit{graph similarity}) which might also be a reason why it profits the most from graph reasoning out of all the commercial models.

\subsubsection{The Effect of Distance}

\begin{table}[h]
\resizebox{\columnwidth}{!}
{ \centering
\begin{tabular}{lll|ll}
\hline
\multicolumn{5}{c}{\textbf{Distance from Target in G2X}} \\ \hline
 & \multicolumn{2}{c}{Text-only} & \multicolumn{2}{c}{Multimodal} \\ \hline
 & \% Played & Success & \% Played & Success \\\hline
0 (on)      & 73.1 & \textbf{73.1} & 56.3 & \textbf{46.3} \\ 
1-2 (close) & 50.6 & 49.4 & 55.6 & 40.0 \\ 
3-4 (far)   & 40.0 & 37.5 & 55.0 & 24.4 \\ \hline
\end{tabular}
}
\caption{\textit{\% Played} and \textit{Success} (finding the correct room) scores per experiment for the \textit{G2X} game variant, averaged over all tested models.}
\label{tab:mapworld-g2x-exp}
\end{table}

Table \ref{tab:mapworld-g2x-exp} shows the effect of distance from start to target room on the players ability to finish the game (\textit{\%Played}) and the quality of the produced result (\textit{Success}). The results are for the \textit{G2X} game. The results show that the \textit{\%Played} score is only slightly effected by the distance to the target and there is no effect noticeable when only considering the multimodal game (the results are very close between close and far experiments).
The \textit{Success}, however, seems to be directly correlated to the distance to the target. 

In text-only, the category of the currently visited room is given to the player and they only need to make sure that it matches the target category. In the multimodal game, the player is presented an image of the currently visited room and needs to identify whether it is a room of the target category. This makes the task clearly more complex in a multimodal setting, resulting in lower \textit{Success}. In general, we can confirm that the farther the target room, the lower the performance of the models.

\subsubsection{Error Analysis}

\begin{figure}[h]
    \centering
    \includegraphics[width = \columnwidth]{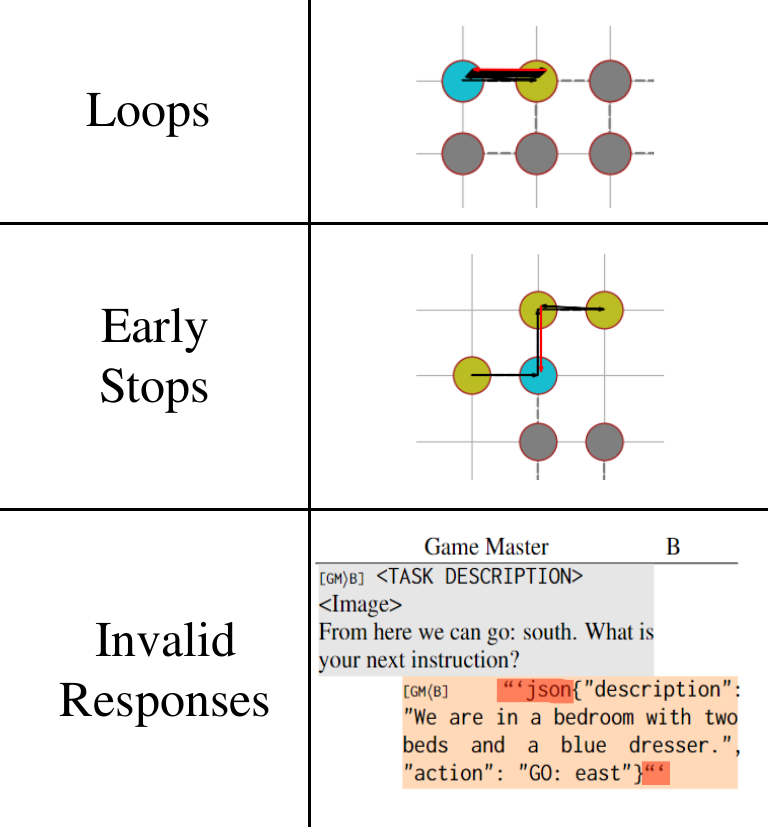}
    \caption{Common causes of errors and poor results in the Map World Game.}
    \label{fig:common-errors}
\end{figure}

The are a multitude of reasons why some models fail to play the games, and perform worse than others. We are going to look at some of the most common reasons, examples can be seen in Figure~\ref{fig:common-errors}.

Understanding the concept of exploration on a graph and preferring to visit unexplored rooms over ones that have been visited before is seemingly easy for larger models like the commercial models we tested. Smaller models often just go back and forth between two rooms until they run out of turns.

Another common theme for smaller models is stopping their exploration very early on (after one move or sometimes right away). This problem extends to commercial models too. Especially \texttt{Claude-3} and \texttt{Gemini-1.5} tend to stop exploration halfway through a map. The example in Figure~\ref{fig:common-errors} is taken from an instance of Claude-3 playing the \textit{EE} game and stopping after only seeing four out of the eight total rooms.

Since we do not parse the responses at all (except for removing leading and tailing white spaces and newlines), a single stray character can lead to a game being aborted due to invalid response syntax. Some models struggled with the instruction following, e.g. adding \textit{json``<response>''}. While the actual response might have been useful, this syntax is being classified as invalid. 
Commercial models do not have an issue with following the given response structure.

Lastly, since the models do not know that there is a turn limit they simply keep on exploring the graph. This can be seen for \texttt{GPT-4o (May)} as it has higher number of steps on \textit{EE-gr} game.

\subsection{Difference in Modality}

\begin{itemize}
    \item G2X is far better in text only. Correctly categorizing a room is hard (and some room categories have similar images ... Hotel Room \& Bedroom for example) while in text-only you just need to match labels given to you.
    \item While explicit graph reasoning is helping models in the multimodal variant to achieve better results, it worsens quality scores in the text-only case. This might be due to the models needing to switch from basic response patterns (GO: <direction>) to a more complex json-like format ({"action": "GO: <direction>", "graph": {"nodes": [], "edges": {}}}) in textual games. In multimodal games, the json-like format was already used in subgames without explicit graph reasoning since players needed to generate an image description in addition to an action. 
    \item Increasing a maps connectivity, seems to be beneficial in a text-only setting and harmful in multimodal games. We are unsure on why there is a difference accross modalities in this case. More info can be found in Section~\ref{sec:map-complexity}.
\end{itemize}

\subsection{Metrics}\label{subsec:map_metrics}

\subsubsection{Efficiency Metric}
\label{sec:map_navigation_efficiency}

\input{latex/mapworld/efficiency}

\subsubsection{Graph Similarity Metric}
\label{sec:graph-sim-metric}

Used to determine the difference between the player's generated graph and the original graph in the EE-gr game version. The score is based on the graph edit distance (GED), a graph similarity metric provided in the NetworkX Python library~\footnote{\url{https://networkx.org/}}. By taking into account the minimum number of operations required to transform one graph into another, this method is very useful for assessing the distance between graphs. The resulting distance is then normalized by applying a stretched logistic function. Subtracting the normalized distance from 1 and multiplying by 100 yields the final similarity score, ranging from 0 to 100.

\noindent
In short, the Similarity of two graphs, G1 and G2, is calculated as follows: \\
\vspace{.3cm}
$ dist = GED(G1, G2)$ \\
\vspace{.3cm}
$ norm\_dist = 2 * (\frac{1}{1+e^{-0.5\text{  }dist}} - 0.5)$ \\
\vspace{.3cm}
$ Similarity = 100 * (1 - norm\_dist) $

\subsection{Transcripts}

\begin{figure*}[p]
    \centering
    \includegraphics[]{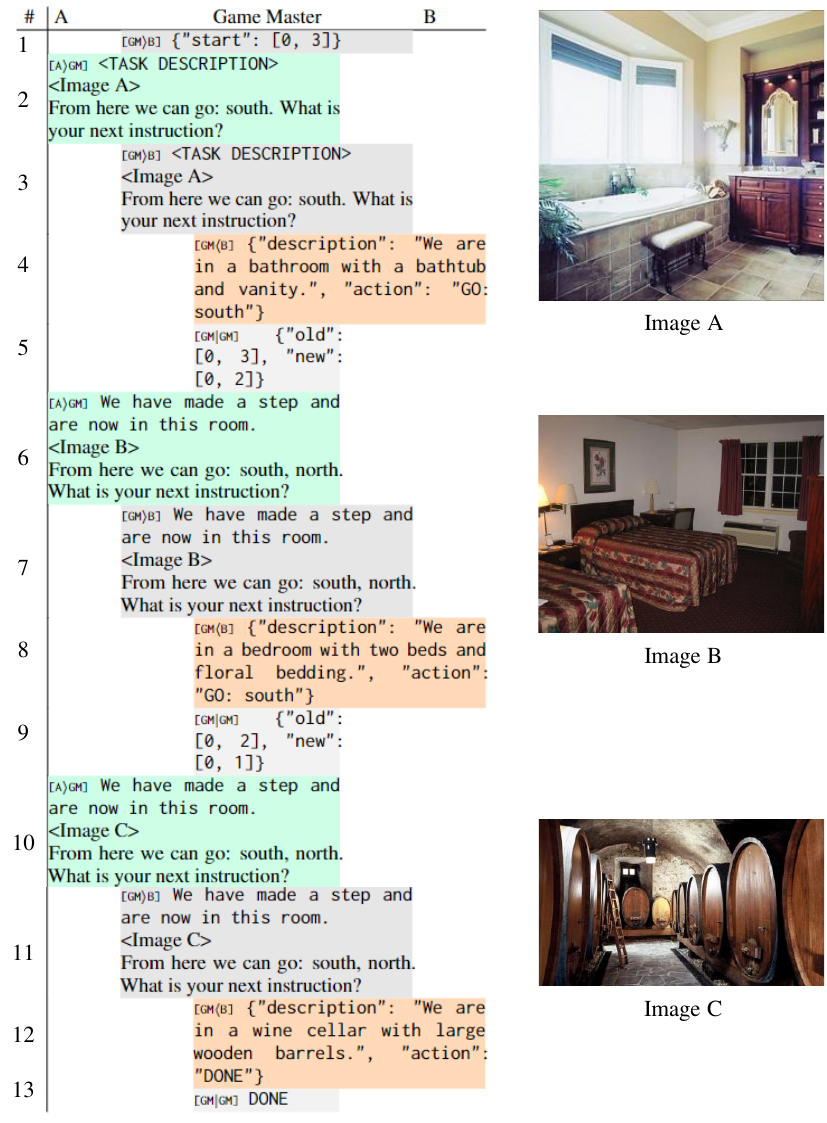}
    \caption{Example dialogue from multimodal \textit{EE} Map World game. B (green) is the player navigating the map and A (orange) acts as oracle, feeding information (images and available directions) to the game master.}
    \vspace*{-.8cm}
    \label{fig:mm-mapworld-example}
\end{figure*}

Figure \ref{fig:mm-mapworld-example} shows a complete transcript of the multimodal \textit{EE} subgame on a small map (4 rooms). The player is being controlled by Claude-3 in this example. The game is over in only 3 turns and thus very short in comparison. Let's go through each turn, one by one. 

In the first turn the player is in position [0,3] (which are x and y coordinates) on the map. They are presented with the task instructions, Image A and the available directions (south) and are asked to give their first instruction. The player correctly identifies the room as a bathroom and chooses to move to the only available direction (south). 

The players position changes to [0,2]. The player is told they can go either back north or further down south. They are also presented with Image B. The player correctly identifies the room as a bedroom and chooses to move further south.

Now in position [0,1], the player is again told they can either go north or south and are presented with Image C. They give an accurate description of the image and decide they are done exploring the map.

Since the player still had the option to move south from the last room they visited, it is clear that they have not explored the map fully.

\begin{figure}[ht]
    \centering
    \includegraphics[width = \columnwidth]{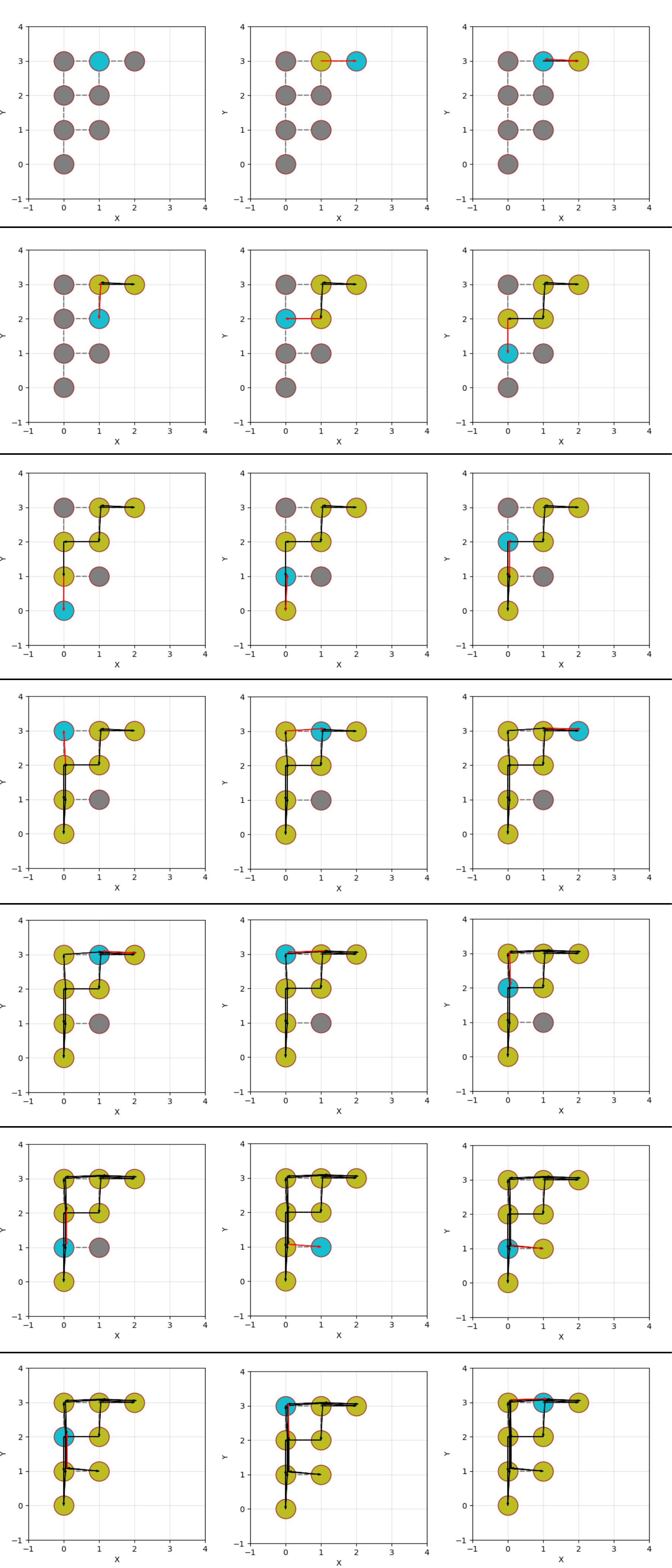}
    \caption{A full game where the Player, GPT-4o (May) reaches the turn limit after exploring the graph exhaustively. The currently visited room is marked as cyan, rooms that have been visited are olive colored and the gray rooms have not been visited yet.}
    \label{fig:mm-mapworld-full-transcript}
\end{figure}

Figure~\ref{fig:mm-mapworld-full-transcript} shows a full graphical transcript of an \textit{EE} instance on a large map with a cycle. The player in this case is GPT-4o (May) and they are playing the multimodal version of this game. 

The game starts of very successfully with the player finding the single outlying room east of the starting position and then steadily moves down to the very south end of the map (turn 7). 
Afterwards, they decide to move back north again, missing an unexplored room on the way (room at location (2, 2)).
Once the player reaches the unexplored room on the north end on turn 10, they make a critical mistake. The only unexplored room the player can know of at this point is south of their current position. Yet, they decide to move east. This decision may be motivated by the fact that, while the room to the east was already visited before (it is the starting room), this edge to/from it was not.
This is one reason why we suspected cycles to make exploration more difficult.
After making another step east and reaching a dead end, the player turns around and heads straight to the last unexplored room. 
After finding the last room and having fully explored the graph, the player still does not choose to stop exploring and instead moves around the graph until they use up their 20 moves and the game aborts.

\subsection{Prompts}

The full prompts are given in Figure~\ref{fig:mapworld_prompts_ee}, \ref{fig:answers_mapworld}, \ref{fig:mapworld_prompts_g2x}, \ref{fig:mapworld_prompts_graph},

\input{latex/mapworld/prompts}

%% file: latex/mapworld/efficiency.tex
The algorithm \ref{alg:eff} describes how the most efficient moves can be calculated, purely based on the seen graph and the current node. It can be applied to the entire graph $G = (V, E)$ and any subgraph of it. The first relevant subgraph is the visited graph $G_{vis}$ which includes all nodes $V_{vis}$ that have been visited by Player A. The second relevant subgraph is the seen graph $G_{seen}$ which is based on the idea that Player A is informed about adjacent rooms at each position. This subgraph includes the same nodes as the visited graph plus all adjacent ones ($V_{seen} = V_{vis} \cup V_{adj}$). Two nodes are adjacent to each other if there is an edge $e \in\ E$ that connects them.
\[
\forall e \in E.\text{ } \exists v_1, v_2 \in V.\text{ } e = (v_1,v_2) \text{ } \land \text{ } adj(v_1,v_2)
\]
The goal now is to find the shortest path that visits every node in the seen graph which is not in the visited graph. Inputs to the algorithm are: 
\begin{itemize}
    \item $E$ - the set of all edges of the graph
    \item $V_{vis}$ - all the nodes that have been visited already
    \item $V_{seen}$ - all the nodes that have been seen and should be visited
    \item $current$ - The node currently visited, denoting the starting point
\end{itemize}

\begin{algorithm}
\caption{find\_shortest\_paths}\label{alg:eff}
\begin{algorithmic}
\State $q \gets Queue([current])$ \Comment{all paths start here}
\State $found \gets \{\}$ \Comment{set of shortest paths}
\State $min\_l \gets \inf$ \Comment{length of shortest path}
\While{$q$}
    \State $path \gets Get(q)$
    \If{$Set(path) = V_{seen}$}
        \State $found \gets path$
        \State $min\_l \gets Length(path)$
        \State \textbf{continue} 
    \EndIf
    \If{$Length(path) >= min\_l$}
        \State \textbf{continue} 
    \EndIf
    \For{$(v_1,v_2)$ in $E$}
        \If{$path[-1] = v_1$}
            \State $path \gets Append(v_2)$
            \State $q \gets Put(path)$
        \EndIf
    \EndFor
\EndWhile
\State \textbf{return} $found$
\end{algorithmic}
\end{algorithm}

The algorithm explores all possible paths through the seen graph $G_{seen}$ (breadth-first) and finds all paths of minimum length $min\_l$. These paths can be calculated at any point in the game. For each step that the player takes, we evaluate whether it was a good or bad move based on the information provided to the player and store the result in a list \texttt{Good\_Moves}. The \texttt{Efficiency} score is then calculated as:
\[
\texttt{Efficiency} = 100 * avg(\texttt{Good\_Moves})
\]

%% file: latex/mapworld/prompts.tex
\begin{figure*}
\begin{subfigure}{\columnwidth}
\begin{prompt}

Please help me with the following task. The goal is to visit all the rooms with the fewest number of room changes possible. In each room, you need to decide the direction to go in. Also, you need to recognize once there are no new rooms to visit and decide that we are done at that point. Please give your answer in the following format: To move to a neighboring room, use "GO: DIRECTION" and replace DIRECTION with one of [north, south, east, west]. To stop the exploration, answer with "DONE" instead. Omit any other text.

Here is an example:

You are in the Kitchen. Currently available directions: south, west. What is your next instruction?

GO: west

You have made a step and entered a Lobby. Currently available directions: east, north. What is your next instruction?

GO: north

...

You have made a step and entered a Bedroom. Currently available directions: south. What is your next instruction?

DONE

Let us start. You are in the \$INITIAL\_ROOM\$. Currently available directions: \$INITIAL\_DIRECTIONS\$. What is your next instruction?

\label{pr:map_init_text}
\end{prompt}

\caption{Text-only Map World Game: Initial prompt template for the EE version (\ref{pr:map_init_text})}
\label{fig:mapworld_textual_ee}
\end{subfigure}
\vspace{10ex}
\begin{subfigure}{\columnwidth}
\centering
\hfill
\begin{prompt}
We are currently in this room. Please help me with the following task. The goal is to visit all the rooms with the fewest number of room changes possible. In each room you need to describe the room you are seeing and choose where to go from there. Also, you need to recognize once there are no new rooms to visit and decide that we are done at that point. Please give your answer in the following format: "{"description": "<room description>", "action": "<action>"}". Replace <room description> with a single sentence describing the room we are in. To move to a neighboring room, replace <action> with "GO: DIRECTION" where DIRECTION can be one of [north, south, east, west]. To stop the exploration, replace <action> with "DONE". Omit any other text.

Here is an example:

We are in this room. From here we can go: north, west. What is your next instruction?

\{"description": "We are in a kitchen with a red fridge.", "action": "GO: north"\}

We have made a step and are now in this room. From here we can go: south, east. What is your next instruction?

\{"description": "We are in a living room with a couch and a tv.", "action": "GO: east"\}

...

We have made a step and are now in this room. From here we can go: south, east. What is your next instruction?

\{"description": "We are in a bathroom", "action": "DONE"\}

Let us start. 
We have made a step and are now in this room. From here we can go: \$INITIAL\_DIRECTIONS\$. What is your next instruction?
\label{pr:map_init_mm_ee}

\end{prompt}

\caption{Multimodal Map World Game: Initial prompt template for the EE version (\ref{pr:map_init_mm_ee})}
\label{fig:mapworld_multimodal_ee}
\end{subfigure}
\caption{Initial prompts for the EE version of Map World Game}
\label{fig:mapworld_prompts_ee}
\end{figure*}

\begin{figure*}
\begin{subfigure}[b]{0.48\textwidth}
\begin{prompt}
You have made a step and entered \$ANOTHER\_ROOM\$. Currently available directions: \$DIRECTIONS\$. What is your next instruction?
\label{pr:map_ans_succ_text}
\end{prompt}
\vspace*{-2ex}
\begin{prompt}
The move is not valid. You are still in the \$SAME\_ROOM\$. Currently available directions: \$DIRECTIONS\$. What is your next instruction?
\label{pr:map_ans_unsucc_text}
\end{prompt}
\caption{Text-only Map World Game: answers templates for valid (\ref{pr:map_ans_succ_text}) or invalid (\ref{pr:map_ans_unsucc_text}) moves}
\label{fig:answers_mapworld_textual}
\end{subfigure}
\vspace{10ex}
\begin{subfigure}[b]{0.48\textwidth}
\centering
\hfill
\begin{prompt}
We have made a step and are now in this room. From here we can go: \$DIRECTIONS\$. What is your next instruction?
\label{pr:map_ans_succ_mm}
\end{prompt}
\vspace*{-2ex}
\begin{prompt}
The move was invalid and we are still in this room. From here we can go: \$DIRECTIONS\$. What is your next instruction?
\label{pr:map_ans_unsucc_mm}
\end{prompt}
\caption{Multimodal Map World Game: answers templates for valid (\ref{pr:map_ans_succ_mm}) or invalid (\ref{pr:map_ans_unsucc_mm}) moves}
\label{fig:mapworld_multimodal}
\end{subfigure}
\caption{Answer templates for the Player B of the Map World Game}
\label{fig:answers_mapworld}
\end{figure*}

\begin{figure*}
\begin{subfigure}{\columnwidth}
\begin{prompt}

Please help me with the following task. The goal is to explore rooms and find the target room. In each room, you need to decide the direction to go in. Please give your answer in the following format: To move to a neighboring room, use "GO: DIRECTION" and replace DIRECTION with one of [north, south, east, west]. Most importantly, once we have found the target room, answer with "DONE" instead. Omit any other text.

Here is an example:

The target room is a Bedroom. You are in the Kitchen. Currently available directions: south, west. What is your next instruction?

GO: west

You have made a step and entered a Lobby. Currently available directions: east, north. What is your next instruction?

GO: north

...

You have made a step and entered a Bedroom. Currently available directions: south. What is your next instruction?

DONE

Let us start. The target room is \$GOAL\$. You are in the \$INITIAL\_ROOM\$. Currently available directions: \$INITIAL\_DIRECTIONS\$. What is your next instruction?

\label{pr:text_map_g2x}
\end{prompt}

\caption{Text-only Map World Game: Initial prompt template for the G2X version (\ref{pr:text_map_g2x})}
\label{fig:mapworld_textual_g2x}
\end{subfigure}
\vspace{10ex}
\begin{subfigure}{\columnwidth}
\centering
\hfill
\begin{prompt}

Please help me with the following task. The goal is to explore rooms and find target room. In each room I will show you an image of the room and tell you in what directions we can go from there. You then give me a description of the room you see in exactly one sentence. Please give your answer in the following format: "{"description": "<room description>", "action": "<action>"}". To move to a neighboring room, replace <action> with "GO: DIRECTION" where DIRECTION can be one of [north, south, east, west]. Most importantly, once we have found the target room, replace <action> with "DONE" instead. Omit any other text.

Here is an example:

The target room is a bathroom.
We have made a step and are now in this room. From here we can go: north, west. What is your next instruction?
{"description": "We are in a kitchen with a red fridge.", "action": "GO: north"}

We have made a step and are now in this room. From here we can go: south, east. What is your next instruction?
{"description": "We are in a living room with a couch and a tv.", "action": "GO: east"}

...
We have made a step and are now in this room. From here we can go: south, east. What is your next instruction?
{"description": "We are in a bathroom, there is a shower and a sink", "action": "DONE"}

Let us start. The target room is a \$GOAL\$
We are now in this room. From here we can go: \$INITIAL\_DIRECTIONS\$. What is your next instruction?

\label{pr:mm_map_g2x}

\end{prompt}

\caption{Multimodal Map World Game: Initial prompt template for the G2X version (\ref{pr:mm_map_g2x})}
\label{fig:mapworld_multimodal_g2x}
\end{subfigure}
\caption{Initial prompt and answer templates for the G2X version of Map World Game}
\label{fig:mapworld_prompts_g2x}
\end{figure*}

\begin{figure*}
\begin{subfigure}{\columnwidth}
\begin{prompt}

Please help me with the following task. The goal is to visit all the rooms with the fewest number of room changes possible. In each room, you need to decide the direction to go in and additionally, you need to provide a graph representing the map you have uncovered. Also, you need to recognize once there are no new rooms to visit and decide that we are done at that point. Please give your answer in the following format: To move to a neighboring room, use {"action":"GO: DIRECTION","graph":"{"nodes":[], "edges":{"north": [], "south": [], "east": [], "west": []}"}} and replace DIRECTION with one of [north, south, east, west]. To stop the exploration, answer with "DONE" instead. Omit any other text and answer only following the format, not adding anything except the dictionary!

Here is an example:

You are in the Living Room. Currently available directions: south, west. What is your next instruction?
{"action": "GO: west", "graph": {"nodes":["Living Room"],
"edges":{"north":[], "south":[], "east":[],
"west":[]}}}

You have made a step and entered a Library. Currently available directions: east, north. What is your next instruction?
{"action": "GO: north", "graph":{"nodes":["Living Room", "Library"], "edges":{"north":[], "south":[],
"east":[],"west":[("Living Room", "Library")]}}}

You have made a step and entered a Kitchen. Currently available directions: south, east. What is your next instruction?
{"action": "GO: east", "graph":{"nodes": ["Living Room", "Library", "Kitchen"], "edges":{"north": [("Library", "Kitchen")], "south": [], "east": [], "west": [("Living Room", "Library")]}}}

...

You have made a step and entered a Bedroom. Currently available directions: south, west. What is your next instruction?
{"action": "DONE", "graph": {...}}

Let us start. You are in the \$INITIAL\_ROOM\$. Currently available directions: \$INITIAL\_DIRECTIONS\$. What is your next instruction?

\label{pr:map_init_text_graph}
\end{prompt}

\caption{Text-only Map World Game: Initial prompt template for EE-gr version (\ref{pr:map_init_text_graph})}
\label{fig:mapworld_textual_graph}
\end{subfigure}
\begin{subfigure}{\columnwidth}
\centering
\hfill
\begin{prompt}

We are currently in this room. Please help me with the following task. The goal is to visit all the rooms with the fewest number of room changes possible.  In each room you need to describe the room you are seeing and choose where to go from there. Additionally, you need to provide a graph representing the map you have uncovered. Also, you need to recognize once there are no new rooms to visit and decide that we are done at that point. Please give your answer in the following format: 
'{"action":"<action>", "description": "<room description>", "graph": <graph>}'.
<action> needs to be in the format "GO: <direction>" where <direction> is one of [north, east, south, west]. Alternatively, choose "DONE" as your action once you have explored the entire map.
<room description> should be a single sentence describing the room shown to you.
<graph> represents the map in this format: {"nodes":[], "edges":{"north": [], "south": [], "east": [], "west": []}}
Omit any other text and answer only following the format, not adding anything except the dictionary!

Here is an example:

We are in this room. From here we can go: south, west. What is your next instruction?

{"action":"GO: north", "description": "We are in a kitchen with a red fridge.", "graph":{"nodes":["Kitchen"], "edges":{"north": [], "south": [], "east": [], "west": []}}}

We have made a step and are now in this room. From here we can go: east. What is your next instruction?

{"action":"GO: east", "description": "We are in a living room with a couch and a tv.", "graph":{"nodes":["Kitchen", "Living Room"], "edges":{"north": [["Kitchen", "Living Room"]], "south": [], "east": [], "west": []}}}

You have made a step and are now in this room. From here we can go: west, south. What is your next instruction?

{"action":"GO: south", "description": "We are in a bedroom with two beds and a nightstand.",  "graph":{"nodes":["Kitchen", "Living Room", "Bedroom"], "edges":{"north": [["Kitchen", "Living Room"]], "south": [], "east": [["Living Room", "Bedroom"]], "west": []}}}

...

You have made a step and are now in this room. From here we can go: north. What is your next instruction?

Example answer:
{"action":"DONE", "description": "We are in a stairwell, the stair is curved.", "graph":"{...}"}

Let us start.
Currently available directions: \$INITIAL\_DIRECTIONS\$. What is your next instruction?

\label{pr:map_init_mm_graph}
\end{prompt}

\caption{Multimodal Map World Game: Initial prompt template for EE-gr version (\ref{pr:map_init_mm_graph}) }
\label{fig:mapworld_multimodal_graph}
\end{subfigure}
\caption{Initial prompts for the EE-gr version of the Map World Game}
\label{fig:mapworld_prompts_graph}
\end{figure*}

%% file: arxiv.bbl
\begin{thebibliography}{53}
\providecommand{\natexlab}[1]{#1}

\bibitem[{Abdin et~al.(2024)Abdin, Jacobs, Awan, Aneja, Awadallah, Awadalla,
  Bach, Bahree, Bakhtiari, Behl, Benhaim, and
  et~al.}]{DBLP:journals/corr/abs-2404-14219}
Marah~I Abdin, Sam~Ade Jacobs, Ammar~Ahmad Awan, Jyoti Aneja, Ahmed Awadallah,
  Hany Awadalla, Nguyen Bach, Amit Bahree, Arash Bakhtiari, Harkirat~S. Behl,
  Alon Benhaim, and et~al. 2024.
\newblock \href {https://doi.org/10.48550/ARXIV.2404.14219} {Phi-3 technical
  report: {A} highly capable language model locally on your phone}.
\newblock \emph{CoRR}, abs/2404.14219.

\bibitem[{Beyer et~al.(2024)Beyer, Chalamalasetti, Hakimov, Madureira, Sadler,
  and Schlangen}]{beyer2024clembench2024}
Anne Beyer, Kranti Chalamalasetti, Sherzod Hakimov, Brielen Madureira, Philipp
  Sadler, and David Schlangen. 2024.
\newblock \href {https://arxiv.org/abs/2405.20859} {clembench$_{2024}$: A
  challenging, dynamic, complementary, multilingual benchmark and underlying
  flexible framework for llms as multi-action agents}.
\newblock \emph{Preprint}, arXiv:2405.20859.

\bibitem[{Brennan and Clark(1996)}]{brenclark:conpact}
Susan~E. Brennan and Herbert~H. Clark. 1996.
\newblock Conceptual pacts and lexical choice in conversation.
\newblock \emph{Journal of Experimental Psychology: Learning, Memory, and
  Cognition}, 22(6):1482--1493.

\bibitem[{Bubeck et~al.(2023)Bubeck, Chandrasekaran, Eldan, Gehrke, Horvitz,
  Kamar, Lee, Lee, Li, Lundberg, Nori, Palangi, Ribeiro, and Zhang}]{sparksagi}
S{\'{e}}bastien Bubeck, Varun Chandrasekaran, Ronen Eldan, Johannes Gehrke,
  Eric Horvitz, Ece Kamar, Peter Lee, Yin~Tat Lee, Yuanzhi Li, Scott~M.
  Lundberg, Harsha Nori, Hamid Palangi, Marco~T{\'{u}}lio Ribeiro, and
  Yi~Zhang. 2023.
\newblock \href {https://doi.org/10.48550/ARXIV.2303.12712} {Sparks of
  artificial general intelligence: Early experiments with {GPT-4}}.
\newblock \emph{CoRR}, abs/2303.12712.

\bibitem[{Chalamalasetti et~al.(2023)Chalamalasetti, G{\"o}tze, Hakimov,
  Madureira, Sadler, and Schlangen}]{chalamalasetti-etal-2023-clembench}
Kranti Chalamalasetti, Jana G{\"o}tze, Sherzod Hakimov, Brielen Madureira,
  Philipp Sadler, and David Schlangen. 2023.
\newblock \href {https://doi.org/10.18653/v1/2023.emnlp-main.689} {clembench:
  Using game play to evaluate chat-optimized language models as conversational
  agents}.
\newblock In \emph{Proceedings of the 2023 Conference on Empirical Methods in
  Natural Language Processing}, pages 11174--11219, Singapore. Association for
  Computational Linguistics.

\bibitem[{Chan et~al.(2023)Chan, Chen, Su, Yu, Xue, Zhang, Fu, and
  Liu}]{DBLP:journals/corr/abs-2308-07201}
Chi{-}Min Chan, Weize Chen, Yusheng Su, Jianxuan Yu, Wei Xue, Shanghang Zhang,
  Jie Fu, and Zhiyuan Liu. 2023.
\newblock \href {https://doi.org/10.48550/ARXIV.2308.07201} {Chateval: Towards
  better llm-based evaluators through multi-agent debate}.
\newblock \emph{CoRR}, abs/2308.07201.

\bibitem[{Chen et~al.(2023)Chen, Wu, Wang, Su, Chen, Xing, Zhong, Zhang, Zhu,
  Lu, Li, Luo, Lu, Qiao, and Dai}]{chen2023internvl}
Zhe Chen, Jiannan Wu, Wenhai Wang, Weijie Su, Guo Chen, Sen Xing, Muyan Zhong,
  Qinglong Zhang, Xizhou Zhu, Lewei Lu, Bin Li, Ping Luo, Tong Lu, Yu~Qiao, and
  Jifeng Dai. 2023.
\newblock Internvl: Scaling up vision foundation models and aligning for
  generic visual-linguistic tasks.
\newblock \emph{arXiv preprint arXiv:2312.14238}.

\bibitem[{Chiang et~al.(2024)Chiang, Zheng, Sheng, Angelopoulos, Li, Li, Zhang,
  Zhu, Jordan, Gonzalez, and Stoica}]{chiang2024chatbot}
Wei-Lin Chiang, Lianmin Zheng, Ying Sheng, Anastasios~Nikolas Angelopoulos,
  Tianle Li, Dacheng Li, Hao Zhang, Banghua Zhu, Michael Jordan, Joseph~E.
  Gonzalez, and Ion Stoica. 2024.
\newblock \href {https://arxiv.org/abs/2403.04132} {Chatbot arena: An open
  platform for evaluating llms by human preference}.
\newblock \emph{Preprint}, arXiv:2403.04132.

\bibitem[{Das et~al.(2017)Das, Kottur, Gupta, Singh, Yadav, Moura, Parikh, and
  Batra}]{das2017visual}
Abhishek Das, Satwik Kottur, Khushi Gupta, Avi Singh, Deshraj Yadav, Jos{\'{e}}
  M.~F. Moura, Devi Parikh, and Dhruv Batra. 2017.
\newblock \href {https://doi.org/10.1109/CVPR.2017.121} {Visual dialog}.
\newblock In \emph{2017 {IEEE} Conference on Computer Vision and Pattern
  Recognition, {CVPR} 2017, Honolulu, HI, USA, July 21-26, 2017}, pages
  1080--1089. {IEEE} Computer Society.

\bibitem[{de~Vries et~al.(2017)de~Vries, Strub, Chandar, Pietquin, Larochelle,
  and Courville}]{devries2017guesswhat}
Harm de~Vries, Florian Strub, Sarath Chandar, Olivier Pietquin, Hugo
  Larochelle, and Aaron~C. Courville. 2017.
\newblock \href {https://doi.org/10.1109/CVPR.2017.475} {Guesswhat?! visual
  object discovery through multi-modal dialogue}.
\newblock In \emph{2017 {IEEE} Conference on Computer Vision and Pattern
  Recognition, {CVPR} 2017, Honolulu, HI, USA, July 21-26, 2017}, pages
  4466--4475. {IEEE} Computer Society.

\bibitem[{Fu et~al.(2023)Fu, Chen, Shen, Qin, Zhang, Lin, Qiu, Lin, Yang,
  Zheng, Li, Sun, and Ji}]{DBLP:journals/corr/abs-2306-13394}
Chaoyou Fu, Peixian Chen, Yunhang Shen, Yulei Qin, Mengdan Zhang, Xu~Lin,
  Zhenyu Qiu, Wei Lin, Jinrui Yang, Xiawu Zheng, Ke~Li, Xing Sun, and Rongrong
  Ji. 2023.
\newblock \href {https://doi.org/10.48550/ARXIV.2306.13394} {{MME:} {A}
  comprehensive evaluation benchmark for multimodal large language models}.
\newblock \emph{CoRR}, abs/2306.13394.

\bibitem[{Gatt and Krahmer(2018)}]{Gatt2018}
Albert Gatt and Emiel Krahmer. 2018.
\newblock \href {https://doi.org/10.1613/jair.5714} {{Survey of the State of
  the Art in Natural Language Generation: Core tasks, applications and
  evaluation}}.
\newblock \emph{Journal of Artificial Intelligence Research}, 61:65--170.

\bibitem[{Goyal et~al.(2017)Goyal, Khot, Summers{-}Stay, Batra, and
  Parikh}]{DBLP:conf/cvpr/GoyalKSBP17}
Yash Goyal, Tejas Khot, Douglas Summers{-}Stay, Dhruv Batra, and Devi Parikh.
  2017.
\newblock \href {https://doi.org/10.1109/CVPR.2017.670} {Making the {V} in
  {VQA} matter: Elevating the role of image understanding in visual question
  answering}.
\newblock In \emph{2017 {IEEE} Conference on Computer Vision and Pattern
  Recognition, {CVPR} 2017, Honolulu, HI, USA, July 21-26, 2017}, pages
  6325--6334. {IEEE} Computer Society.

\bibitem[{Gu et~al.(2022)Gu, Stefani, Wu, Thomason, and
  Wang}]{gu-etal-2022-vision}
Jing Gu, Eliana Stefani, Qi~Wu, Jesse Thomason, and Xin Wang. 2022.
\newblock \href {https://doi.org/10.18653/v1/2022.acl-long.524}
  {Vision-and-language navigation: A survey of tasks, methods, and future
  directions}.
\newblock In \emph{Proceedings of the 60th Annual Meeting of the Association
  for Computational Linguistics (Volume 1: Long Papers)}, pages 7606--7623,
  Dublin, Ireland. Association for Computational Linguistics.

\bibitem[{Gurari et~al.(2018)Gurari, Li, Stangl, Guo, Lin, Grauman, Luo, and
  Bigham}]{DBLP:conf/cvpr/Gurari0SGLGLB18}
Danna Gurari, Qing Li, Abigale~J. Stangl, Anhong Guo, Chi Lin, Kristen Grauman,
  Jiebo Luo, and Jeffrey~P. Bigham. 2018.
\newblock \href {https://doi.org/10.1109/CVPR.2018.00380} {Vizwiz grand
  challenge: Answering visual questions from blind people}.
\newblock In \emph{2018 {IEEE} Conference on Computer Vision and Pattern
  Recognition, {CVPR} 2018, Salt Lake City, UT, USA, June 18-22, 2018}, pages
  3608--3617. Computer Vision Foundation / {IEEE} Computer Society.

\bibitem[{Haber et~al.(2019)Haber, Baumg{\"a}rtner, Takmaz, Gelderloos, Bruni,
  and Fern{\'a}ndez}]{haber-etal-2019-photobook}
Janosch Haber, Tim Baumg{\"a}rtner, Ece Takmaz, Lieke Gelderloos, Elia Bruni,
  and Raquel Fern{\'a}ndez. 2019.
\newblock \href {https://doi.org/10.18653/v1/P19-1184} {The {P}hoto{B}ook
  dataset: Building common ground through visually-grounded dialogue}.
\newblock In \emph{Proceedings of the 57th Annual Meeting of the Association
  for Computational Linguistics}, pages 1895--1910, Florence, Italy.
  Association for Computational Linguistics.

\bibitem[{He et~al.(2024)He, Liu, Chen, Tian, Liu, Chi, Liu, Yuan, Xing, Wang,
  Dai, Zhang, Xue, Liu, Guo, and Chen}]{DBLP:journals/corr/abs-2405-19334}
Yingqing He, Zhaoyang Liu, Jingye Chen, Zeyue Tian, Hongyu Liu, Xiaowei Chi,
  Runtao Liu, Ruibin Yuan, Yazhou Xing, Wenhai Wang, Jifeng Dai, Yong Zhang,
  Wei Xue, Qifeng Liu, Yike Guo, and Qifeng Chen. 2024.
\newblock \href {https://doi.org/10.48550/ARXIV.2405.19334} {Llms meet
  multimodal generation and editing: {A} survey}.
\newblock \emph{CoRR}, abs/2405.19334.

\bibitem[{Ilinykh et~al.(2019)Ilinykh, Zarrie{\ss}, and Schlangen}]{mapworld}
Nikolai Ilinykh, Sina Zarrie{\ss}, and David Schlangen. 2019.
\newblock \href {http://semdial.org/anthology/Z19-Ilinykh_semdial_0006.pdf}
  {Meet up! a corpus of joint activity dialogues in a visual environment}.
\newblock In \emph{Proceedings of the 23rd Workshop on the Semantics and
  Pragmatics of Dialogue - Full Papers}, London, United Kingdom. SEMDIAL.

\bibitem[{Johnson et~al.(2017)Johnson, Hariharan, van~der Maaten, Fei{-}Fei,
  Zitnick, and Girshick}]{DBLP:conf/cvpr/JohnsonHMFZG17}
Justin Johnson, Bharath Hariharan, Laurens van~der Maaten, Li~Fei{-}Fei,
  C.~Lawrence Zitnick, and Ross~B. Girshick. 2017.
\newblock \href {https://doi.org/10.1109/CVPR.2017.215} {{CLEVR:} {A}
  diagnostic dataset for compositional language and elementary visual
  reasoning}.
\newblock In \emph{2017 {IEEE} Conference on Computer Vision and Pattern
  Recognition, {CVPR} 2017, Honolulu, HI, USA, July 21-26, 2017}, pages
  1988--1997. {IEEE} Computer Society.

\bibitem[{Krauss and Weinheimer(1964)}]{krausswein:1964}
Robert~M. Krauss and Sidney Weinheimer. 1964.
\newblock Changes in reference phrases as a function of frequency of usage in
  social interaction: A preliminary study.
\newblock \emph{Psychonomic Science}, 1:266--278.

\bibitem[{Krishna et~al.(2017)Krishna, Zhu, Groth, Johnson, Hata, Kravitz,
  Chen, Kalantidis, Li, Shamma, Bernstein, and
  Fei{-}Fei}]{krishna2017visgenome}
Ranjay Krishna, Yuke Zhu, Oliver Groth, Justin Johnson, Kenji Hata, Joshua
  Kravitz, Stephanie Chen, Yannis Kalantidis, Li{-}Jia Li, David~A. Shamma,
  Michael~S. Bernstein, and Li~Fei{-}Fei. 2017.
\newblock \href {https://doi.org/10.1007/S11263-016-0981-7} {Visual genome:
  Connecting language and vision using crowdsourced dense image annotations}.
\newblock \emph{Int. J. Comput. Vis.}, 123(1):32--73.

\bibitem[{Laurençon et~al.(2023)Laurençon, Saulnier, Tronchon, Bekman, Singh,
  Lozhkov, Wang, Karamcheti, Rush, Kiela, Cord, and Sanh}]{idefics}
Hugo Laurençon, Lucile Saulnier, Léo Tronchon, Stas Bekman, Amanpreet Singh,
  Anton Lozhkov, Thomas Wang, Siddharth Karamcheti, Alexander~M. Rush, Douwe
  Kiela, Matthieu Cord, and Victor Sanh. 2023.
\newblock \href {https://arxiv.org/abs/2306.16527} {Obelics: An open web-scale
  filtered dataset of interleaved image-text documents}.
\newblock \emph{Preprint}, arXiv:2306.16527.

\bibitem[{Laurençon et~al.(2024)Laurençon, Tronchon, Cord, and
  Sanh}]{laurençon2024mattersbuildingvisionlanguagemodels}
Hugo Laurençon, Léo Tronchon, Matthieu Cord, and Victor Sanh. 2024.
\newblock \href {https://arxiv.org/abs/2405.02246} {What matters when building
  vision-language models?}
\newblock \emph{Preprint}, arXiv:2405.02246.

\bibitem[{Lewis(1969)}]{lewis:conv}
David Lewis. 1969.
\newblock \emph{Convention}.
\newblock Harvard University Press.

\bibitem[{Li et~al.(2023{\natexlab{a}})Li, Ge, Ge, Wang, Wang, Zhang, and
  Shan}]{seedbench-2}
Bohao Li, Yuying Ge, Yixiao Ge, Guangzhi Wang, Rui Wang, Ruimao Zhang, and Ying
  Shan. 2023{\natexlab{a}}.
\newblock \href {https://doi.org/10.48550/ARXIV.2311.17092} {Seed-bench-2:
  Benchmarking multimodal large language models}.
\newblock \emph{CoRR}, abs/2311.17092.

\bibitem[{Li et~al.(2023{\natexlab{b}})Li, Wang, Wang, Ge, Ge, and
  Shan}]{seedbench-1}
Bohao Li, Rui Wang, Guangzhi Wang, Yuying Ge, Yixiao Ge, and Ying Shan.
  2023{\natexlab{b}}.
\newblock \href {https://doi.org/10.48550/ARXIV.2307.16125} {Seed-bench:
  Benchmarking multimodal llms with generative comprehension}.
\newblock \emph{CoRR}, abs/2307.16125.

\bibitem[{Li et~al.(2024)Li, Zhang, Zhou, Collier, Korhonen, and
  Vulic}]{DBLP:journals/corr/abs-2406-02537}
Chengzu Li, Caiqi Zhang, Han Zhou, Nigel Collier, Anna Korhonen, and Ivan
  Vulic. 2024.
\newblock \href {https://doi.org/10.48550/ARXIV.2406.02537} {Topviewrs:
  Vision-language models as top-view spatial reasoners}.
\newblock \emph{CoRR}, abs/2406.02537.

\bibitem[{Liang et~al.(2022)Liang, Bommasani, Lee, Tsipras, Soylu, Yasunaga,
  and \textit{alia}}]{helm2022}
Percy Liang, Rishi Bommasani, Tony Lee, Dimitris Tsipras, Dilara Soylu,
  Michihiro Yasunaga, and \textit{alia}. 2022.
\newblock \href {https://doi.org/10.48550/arXiv.2211.09110} {Holistic
  evaluation of language models}.
\newblock \emph{CoRR}, abs/2211.09110.

\bibitem[{Liu and Wu(2023)}]{liu2023evaluating}
Chang Liu and Bo~Wu. 2023.
\newblock \href {https://arxiv.org/abs/2308.11224} {Evaluating large language
  models on graphs: Performance insights and comparative analysis}.
\newblock \emph{Preprint}, arXiv:2308.11224.

\bibitem[{Liu et~al.(2023)Liu, Duan, Zhang, Li, Zhang, Zhao, Yuan, Wang, He,
  Liu, Chen, and Lin}]{DBLP:journals/corr/abs-2307-06281}
Yuan Liu, Haodong Duan, Yuanhan Zhang, Bo~Li, Songyang Zhang, Wangbo Zhao, Yike
  Yuan, Jiaqi Wang, Conghui He, Ziwei Liu, Kai Chen, and Dahua Lin. 2023.
\newblock \href {https://doi.org/10.48550/ARXIV.2307.06281} {Mmbench: Is your
  multi-modal model an all-around player?}
\newblock \emph{CoRR}, abs/2307.06281.

\bibitem[{Lopes et~al.(2018)Lopes, Hemmingsson, and
  {\AA}strand}]{lopes2018spot}
Jos{\'{e}} Lopes, Nils Hemmingsson, and Oliver {\AA}strand. 2018.
\newblock \href
  {http://www.lrec-conf.org/proceedings/lrec2018/summaries/410.html} {The spot
  the difference corpus: a multi-modal corpus of spontaneous task oriented
  spoken interactions}.
\newblock In \emph{Proceedings of the Eleventh International Conference on
  Language Resources and Evaluation, {LREC} 2018, Miyazaki, Japan, May 7-12,
  2018}. European Language Resources Association {(ELRA)}.

\bibitem[{Magar and Schwartz(2022)}]{magar-schwartz-2022-data}
Inbal Magar and Roy Schwartz. 2022.
\newblock \href {https://doi.org/10.18653/v1/2022.acl-short.18} {Data
  contamination: From memorization to exploitation}.
\newblock In \emph{Proceedings of the 60th Annual Meeting of the Association
  for Computational Linguistics (Volume 2: Short Papers)}, pages 157--165,
  Dublin, Ireland. Association for Computational Linguistics.

\bibitem[{Momennejad et~al.(2023)Momennejad, Hasanbeig, Frujeri, Sharma,
  Redmond, sharma Robert~Osazuwa, Ness, Jojic, Palangi, and
  Larson}]{Momennejad2023EvaluatingCM}
Ida Momennejad, Hosein Hasanbeig, Felipe~Vieira Frujeri, Hiteshi Sharma,
  Microsoft~Research Redmond, sharma Robert~Osazuwa, Ness, Nebojsa Jojic, Hamid
  Palangi, and Jonathan Larson. 2023.
\newblock \href {https://api.semanticscholar.org/CorpusID:263257355}
  {Evaluating cognitive maps in large language models with cogeval: No emergent
  planning}.
\newblock In \emph{Proceedings of NeurIPS 2023}.

\bibitem[{Onoe et~al.(2024)Onoe, Rane, Berger, Bitton, Cho, Garg, Ku, Parekh,
  Pont-Tuset, Tanzer, Wang, and Baldridge}]{onoe2024docci}
Yasumasa Onoe, Sunayana Rane, Zachary Berger, Yonatan Bitton, Jaemin Cho,
  Roopal Garg, Alexander Ku, Zarana Parekh, Jordi Pont-Tuset, Garrett Tanzer,
  Su~Wang, and Jason Baldridge. 2024.
\newblock \href {https://arxiv.org/abs/2404.19753} {Docci: Descriptions of
  connected and contrasting images}.
\newblock \emph{Preprint}, arXiv:2404.19753.

\bibitem[{Qiao et~al.(2023)Qiao, Wu, Liang, Li, and
  Duan}]{DBLP:journals/corr/abs-2308-10032}
Dan Qiao, Chenfei Wu, Yaobo Liang, Juntao Li, and Nan Duan. 2023.
\newblock \href {https://doi.org/10.48550/ARXIV.2308.10032} {Gameeval:
  Evaluating llms on conversational games}.
\newblock \emph{CoRR}, abs/2308.10032.

\bibitem[{Radford et~al.(2021)Radford, Kim, Hallacy, Ramesh, Goh, Agarwal,
  Sastry, Askell, Mishkin, Clark, Krueger, and Sutskever}]{radford2021clip}
Alec Radford, Jong~Wook Kim, Chris Hallacy, Aditya Ramesh, Gabriel Goh,
  Sandhini Agarwal, Girish Sastry, Amanda Askell, Pamela Mishkin, Jack Clark,
  Gretchen Krueger, and Ilya Sutskever. 2021.
\newblock \href {http://proceedings.mlr.press/v139/radford21a.html} {Learning
  transferable visual models from natural language supervision}.
\newblock In \emph{Proceedings of the 38th International Conference on Machine
  Learning, {ICML} 2021, 18-24 July 2021, Virtual Event}, volume 139 of
  \emph{Proceedings of Machine Learning Research}, pages 8748--8763. {PMLR}.

\bibitem[{Rizvi et~al.(2024)Rizvi, Zhu, and
  Gurevych}]{DBLP:journals/corr/abs-2406-04566}
Md~Imbesat~Hassan Rizvi, Xiaodan Zhu, and Iryna Gurevych. 2024.
\newblock \href {https://doi.org/10.48550/ARXIV.2406.04566} {Sparc and sparp:
  Spatial reasoning characterization and path generation for understanding
  spatial reasoning capability of large language models}.
\newblock \emph{CoRR}, abs/2406.04566.

\bibitem[{Sadler et~al.(2024)Sadler, Hakimov, and
  Schlangen}]{sadler-etal-2024-sharing-cost}
Philipp Sadler, Sherzod Hakimov, and David Schlangen. 2024.
\newblock \href {https://aclanthology.org/2024.lrec-main.1287} {Sharing the
  cost of success: A game for evaluating and learning collaborative multi-agent
  instruction giving and following policies}.
\newblock In \emph{Proceedings of the 2024 Joint International Conference on
  Computational Linguistics, Language Resources and Evaluation (LREC-COLING
  2024)}, pages 14770--14783, Torino, Italia. ELRA and ICCL.

\bibitem[{Sadovnik et~al.(2012)Sadovnik, Chiu, Snavely, Edelman, and
  Chen}]{DBLP:conf/cvpr/SadovnikCSEC12}
Amir Sadovnik, Yi{-}I Chiu, Noah Snavely, Shimon Edelman, and Tsuhan Chen.
  2012.
\newblock \href {https://doi.org/10.1109/CVPR.2012.6248003} {Image description
  with a goal: Building efficient discriminating expressions for images}.
\newblock In \emph{2012 {IEEE} Conference on Computer Vision and Pattern
  Recognition, Providence, RI, USA, June 16-21, 2012}, pages 2791--2798. {IEEE}
  Computer Society.

\bibitem[{Sagi et~al.(2012)Sagi, Gentner, and Lovett}]{sagi2012difference}
Eyal Sagi, Dedre Gentner, and Andrew~M. Lovett. 2012.
\newblock \href {https://doi.org/10.1111/J.1551-6709.2012.01250.X} {What
  difference reveals about similarity}.
\newblock \emph{Cogn. Sci.}, 36(6):1019--1050.

\bibitem[{Schlangen(2019)}]{schlangen2019grounded}
David Schlangen. 2019.
\newblock \href {https://arxiv.org/abs/1908.11279} {Grounded agreement games:
  Emphasizing conversational grounding in visual dialogue settings}.
\newblock \emph{CoRR}, abs/1908.11279.

\bibitem[{Schlangen(2023{\natexlab{a}})}]{Schlangen-2023-1}
David Schlangen. 2023{\natexlab{a}}.
\newblock \href {https://doi.org/10.48550/arXiv.2304.07007} {Dialogue games for
  benchmarking language understanding: Motivation, taxonomy, strategy}.
\newblock \emph{CoRR}, abs/2304.07007.

\bibitem[{Schlangen(2023{\natexlab{b}})}]{schlangen-2023-general}
David Schlangen. 2023{\natexlab{b}}.
\newblock \href {https://doi.org/10.18653/v1/2023.findings-emnlp.591} {On
  general language understanding}.
\newblock In \emph{Findings of the Association for Computational Linguistics:
  EMNLP 2023}, pages 8818--8825, Singapore. Association for Computational
  Linguistics.

\bibitem[{Schlangen(2023{\natexlab{c}})}]{schlangen-2023}
David Schlangen. 2023{\natexlab{c}}.
\newblock \href {https://doi.org/10.48550/arXiv.2302.08590} {{What A Situated
  Language-Using Agent Must be Able to Do: {A} Top-Down Analysis}}.
\newblock \emph{CoRR}, abs/2302.08590.

\bibitem[{Shen et~al.(2018)Shen, Hofer, Felbo, and
  Levy}]{shen-etal-2018-comparing}
Judy~Hanwen Shen, Matthias Hofer, Bjarke Felbo, and Roger Levy. 2018.
\newblock \href {https://doi.org/10.18653/v1/K18-1029} {Comparing models of
  associative meaning: An empirical investigation of reference in simple
  language games}.
\newblock In \emph{Proceedings of the 22nd Conference on Computational Natural
  Language Learning}, pages 292--301, Brussels, Belgium. Association for
  Computational Linguistics.

\bibitem[{Shi et~al.(2022)Shi, Zhang, and Lipani}]{DBLP:conf/aaai/ShiZL22}
Zhengxiang Shi, Qiang Zhang, and Aldo Lipani. 2022.
\newblock \href {https://doi.org/10.1609/AAAI.V36I10.21383} {Stepgame: {A} new
  benchmark for robust multi-hop spatial reasoning in texts}.
\newblock In \emph{Thirty-Sixth {AAAI} Conference on Artificial Intelligence,
  {AAAI} 2022, Thirty-Fourth Conference on Innovative Applications of
  Artificial Intelligence, {IAAI} 2022, The Twelveth Symposium on Educational
  Advances in Artificial Intelligence, {EAAI} 2022 Virtual Event, February 22 -
  March 1, 2022}, pages 11321--11329. {AAAI} Press.

\bibitem[{Srivastava et~al.(2022)Srivastava, Rastogi, Rao, Shoeb, Abid, Fisch,
  Brown, Santoro, Gupta, Garriga{-}Alonso, Kluska, Lewkowycz, Agarwal, and
  et~al.}]{bigbench2022}
Aarohi Srivastava, Abhinav Rastogi, Abhishek Rao, Abu Awal~Md Shoeb, Abubakar
  Abid, Adam Fisch, Adam~R. Brown, Adam Santoro, Aditya Gupta, Adri{\`{a}}
  Garriga{-}Alonso, Agnieszka Kluska, Aitor Lewkowycz, Akshat Agarwal, and
  et~al. 2022.
\newblock \href {https://doi.org/10.48550/arXiv.2206.04615} {Beyond the
  imitation game: Quantifying and extrapolating the capabilities of language
  models}.
\newblock \emph{CoRR}, abs/2206.04615.

\bibitem[{Suglia et~al.(2024)Suglia, Konstas, and Lemon}]{Suglia2024}
Alessandro Suglia, Ioannis Konstas, and Oliver Lemon. 2024.
\newblock {Visually Grounded Language Learning: a review of language games,
  datasets, tasks, and models}.
\newblock \emph{Journal of Artificial Intelligence Research}, 79:173--239.

\bibitem[{Wang et~al.(2019)Wang, Pruksachatkun, Nangia, Singh, Michael, Hill,
  Levy, and Bowman}]{superGLUE}
Alex Wang, Yada Pruksachatkun, Nikita Nangia, Amanpreet Singh, Julian Michael,
  Felix Hill, Omer Levy, and Samuel~R. Bowman. 2019.
\newblock \href {https://arxiv.org/abs/1905.00537} {{SuperGLUE: A Stickier
  Benchmark for General-Purpose Language Understanding Systems}}.
\newblock In \emph{NeurIPS}, July, pages 1--30.

\bibitem[{Wang et~al.(2023)Wang, Feng, He, Tan, Han, and
  Tsvetkov}]{wang2023language}
Heng Wang, Shangbin Feng, Tianxing He, Zhaoxuan Tan, Xiaochuang Han, and Yulia
  Tsvetkov. 2023.
\newblock \href {https://arxiv.org/abs/2305.10037} {Can language models solve
  graph problems in natural language?}
\newblock \emph{Preprint}, arXiv:2305.10037.

\bibitem[{Yue et~al.(2023)Yue, Ni, Zhang, Zheng, Liu, Zhang, Stevens, Jiang,
  and et~al.}]{mmmu}
Xiang Yue, Yuansheng Ni, Kai Zhang, Tianyu Zheng, Ruoqi Liu, Ge~Zhang, Samuel
  Stevens, Dongfu Jiang, and et~al. 2023.
\newblock \href {https://doi.org/10.48550/ARXIV.2311.16502} {{MMMU:} {A}
  massive multi-discipline multimodal understanding and reasoning benchmark for
  expert {AGI}}.
\newblock \emph{CoRR}, abs/2311.16502.

\bibitem[{Zhang et~al.(2024)Zhang, Dong, Zang, Cao, Qian, Chen, Guo, Duan, and
  et~al.}]{internlmxcomposer2_5}
Pan Zhang, Xiaoyi Dong, Yuhang Zang, Yuhang Cao, Rui Qian, Lin Chen, Qipeng
  Guo, Haodong Duan, and et~al. 2024.
\newblock Internlm-xcomposer-2.5: A versatile large vision language model
  supporting long-contextual input and output.
\newblock \emph{arXiv preprint arXiv:2407.03320}.

\bibitem[{Zhou et~al.(2017)Zhou, Zhao, Puig, Fidler, Barriuso, and
  Torralba}]{DBLP:conf/cvpr/ZhouZPFB017}
Bolei Zhou, Hang Zhao, Xavier Puig, Sanja Fidler, Adela Barriuso, and Antonio
  Torralba. 2017.
\newblock \href {https://doi.org/10.1109/CVPR.2017.544} {Scene parsing through
  {ADE20K} dataset}.
\newblock In \emph{2017 {IEEE} Conference on Computer Vision and Pattern
  Recognition, {CVPR} 2017, Honolulu, HI, USA, July 21-26, 2017}, pages
  5122--5130. {IEEE} Computer Society.

\end{thebibliography}
